\documentclass{article}

\usepackage{microtype}
\usepackage{graphicx}
\usepackage{booktabs} 

\usepackage{hyperref}

 \usepackage[accepted]{icml2023}

\usepackage[utf8]{inputenc} 
\usepackage[T1]{fontenc}    
\usepackage{url}            
\usepackage{amsfonts}       
\usepackage{nicefrac}       
\usepackage{xcolor, color, colortbl}         

\usepackage{amsmath}
\usepackage{amssymb}
\usepackage{mathtools}
\usepackage{amsthm}
\usepackage{bm}
\usepackage{pifont}
\usepackage{enumitem}

\usepackage{subcaption}
\usepackage{graphics}
\usepackage{graphicx}
\usepackage{caption}
\usepackage{wrapfig}

\usepackage{threeparttable}
\usepackage{adjustbox}
\usepackage{multirow}
\usepackage{threeparttable}
\usepackage{arydshln}
\usepackage{stfloats}

\usepackage[capitalize,noabbrev]{cleveref}
\usepackage{minitoc}
\input{insbox}

\theoremstyle{plain}
\newtheorem{theorem}{Theorem}

\newtheorem{lemma}{Lemma}

\theoremstyle{definition}
\newtheorem{definition}{Definition}

\theoremstyle{remark}

\newcommand*\diff{\mathop{}\!\mathrm{d}}
\newcommand{\cmark}{\ding{51}}%
\newcommand{\xmark}{\ding{55}}%
\definecolor{Gray}{gray}{0.9}

\newcommand{\cc}[1]{\cellcolor{gray!#1}}
\newcolumntype{g}{>{\columncolor{Gray}}c}
\newcolumntype{f}{>{\columncolor{Gray}}l}
\makeatletter
\def\smallunderbrace#1{\mathop{\vtop{\m@th\ialign{##\crcr
				$\hfil\displaystyle{#1}\hfil$\crcr
				\noalign{\kern3\p@\nointerlineskip}%
				\tiny\upbracefill\crcr\noalign{\kern3\p@}}}}\limits}
\makeatother
\makeatletter
\allowdisplaybreaks
\newcommand{\printfnsymbol}[1]{%
	\textsuperscript{\@fnsymbol{#1}}%
}
\def\smalloverbrace#1{\mathop{\vtop{\m@th\ialign{##\crcr
				$\hfil\displaystyle{#1}\hfil$\crcr
				\noalign{\kern3\p@\nointerlineskip}%
				\tiny\upbracefill\crcr\noalign{\kern3\p@}}}}\limits}
\def\adl@drawiv#1#2#3{%
	\hskip.5\tabcolsep
	\xleaders#3{#2.5\@tempdimb #1{1}#2.5\@tempdimb}%
	#2\z@ plus1fil minus1fil\relax
	\hskip.5\tabcolsep}
\newcommand{\cdashlinelr}[1]{%
	\noalign{\vskip\aboverulesep
		\global\let\@dashdrawstore\adl@draw
		\global\let\adl@draw\adl@drawiv}
	\cdashline{#1}
	\noalign{\global\let\adl@draw\@dashdrawstore
		\vskip\belowrulesep}}				
\newcommand{\wrapfill}{\par\ifnum\value{WF@wrappedlines}>0
	\addtocounter{WF@wrappedlines}{-1}%
	\null\vspace{\arabic{WF@wrappedlines}\baselineskip}%
	\WFclear
	\fi}
\newcommand{\appropto}{\mathrel{\vcenter{
			\offinterlineskip\halign{\hfil$##$\cr
				\propto\cr\noalign{\kern2pt}\sim\cr\noalign{\kern-2pt}}}}}

\usepackage[textsize=tiny]{todonotes}

\icmltitlerunning{Refining Generative Process with Discriminator Guidance in Score-based Diffusion Models}

\begin{document}

\twocolumn[
\icmltitle{Refining Generative Process with Discriminator Guidance\\in Score-based Diffusion Models}



\icmlsetsymbol{equal}{*}

\begin{icmlauthorlist}
\icmlauthor{Dongjun Kim}{equal,kaist}
\icmlauthor{Yeongmin Kim}{equal,kaist}
\icmlauthor{Se Jung Kwon}{comp}
\icmlauthor{Wanmo Kang}{kaist}
\icmlauthor{Il-Chul Moon}{kaist,comp2}
\end{icmlauthorlist}

\icmlaffiliation{kaist}{KAIST, South Korea}
\icmlaffiliation{comp}{NAVER Cloud}
\icmlaffiliation{comp2}{Summary.AI}

\icmlcorrespondingauthor{Dongjun Kim}{dongjoun57@kaist.ac.kr}

\icmlkeywords{Machine Learning, ICML}

\vskip 0.3in
]



\printAffiliationsAndNotice{\icmlEqualContribution} 

\begin{figure*}[hb]
	\centering
	\includegraphics[width=0.95\linewidth]{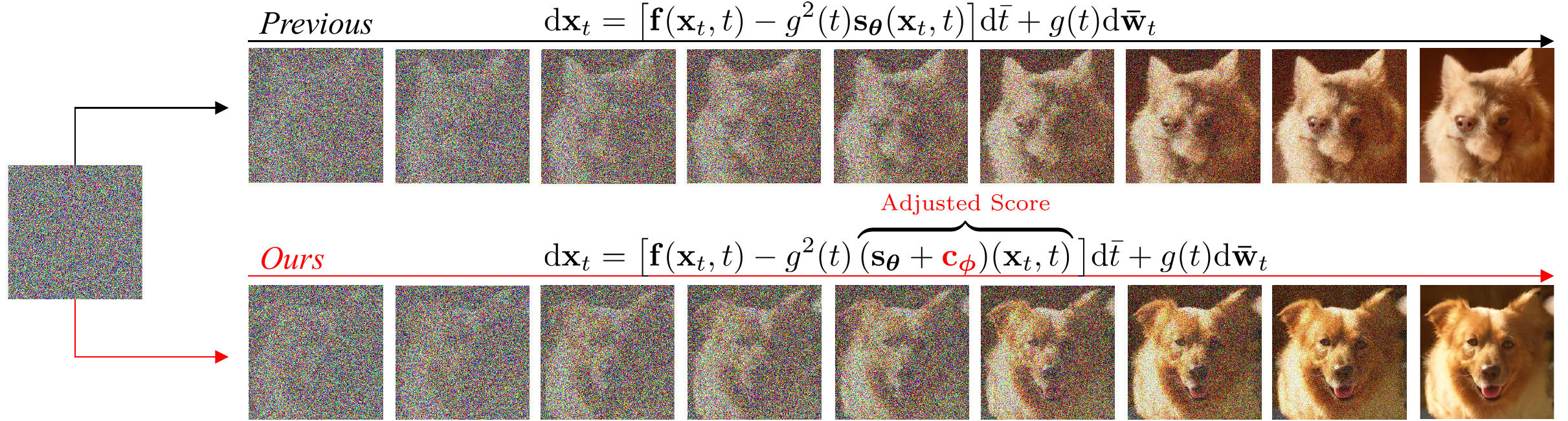}
	\caption{Comparison of the denoising processes. Discriminator Guidance adjusts the score function by estimating the gap $\mathbf{c}_{\bm{\phi}}$ between the predicted model score and the true data score. As a result, the sample generated using Discriminator Guidance is indistinguishable from real data according to the discriminator.}
	\label{fig:thumbnail}
\end{figure*}

\begin{abstract}
	The proposed method, \textbf{Discriminator Guidance}, aims to improve sample generation of pre-trained diffusion models. The approach introduces a discriminator that gives explicit supervision to a denoising sample path whether it is realistic or not. Unlike GANs, our approach does not require joint training of score and discriminator networks. Instead, we train the discriminator after score training, making discriminator training stable and fast to converge. In sample generation, we add an auxiliary term to the pre-trained score to deceive the discriminator. This term corrects the model score to the data score at the optimal discriminator, which implies that the discriminator helps better score estimation in a complementary way. Using our algorithm, we achive state-of-the-art results on ImageNet 256x256 with FID 1.83 and recall 0.64, similar to the validation data's FID (1.68) and recall (0.66). We release the code at \url{https://github.com/alsdudrla10/DG}.
\end{abstract}

\section{Introduction}

The diffusion model has recently been highlighted for its success in image generation \cite{dhariwal2021diffusion, ho2022cascaded, karras2022elucidating, song2020score}, video generation \cite{singer2022make, ho2022video, voleti2022mcvd}, and text-to-image generation \cite{rombach2022high, ramesh2022hierarchical, saharia2022photorealistic}. The State-Of-The-Art (SOTA) models perform human-level generation, but there is still much more room to be investigated for a deep understanding on diffusion models.	

\begin{figure*}[t]
	\centering
	\begin{subfigure}{0.29\linewidth}
		\centering
		\includegraphics[width=\linewidth]{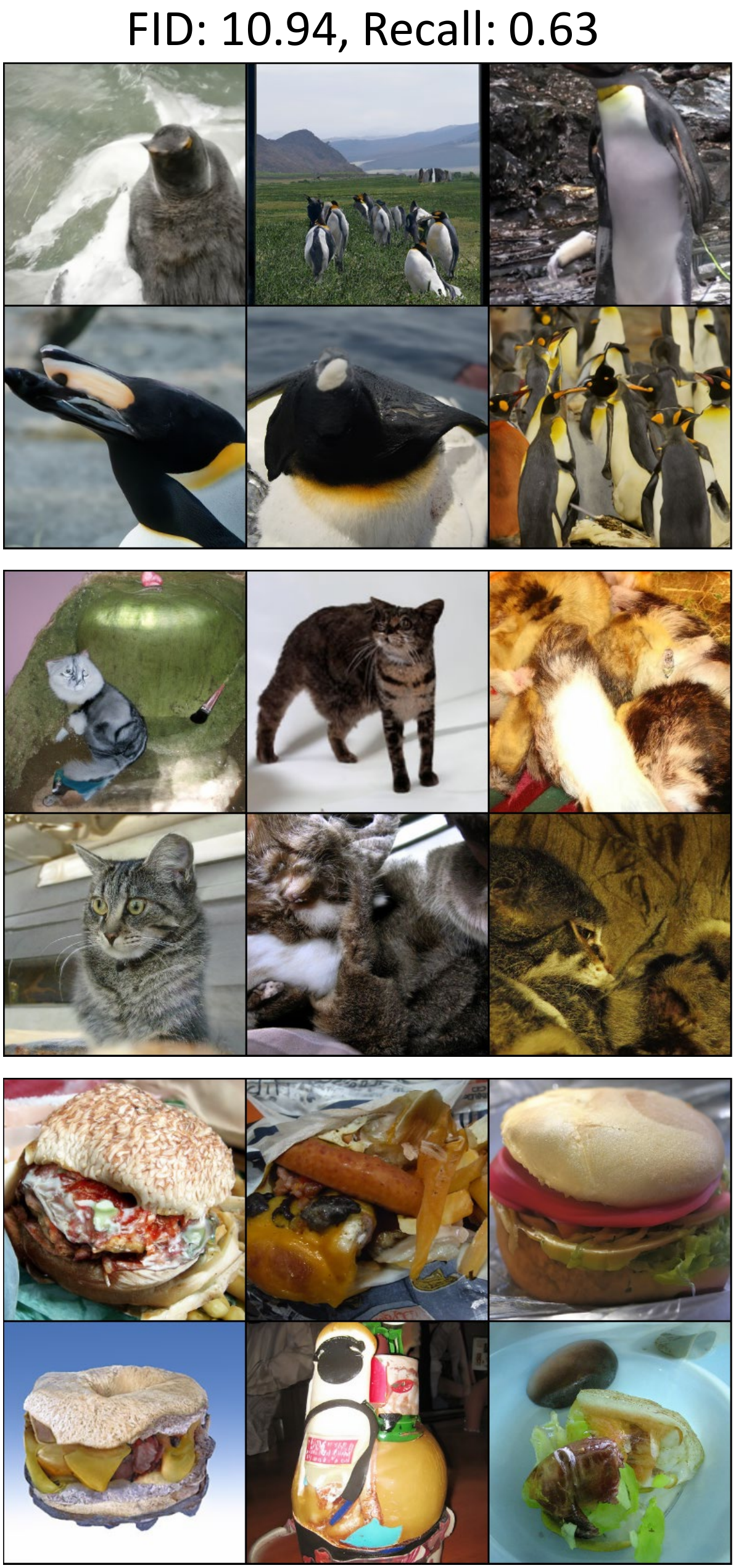}
		\subcaption{ADM \cite{dhariwal2021diffusion}}
	\end{subfigure}
	\hfil
	\begin{subfigure}{0.29\linewidth}
		\centering
		\includegraphics[width=\linewidth]{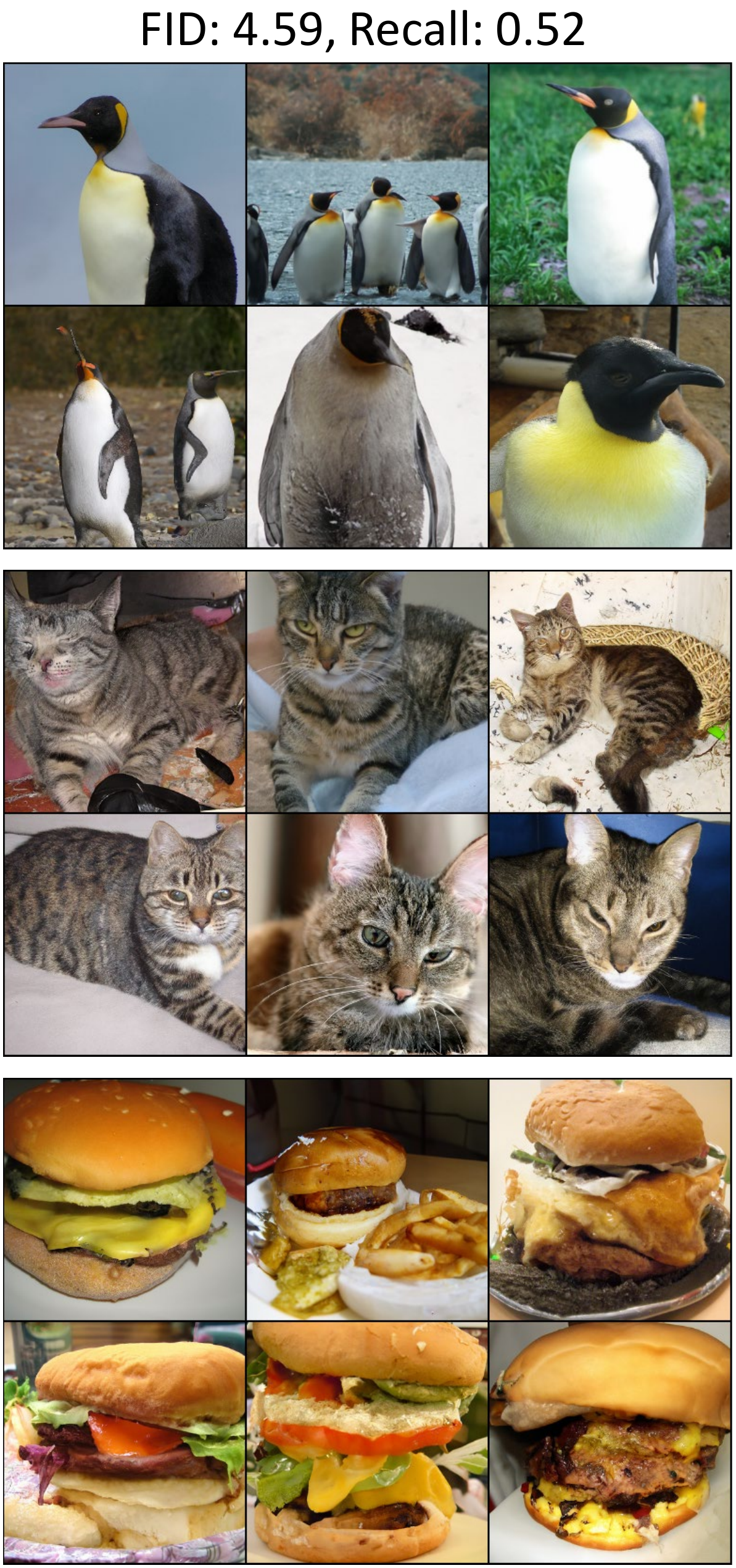}
		\subcaption{ADM-G \cite{dhariwal2021diffusion}}
	\end{subfigure}		
	\hfil
	\begin{subfigure}{0.29\linewidth}
		\centering
		\includegraphics[width=\linewidth]{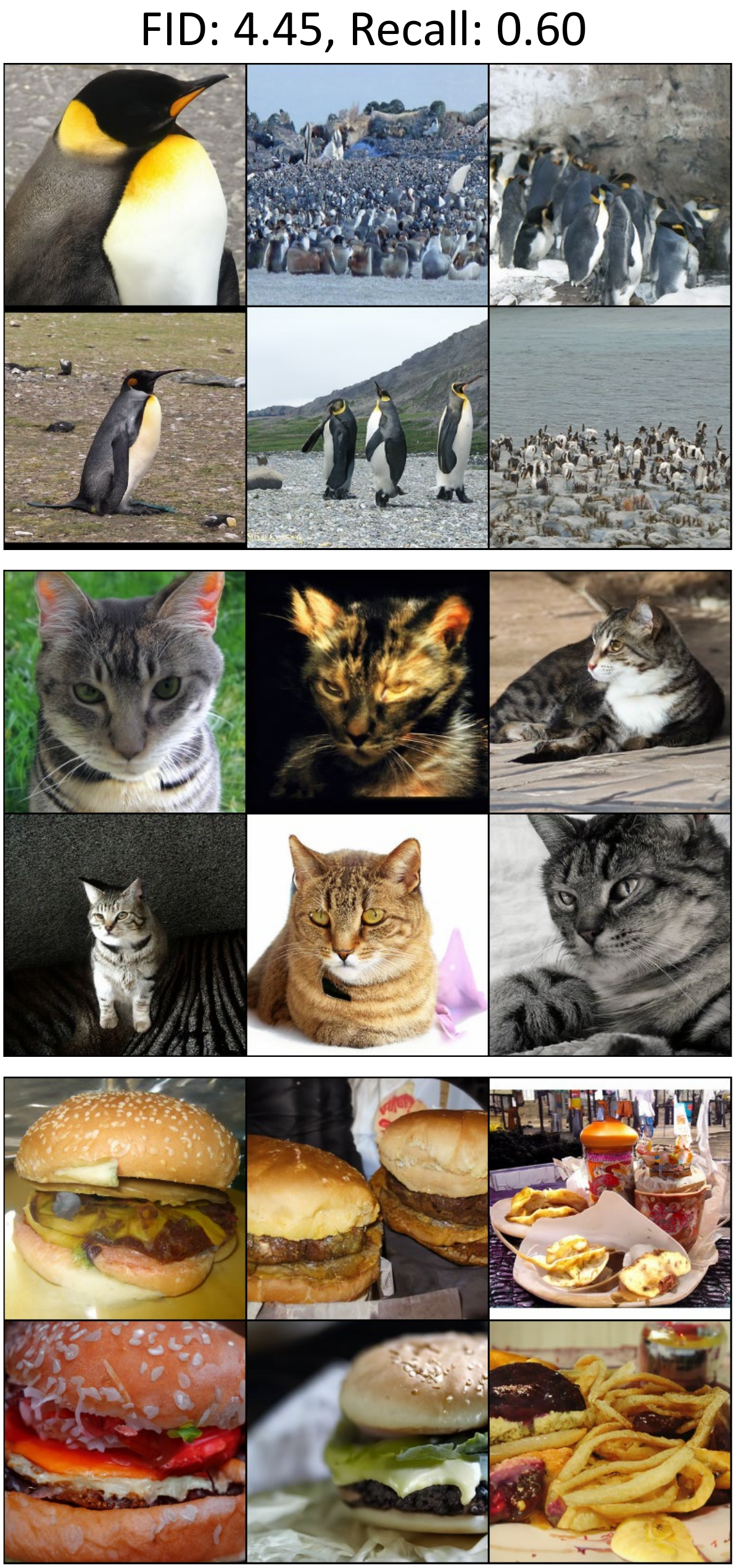}
		\subcaption{ADM-G++ (Ours)}
	\end{subfigure}
	\caption{Samples from ImageNet 256x256 on (a) ADM, (b) ADM with Classifier Guidance. Classifier Guidance generates high-fidelity but mode degenerated samples. (c) Classifier Guidance combined with Discriminator Guidance improves both sample quality and intra-class diversity. See Appendix \ref{sec:uncurated} for uncurated samples of SOTA models.}
	\label{fig:ImageNet256}
	\vskip -0.1in
\end{figure*}

The generative model community widely uses well-trained score models \cite{dhariwal2021diffusion, rombach2022high} in downstream tasks \cite{meng2021sdedit, kawar2022denoising, su2022dual, kim2022unsupervised}. This is partially because training a new score model from scratch can be computationally expensive. However, as the demand for reusing pre-trained models increases, there are only a few research  efforts that focus on improving sample quality with a pre-trained score model.

To avoid issues such as overfitting \cite{nichol2021improved} or memorization \cite{carlini2023extracting} that may arise from further score training (Figure \ref{fig:further_score_training}), our approach keeps the pre-trained score model fixed and introduces a new component that provides a supervision during sample generation. Specifically, we propose using a discriminator as an auxiliary degree of freedom to the pre-trained model. This discriminator classifies real and generated data at all noise scales, providing direct feedback to the sample denoising process, indicating whether the sample path is realistic or not. We achieve this by adding a correction term to the model score, constructed by the discriminator, which steers the sample path towards more realistic regions (Figure \ref{fig:thumbnail}). This term is designed to adjust the model score to match the data score at the optimal discriminator (Theorem \ref{thm:1}), allowing our approach to find a realistic sample path by adjusting the model score. In experiments, we achieve new SOTA performances on image datasets such as CIFAR-10, CelebA/FFHQ 64x64, and ImageNet 256x256. As discriminator training is a minimization problem that is stable and fast to converge (Figure \ref{fig:discriminator_training}), such a significant gain can be achieved with a cheap budget (Table \ref{tab:budget}). We summarize the contributions as follows.

\vspace{-2mm}
\begin{itemize}\setlength\itemsep{0.2em}
	\item[\checkmark] We propose a new generative process, \textbf{Discriminator Guidance}, with an adjusted score of a given \textit{pre-trained} score model.
	\item[\checkmark] We show that the discriminator-guided samples are \textit{closer} to the real-world data than the non-guided samples, theoretically and empirically.
\end{itemize}
\vspace{-2mm}

\section{Preliminary}

Suppose $p_{r}(\mathbf{x}_{0})$ be the data distribution and $p_{\bm{\theta}}(\mathbf{x}_{0})$ be the model distribution. Likelihood-based latent variable models optimize their parameters by minimizing the upper bound of the KL divergence $D_{KL}(p_{r}(\mathbf{x}_{0})\Vert p_{\bm{\theta}}(\mathbf{x}_{0}))$, given by
\begin{align*}
D_{KL}(p_{r}(\mathbf{x}_{0})\Vert p_{\bm{\theta}}(\mathbf{x}_{0}))\le D_{KL}(q(\mathbf{x}_{0:T})\Vert p_{\bm{\theta}}(\mathbf{x}_{0:T})),
\end{align*}
where $\mathbf{x}_{1:T}$ are $T$ latent variables; $q(\mathbf{x}_{0:T})$ is an inference distribution with marginal density $q(\mathbf{x}_{0}):=p_{r}(\mathbf{x}_{0})$; and $p_{\bm{\theta}}(\mathbf{x}_{0:T})$ is a generative distribution with marginal density $p_{\bm{\theta}}(\mathbf{x}_{T}):=\pi(\mathbf{x}_{T})$, where $\pi$ is an easy-to-sample prior distribution for generation purpose.

Denoising Diffusion Probabilistic Models (DDPM) \cite{ho2020denoising} perturb the data variable $\mathbf{x}_{0}$ step-by-step to construct $\mathbf{x}_{1:T}$ by adding iterative Gaussian noises, leading $q$ to be a non-parametrized fixed inference distribution with $q(\mathbf{x}_{0:T})=p_{r}(\mathbf{x}_{0})\prod_{t=1}^{T}q(\mathbf{x}_{t}\vert\mathbf{x}_{t-1})$. Most \cite{okhotin2023star} of diffusion models assume a Markov chain for the generative process so to satisfy $p_{\bm{\theta}}(\mathbf{x}_{0:T})=\pi(\mathbf{x}_{T})\prod_{t=1}^{T}p_{\bm{\theta}}(\mathbf{x}_{t-1}\vert\mathbf{x}_{t})$, and this modeling choice enables to optimize the surrogate objective $D_{KL}(q(\mathbf{x}_{0:T})\Vert p_{\bm{\theta}}(\mathbf{x}_{0:T}))$ in a tractable way.

The continuous-time counterpart \cite{song2020score} of DDPM describes the diffusion process in the language of stochastic differential equations (SDE) by
\begin{align}\label{eq:forward_sde}
\diff\mathbf{x}_{t}=\mathbf{f}(\mathbf{x}_{t},t)\diff t+g(t)\diff \mathbf{w}_{t},
\end{align}
with $t$ now being a continuum of the diffusion index in $[0,T]$, and $\mathbf{f}(\mathbf{x}_{t},t)$ and $g(t)$ being the drift and the volatility coefficients, respectively. We describe our model under the continuous-time framework mainly for notational simplicity. Our model is applicable to both discrete- and continuous-time settings.

Under the continuous-time framework, the forward-time diffusion process of Eq. \eqref{eq:forward_sde} has a unique reverse-time diffusion process \cite{anderson1982reverse}
\begin{align}\label{eq:reverse_sde}
\diff\mathbf{x}_{t}=\big[\mathbf{f}(\mathbf{x}_{t},t)-g^{2}(t)\nabla\log{p_{r}^{t}(\mathbf{x}_{t})}\big]\diff \bar{t}+g(t)\diff\bar{\mathbf{w}}_{t},
\end{align}
where $\diff\bar{t}$ and $\bar{\mathbf{w}}_{t}$ are the infinitesimal reverse-time and the reverse-time Brownian motion, respectively. Subsequently, the continuous-time generative process becomes
\begin{align*}
\diff\mathbf{x}_{t}=\big[\mathbf{f}(\mathbf{x}_{t},t)-g^{2}(t)\mathbf{s}_{\bm{\theta}}(\mathbf{x}_{t},t)\big]\diff \bar{t}+g(t)\diff\bar{\mathbf{w}}_{t},
\end{align*}
where the estimation target of the score network $\mathbf{s}_{\bm{\theta}}(\mathbf{x}_{t},t)$ is the actual data score $\nabla\log{p_{r}^{t}(\mathbf{x}_{t})}$. Here, $p_{r}^{t}$ is the diffused probability density of the data distribution following the forward-time diffusion process in Eq. \eqref{eq:forward_sde}.

The continuous-time model trains the score network with the denoising score matching loss \cite{song2019generative}
\begin{align*}
\mathcal{L}_{\bm{\theta}}=\frac{1}{2}\int_{0}^{T}\xi(t)\mathbb{E}\big[\Vert\mathbf{s}_{\bm{\theta}}(\mathbf{x}_{t},t)-\nabla\log{p_{0t}(\mathbf{x}_{t}\vert\mathbf{x}_{0})}\Vert_{2}^{2}\big]\diff t,
\end{align*}
where $\xi$ is the temporal weight and $p_{0t}$ is the transition probability from $\mathbf{x}_{0}$ to $\mathbf{x}_{t}$. This denoising score objective coincides to the joint KL divergence $D_{KL}(q(\mathbf{x}_{0:T})\Vert p_{\bm{\theta}}(\mathbf{x}_{0:T}))$ if $\xi(t)=g^{2}(t)$ \cite{chen2016relation, song2021maximum}. Also, under different weighting functions, this objective could be equivalently interpreted as the noise matching loss $\int_{0}^{T}\mathbb{E}[\Vert\bm{\epsilon}_{\bm{\theta}}-\bm{\epsilon}\Vert_{2}^{2}]$ \cite{ho2020denoising} or the data reconstruction loss $\int_{0}^{T}\mathbb{E}[\Vert\hat{\mathbf{x}}_{\bm{\theta}}(\mathbf{x}_{t})-\mathbf{x}_{0}\Vert_{2}^{2}]$ \cite{kingma2021variational}.

There are various approaches to enhance the precision of score training. For instance, \citet{kim2022soft, kingma2023understanding, hang2023efficient} have proposed updating the score network using Maximum Perturbed Likelihood Estimation to improve large-time denoising accuracy. Conversely, \citet{lai2022regularizing, daras2023consistent} have studied the invariant characteristics of the data diffusion process and recommended adding an extra regularization term to the denoising score loss to meet these invariant properties. Our work, on the other hand, aims to refine the fixed model score with noise contrastive estimation, which is distinct from prior attempts to improve score accuracy.

\begin{algorithm}[t]
	\centering
	\caption{Discriminator Training}\label{alg:discriminator}
	\begin{algorithmic}[1]
		\STATE Construct $\mathcal{D}=\{\mathbf{x}_{1},...,\mathbf{x}_{M}\}$ from the real-world
		\STATE Construct $\mathcal{G}=\{\mathbf{\hat{x}}_{1},...,\mathbf{\hat{x}}_{N}\}$ by sampling from $p_{\bm{\theta}}$
		\WHILE{converged}
		\STATE Sample $\mathbf{x}_{1},...,\mathbf{x}_{B/2}$ from the real dataset $\mathcal{D}$
		\STATE Sample $\mathbf{x}_{B/2+1},...,\mathbf{x}_{B}$ from the sample dataset $\mathcal{G}$
		\STATE Sample $t_{1},...,t_{B}$ from $[0,T]$
		\STATE Diffuse $\mathbf{x}_{i}^{t_{i}}\leftarrow e^{-\int_{0}^{t_{i}}\beta_{s}\diff s}\mathbf{x}_{i}+\sqrt{1-e^{-\int_{0}^{t_{i}}\beta_{s}\diff s}}\bm{\epsilon}_{i}$ for $\bm{\epsilon}_{i}\sim\mathcal{N}(0,\mathbf{I})$, $\forall i=1,...,B$
		\STATE Calculate $\mathcal{\hat{L}}_{\bm{\phi}}\leftarrow-\sum_{i=1}^{B/2}\lambda(t_{i})\log{d_{\bm{\phi}}(\mathbf{x}_{i}^{t_{i}},t_{i})}-\sum_{i=B/2+1}^{B}\lambda(t_{i})\log{(1-d_{\bm{\phi}}(\mathbf{x}_{i}^{t_{i}},t_{i}))}$
		\STATE Update $\bm{\phi}\leftarrow \bm{\phi}-\frac{\partial\mathcal{\hat{L}}_{\bm{\phi}}}{\partial\bm{\phi}}$
		\ENDWHILE
	\end{algorithmic}
\end{algorithm}

\section{Refining Generative Process with Discriminator Guidance}\label{sec:methodology}

\subsection{Correction of Pre-trained Model Score}\label{sec:error}

After score training, we synthesize samples with the time-reversal generative process
\begin{align}\label{eq:generative_process}
\diff\mathbf{x}_{t}=\big[\mathbf{f}(\mathbf{x}_{t},t)-g^{2}(t)\mathbf{s}_{\bm{\theta}_{\infty}}(\mathbf{x}_{t},t)\big]\diff \bar{t}+g(t)\diff\bar{\mathbf{w}}_{t},
\end{align}
where $\mathbf{s}_{\bm{\theta}_{\infty}}$ represents the score network after the convergence. This generative process could differ from the reverse-time data process if the local optimum $\bm{\theta}_{\infty}$ deviates from the global optimum $\bm{\theta}_{*}$. We show in Theorem \ref{thm:1} that the generative process of Eq. \eqref{eq:generative_process} coincides with the data process of Eq. \eqref{eq:reverse_sde} if we adjust the model score. We call this gap by the \textit{correction term}, which is \textit{nonzero} as long as $\bm{\theta}_{\infty}\neq \bm{\theta}_{*}$.
\begin{theorem}\label{thm:1}
	Suppose $p_{\bm{\theta}_{\infty}}$ be the solution of the time-reversal generative process of Eq. \eqref{eq:generative_process}. Let $p_{r}^{t}$ and $p_{\bm{\theta}_{\infty}}^{t}$ be the marginal densities (at $t$) of the forward-time SDE $\diff\mathbf{x}_{t}=\mathbf{f}(\mathbf{x}_{t},t)\diff t+g(t)\diff \mathbf{w}_{t}$ starting from $p_{r}$ and $p_{\bm{\theta}_{\infty}}$, respectively. If $\mathbf{s}_{\bm{\theta}_{\infty}}(\mathbf{x},T)=\nabla\log{\pi(\mathbf{x})}$, where $\pi$ is the prior distribution, and the log-likelihood $\log{p_{\bm{\theta}_{\infty}}}$ equals its evidence lower bound $\mathcal{L}_{\bm{\theta}_{\infty}}$, then the reverse-time SDE
	\begin{align*}
	\diff\mathbf{x}_{t}=\big[\mathbf{f}(\mathbf{x}_{t},t)-g^{2}(t)\nabla\log{p_{r}^{t}(\mathbf{x}_{t})}\big]\diff \bar{t}+g(t)\diff\bar{\mathbf{w}}_{t},
	\end{align*}
	coincides with a diffusion process with adjusted score,
	\begin{align*}
	\diff\mathbf{x}_{t}&=\big[\mathbf{f}(\mathbf{x}_{t},t)-g^{2}(t)(\mathbf{s}_{\bm{\theta}_{\infty}}+\mathbf{c}_{\bm{\theta}_{\infty}})(\mathbf{x}_{t},t)\big]\diff\bar{t}+g(t)\diff\bar{\mathbf{w}}_{t},
	\end{align*}
	for $\mathbf{c}_{\bm{\theta}_{\infty}}(\mathbf{x}_{t},t):=\nabla\log{\frac{p_{r}^{t}(\mathbf{x}_{t})}{p_{\bm{\theta}_{\infty}}^{t}(\mathbf{x}_{t})}}$.
\end{theorem}

\subsection{Discriminator Guidance}\label{sec:adjusted_generative_process}

\begin{figure}[t]
	\centering
	\includegraphics[width=0.9\linewidth]{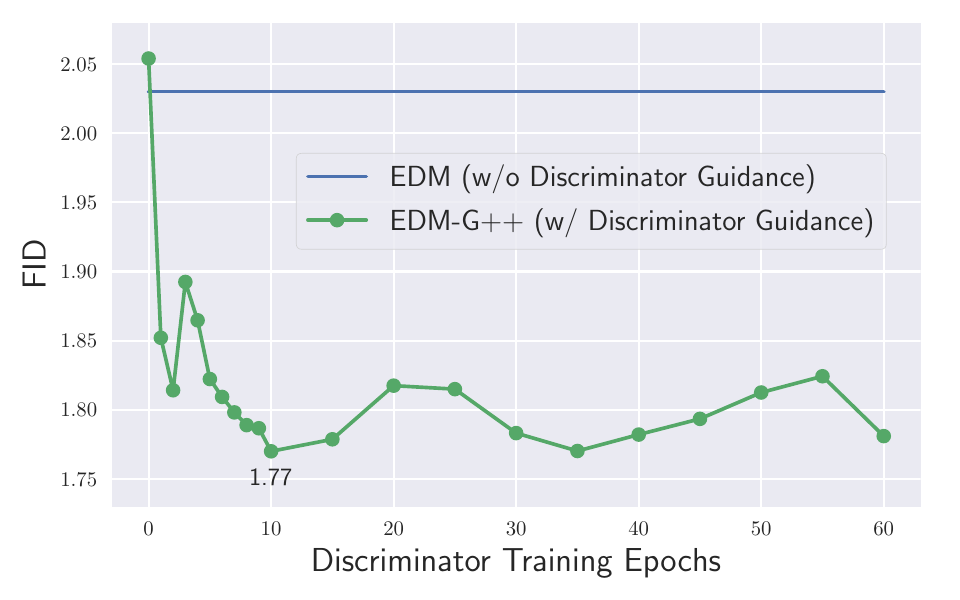}
	\vskip -0.1in
	\caption{Discriminator guidance refines FID on CIFAR-10.}
	\label{fig:discriminator_training}
\end{figure}

The correction term $\mathbf{c}_{\bm{\theta}_{\infty}}(\mathbf{x}_{t},t)=\nabla\log{\frac{p_{r}^{t}(\mathbf{x}_{t})}{p_{\bm{\theta}_{\infty}}^{t}(\mathbf{x}_{t})}}$ is intractable in general because the density-ratio $\frac{p_{r}^{t}}{p_{\bm{\theta}_{\infty}}^{t}}$ is inaccessible. Therefore, we estimate this density-ratio by training a discriminator at all noise level $t$. For discriminator training, we first draw fake samples from the generative process of Eq. \eqref{eq:generative_process} as many as data instances. Then we classify the real and fake data using the noise-embedded Binary Cross Entropy (BCE)
\begin{align}\label{eq:discriminator_loss}
\begin{split}
\mathcal{L}_{\bm{\phi}}&=\int \lambda(t)\big(\mathbb{E}_{p_{r}^{t}(\mathbf{x}_{t})}[-\log{d_{\bm{\phi}}(\mathbf{x}_{t},t)}]\\
&\quad+\mathbb{E}_{p_{\bm{\theta}_{\infty}}^{t}(\mathbf{x}_{t})}[-\log{(1-d_{\bm{\phi}}(\mathbf{x}_{t},t))}]\big)\diff t,
\end{split}
\end{align}
where $\lambda$ is the temporal weight, see Algorithm \ref{alg:discriminator} and Appendix \ref{sec:forward_instead_of_generative} for details.

As the correction term is represented by
\begin{align*}
\mathbf{c}_{\bm{\theta}_{\infty}}(\mathbf{x}_{t},t)=\nabla\log{\frac{d_{\bm{\phi}_{*}}(\mathbf{x}_{t},t)}{1-d_{\bm{\phi}_{*}}(\mathbf{x}_{t},t)}},
\end{align*}
in terms of the optimal discriminator $d_{\bm{\phi}_{*}}$ of $\mathcal{L}_{\bm{\phi}}$, we estimate the correction term $\mathbf{c}_{\bm{\theta}_{\infty}}$ with a neural discriminator $d_{\bm{\phi}}$ by
\begin{align*}
\mathbf{c}_{\bm{\theta}_{\infty}}(\mathbf{x}_{t},t)\approx\mathbf{c}_{\bm{\phi}}(\mathbf{x}_{t},t):=\nabla\log{\frac{d_{\bm{\phi}}(\mathbf{x}_{t},t)}{1-d_{\bm{\phi}}(\mathbf{x}_{t},t)}}.
\end{align*}
With the above tractable correction estimate, we define the \textbf{Discriminator Guidance} (DG) by
\begin{align}\label{eq:dg}
\diff\mathbf{x}_{t}&=\big[\mathbf{f}(\mathbf{x}_{t},t)-g^{2}(t)(\mathbf{s}_{\bm{\theta}_{\infty}}+\mathbf{c}_{\bm{\phi}})(\mathbf{x}_{t},t)\big]\diff\bar{t}+g(t)\diff\bar{\mathbf{w}}_{t}.
\end{align}
Figure \ref{fig:discriminator_training} shows that the discriminator indeed improves sample quality with a quick convergence. 

\begin{figure}[t]
	\centering
	\includegraphics[width=0.9\linewidth]{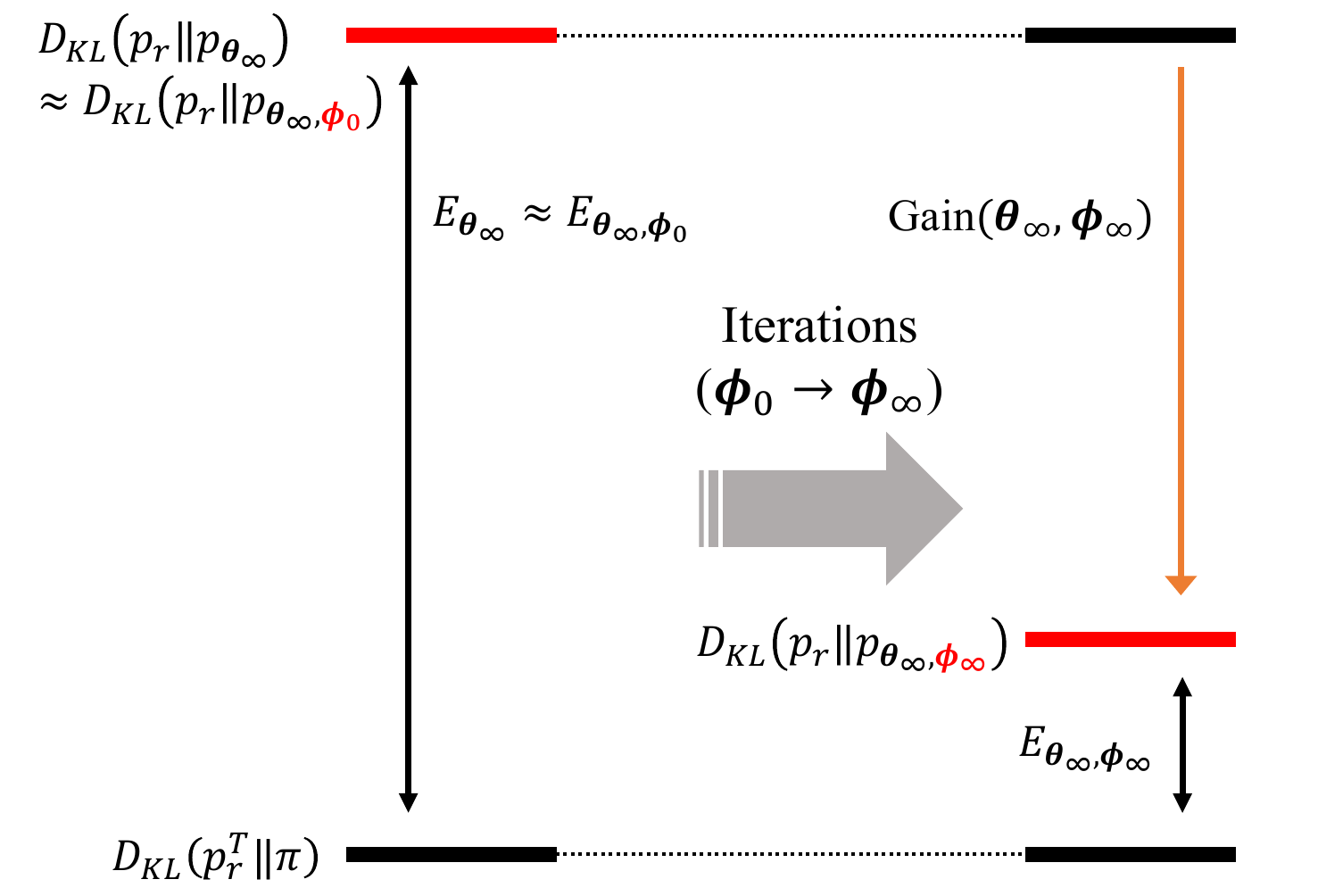}
	\caption{Schematic illustration of the analysis in Section \ref{sec:theory}. The gain increases as discriminator learns.}
	\label{fig:loss_visualization}
\end{figure}
\subsection{Theoretical Analysis}\label{sec:theory}
Although we introduced Discriminator Guidance in the context of differential equations, this section examines the approach from the perspective of statistical divergence between the data and the sample distributions. Specifically, we define $p_{\bm{\theta}_{\infty},\bm{\phi}}$ as the discriminator-guided sample distribution of Eq. \eqref{eq:dg}. The central question becomes
\begin{align*}
\textit{Is $p_{\bm{\theta}_{\infty},\bm{\phi}}$ closer to the data distribution $p_{r}$ than $p_{\bm{\theta}_{\infty}}$?}
\end{align*}
We answer the question in Theorem \ref{thm:2}. 
\begin{theorem}\label{thm:2}
	If the assumptions of Theorem \ref{thm:1} hold, then
	\begin{align*}
	&D_{KL}(p_{r}\Vert p_{\bm{\theta}_{\infty}})=D_{KL}(p_{r}^{T}\Vert \pi)+E_{\bm{\theta}_{\infty}},\\
	&D_{KL}(p_{r}\Vert p_{\bm{\theta}_{\infty},\bm{\phi}})\le D_{KL}(p_{r}^{T}\Vert\pi)+E_{\bm{\theta}_{\infty},\bm{\phi}},
	\end{align*}
	where $E_{\bm{\theta}_{\infty}}$ is the score error
	\begin{align*}
	E_{\bm{\theta}_{\infty}}=\frac{1}{2}\int_{0}^{T}g^{2}(t)\mathbb{E}_{p_{r}^{t}}\big[\Vert\nabla\log{p_{r}^{t}}-\mathbf{s}_{\bm{\theta}_{\infty}}\Vert_{2}^{2}\big]\diff t,
	\end{align*}
	and $E_{\bm{\theta}_{\infty},\bm{\phi}}$ is the discriminator-adjusted score error
	\begin{align*}
	&E_{\bm{\theta}_{\infty},\bm{\phi}}=\frac{1}{2}\int_{0}^{T}g^{2}(t)\mathbb{E}_{p_{r}^{t}}\big[\Vert\nabla\log{p_{r}^{t}}-(\mathbf{s}_{\bm{\theta}_{\infty}}+\mathbf{c}_{\bm{\phi}})\Vert_{2}^{2}\big]\diff t\\
	&\!\quad\quad\quad=\frac{1}{2}\int_{0}^{T}g^{2}(t)\mathbb{E}_{p_{r}^{t}}\big[\Vert \mathbf{c}_{\bm{\theta}_{\infty}}-\mathbf{c}_{\bm{\phi}}\Vert_{2}^{2}\big]\diff t.
	\end{align*}
\end{theorem}

\begin{table}[t]
	\caption{Discriminator-adjusted score error $E_{\bm{\theta}_{\infty},\bm{\phi}}$ and corresponding Gain.}
	\label{tab:gain}
	\scriptsize
	\centering
		\begin{tabular}{lcc}
			\toprule
			Discriminator & $E_{\bm{\theta}_{\infty},\bm{\phi}}$ & Gain \\\midrule
			Blind $d_{\bm{\phi}_{b}}(\equiv 0.5)$ & $E_{\bm{\theta}_{\infty}}$ & 0 \\
			Optimal $d_{\bm{\phi}_{*}}$ & 0 & $E_{\bm{\theta}_{\infty}}$ (Maximum)\\\cdashlinelr{1-3}
			Untrained $d_{\bm{\phi}_{0}}(\approx 0.5)$ & $\approx E_{\bm{\theta}_{\infty}}$ & $\approx 0$ \\
			Trained $d_{\bm{\phi}_{\infty}}$ & $\ll E_{\bm{\theta}_{\infty}}$ & $\nearrow E_{\bm{\theta}_{\infty}}$ \\
			\bottomrule
		\end{tabular}
\end{table}
To measure the effect of discriminator training, we use Theorem \ref{thm:2} to compute the gain by subtracting two KLs,
\begin{align*}
&D_{KL}(p_{r}\Vert p_{\bm{\theta}_{\infty},\bm{\phi}})\le D_{KL}(p_{r}\Vert p_{\bm{\theta}_{\infty}})-\textup{Gain}(\bm{\theta}_{\infty},\bm{\phi}),
\end{align*}
where $\textup{Gain}(\bm{\theta}_{\infty},\bm{\phi})=E_{\bm{\theta}_{\infty}}-E_{\bm{\theta}_{\infty},\bm{\phi}}$ represents the difference between the score error and the discriminator-adjusted score error. Note that while Theorem \ref{thm:2} does not guarantee that the gain is strictly positive, it is initialized near zero and gradually increases throughout discriminator training, as summarized in Table \ref{tab:gain}. Specifically, when the discriminator is completely blind ($d_{\bm{\phi}_{b}}\equiv 0.5$), there is no signal from the discriminator gradient, and the discriminator-adjusted score error $E_{\bm{\theta}_{\infty},\bm{\phi}_{b}}$ equals the score error $E_{\bm{\theta}_{\infty}}$. Therefore, the gain is approximately zero when the discriminator is untrained ($d_{\bm{\phi}_{0}}\approx 0.5$) as shown in Figure \ref{fig:discriminator_training}. On the other hand, at the optimal discriminator $d_{\bm{\phi}_{*}}$, the neural correction $\mathbf{c}_{\bm{\phi}_{*}}$ matches the target correction $\mathbf{c}_{\bm{\theta}_{\infty}}$ and satisfies $E_{\bm{\theta}_{\infty},\bm{\phi}_{*}}=0$, allowing Gain to be maximized as discriminator parameters are updated. See Figure \ref{fig:loss_visualization} for a schematic visualization.

In other words, we can interpret that Discriminator Guidance introduces an additional axial degree of freedom $\bm{\phi}$ that reparametrizes the score error $E_{\bm{\theta}_{\infty}}$ into a discriminator-adjusted score error $E_{\bm{\theta}_{\infty},\bm{\phi}}$. As a result, the score error $E_{\bm{\theta}_{\infty}}$ is no longer optimized with the denoising score loss $\mathcal{L}_{\bm{\theta}}$, but the reparametrized error $E_{\bm{\theta}_{\infty},\bm{\phi}}$ can be further optimized with an alternative loss $\mathcal{L}_{\bm{\phi}}$ of Eq. \eqref{eq:discriminator_loss}.

\subsection{Optimality Analysis}\label{sec:mode_coverage}

\begin{figure}[t]
	\centering
	\includegraphics[width=0.9\linewidth]{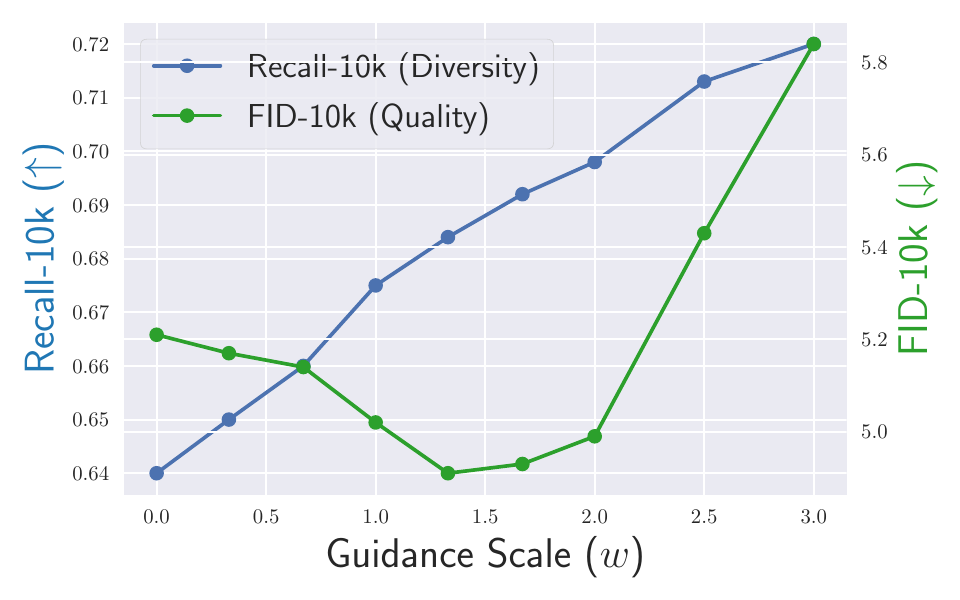}
	\caption{FID and Recall trade-off on DiT-XL-G++.}
	\vskip -0.1in
	\label{fig:precision_recall}
\end{figure}

Let's take a closer look at the score component. Our discriminator training is stable because we keep a pre-trained score model fixed during training, unlike unstable GAN training. Hence, after the discriminator has reached its optimal point, the resulting adjusted model score is given by
\begin{align*}
&\mathbf{s}_{\bm{\theta}_{\infty}}(\mathbf{x}_{t},t)+w \mathbf{c}_{\bm{\phi}_{*}}(\mathbf{x}_{t},t)\\
&\quad=\nabla\log{p_{\bm{\theta}_{\infty}}(\mathbf{x}_{t})}+w\nabla\log{\frac{p_{r}^{t}(\mathbf{x}_{t})}{p_{\bm{\theta}_{\infty}}^{t}(\mathbf{x}_{t})}}\\
&\quad=\nabla\log{\big[(p_{r}^{t}(\mathbf{x}_{t}))^{w}(p_{\bm{\theta}_{\infty}}^{t}(\mathbf{x}_{t}))^{1-w}\big]},
\end{align*}
Therefore, the sample distribution $(p_{r}^{t})^{w}(p_{\bm{\theta}_{\infty}}^{t})^{1-w}$ balances data distribution and non-guided distribution. The argument also holds for the conditional case, leading DG as a controller for the intra-class diversity. Figure \ref{fig:precision_recall} experiments on ImageNet, demonstrating that there is a sweet spot of DG weight regarding both quality (FID) and diversity (recall).

\subsection{Connection with Classifier Guidance}

Classifier Guidance (CG) \cite{dhariwal2021diffusion} is a milestone technique to guide a sample with a pre-trained classifier $p_{\psi_{\infty}}(c\vert\mathbf{x}_{t},t)$. The classifier-guided generative process is $\diff\mathbf{x}_{t}=[\mathbf{f}(\mathbf{x}_{t},t)-g^{2}(t)(\mathbf{s}_{\bm{\theta}_{\infty}}(\mathbf{x}_{t},t)+\nabla\log{p_{\bm{\psi}_{\infty}}(y\vert\mathbf{x}_{t},t)})]\diff \bar{t}+g_{t}\diff\bar{\mathbf{w}}_{t}$. This is equivalent to sampling from the joint distribution of $(\mathbf{x}_{t},y)$ because
\begin{align*}
\nabla\log{p_{r}^{t}(\mathbf{x}_{t},y)}&=\nabla\log{p_{r}^{t}(\mathbf{x}_{t})}+\nabla\log{p(y\vert\mathbf{x}_{t},t)}\\
&\approx\mathbf{s}_{\bm{\theta}_{\infty}}(\mathbf{x}_{t},t)+\nabla\log{p_{\bm{\psi}_{\infty}}(y\vert\mathbf{x}_{t},t)},
\end{align*}
where $p(y\vert\mathbf{x}_{t},t)$ is the oracle classifier at $t$. Classifier Guidance provides supervision information on a sample path, evaluating whether the sample is correctly classified by the class label $y$, or not. However, using Classifier Guidance may lead to mode collapse as it maximizes the classifier probability $p(y\vert\mathbf{x}_{t},t)$. In contrast, Discriminator Guidance offers enhanced mode coverage, as elaborated in Section \ref{sec:mode_coverage}, by providing distinctive supervision information on whether a sample path is realistic or not.

\begin{algorithm}[t]
	\centering
	\caption{Sampling with Guidance Techniques}\label{alg:sampler}
	\begin{algorithmic}[1]
		\STATE Sample $\mathbf{x}_{T}\sim\mathcal{N}(0,\sigma_{max}^{2}\mathbf{I})$
		\FOR{$i=N$ to $1$}
		\STATE Sample $\bm{\epsilon}_{i}\sim\mathcal{N}(0,S_{1}^{2}\mathbf{I})$ and $\bm{\epsilon}_{i}'\sim\mathcal{N}(0,\mathbf{I})$
		\STATE $\hat{t}_{i}\leftarrow\sigma^{-1}((1+\gamma_{t_{i}})\sigma(t_{i}))$ \cite{karras2022elucidating}
		\STATE $\mathbf{x}_{\hat{t}_{i}}\leftarrow\mathbf{x}_{t_{i}}+\sqrt{\sigma^{2}(\hat{t}_{i})-\sigma^{2}(t_{i})}\bm{\epsilon}_{t_{i}}$
		\STATE $\mathbf{s}_{\hat{t}_{i}}\leftarrow \mathbf{f}(\mathbf{x}_{\hat{t}_{i}},\hat{t}_{i})-\frac{1+\eta^{2}}{2}g^{2}_{\hat{t}_{i}}\mathbf{s}_{\bm{\theta}_{\infty}}(\mathbf{x}_{\hat{t}_{i}},\hat{t}_{i})$
		\STATE $\mathbf{c}_{\hat{t}_{i}}\leftarrow-\frac{1+\eta^{2}}{2}g^{2}_{\hat{t}_{i}}\nabla\log{\frac{d_{\bm{\phi}_{\infty}}(\nu_{\tau_{\hat{t}_{i}}}\mathbf{x}_{\hat{t}_{i}},\tau_{\hat{t}_{i}})}{1-d_{\bm{\phi}_{\infty}}(\nu_{\tau_{\hat{t}_{i}}}\mathbf{x}_{\hat{t}_{i}},\tau_{\hat{t}_{i}})}}$ (Eq. \eqref{eq:tau})
		\STATE $\mathbf{g}_{\hat{t}_{i}}\leftarrow -\frac{1+\eta^{2}}{2}g^{2}_{\hat{t}_{i}}\nabla\log{p_{\bm{\psi}_{\infty}}(y\vert\mathbf{x}_{\hat{t}_{i}},\hat{t}_{i})}$
		\STATE $\mathbf{x}_{t_{i-1}}\leftarrow\mathbf{x}_{\hat{t}_{i}}+(t_{i-1}-\hat{t}_{i})\big(\mathbf{s}_{\hat{t}_{i}}+w_{\hat{t}_{i}}^{DG}\mathbf{c}_{\hat{t}_{i}}+w_{\hat{t}_{i}}^{CG}\mathbf{g}_{\hat{t}_{i}}\big)$
		\STATE $\mathbf{x}_{t_{i-1}}\leftarrow\mathbf{x}_{t_{i-1}}+\eta g_{\hat{t}_{i}}\sqrt{t_{i-1}-\hat{t}_{i}}\bm{\epsilon}_{i}'$
		\ENDFOR
	\end{algorithmic}
\end{algorithm}

As sampling from the joint distribution of $(\mathbf{x}_{t},y)$ requires accurate score estimation, Discriminator Guidance and Classifier Guidance can be combined for a synergistic effect. We suggest the combination of guidance techniques by
\begin{align}\label{eq:dg_cg}
\begin{split}
\diff\mathbf{x}_{t}&=\Big[\mathbf{f}(\mathbf{x}_{t},t)-g^{2}(t)\big((\mathbf{s}_{\bm{\theta}_{\infty}}+w_{t}^{DG}\mathbf{c}_{\bm{\phi}_{\infty}})(\mathbf{x}_{t},t)\\
&\quad+w_{t}^{CG}\nabla\log{p_{\bm{\psi}_{\infty}}(c\vert\mathbf{x}_{t},t)}\big)\Big]\diff \bar{t}+g(t)\diff\bar{\mathbf{w}}_{t},
\end{split}
\end{align}
\begin{wraptable}{r}{0.289\textwidth}
	\vskip -0.17in
	\centering
	\caption{Algorithm \ref{alg:sampler} includes DDPM, DDIM, and EDM samplers.}
	\label{tab:sampler}
	\tiny
	\centering
	\begin{tabular}{lcccc}
		\toprule
		Sampler & $\gamma_{t}$ & $\eta$ & $w_{t}^{CG}$ & $w_{t}^{DG}$ \\\midrule
		DDPM & 0 & 1 & 0 & 0 \\
		DDIM & 0 & 0 & 0 & 0 \\
		EDM & $\ge$0 & 0 & 0 & 0 \\
		ADM-G & 0 & 1 & $>$0 & 0 \\\cdashlinelr{1-5}
		\cc{15}EDM-G++ & \cc{15}$\ge$0 & \cc{15}0 & \cc{15}0 & \cc{15}$>$0\\
		\cc{15}ADM-G++ & \cc{15}0 & \cc{15}1 & \cc{15}$>$0 & \cc{15}$>$0 \\
		\bottomrule
	\end{tabular}
	\vskip -0.1in
\end{wraptable}
where $w_{t}^{DG}$ and $w_{t}^{CG}$ are the time-dependent weights, respectively. The two pieces of information could ideally guide the sample toward the common likely region of classifier and discriminator in a complementary way. 

Algorithm \ref{alg:sampler} describes the full details of our sampling procedure for Eq. \eqref{eq:dg_cg}. The algorithm reduces the samplers of DDPM \cite{ho2020denoising, dhariwal2021diffusion}, DDIM \cite{song2020denoising}, and EDM \cite{karras2022elucidating} with corresponding hyperparameters in Table \ref{tab:sampler}. Our sampler is denoted as the postfix with \textbf{G++} upon a basic sampler by a prefix. See Appendix \ref{sec:sampling} for detailed sampling procedure.

\section{Related Works}

A line of research merges diffusion models with GAN models. \citet{zheng2022truncated, lyu2022accelerating} synthesize the diffused data $\mathbf{x}_{\sigma_{mid}}$ with a GAN generator (by putting $\mathbf{x}_{T}$ as generator's input), and denoise $\mathbf{x}_{\sigma_{mid}}$ to $\mathbf{x}_{0}$ with a diffusion model. \citet{xiao2021tackling} exchange thousands of denoising steps with a small number of sequential conditional GAN generators. \citet{wang2022diffusion} utilize the diffusion concept to train GAN. On the contrary, \citet{jolicoeur2020adversarial} train the diffusion model with an adversarial loss. The diffusion model and GAN in previous works, however, do not interplay with each other in generation process after their training. In contrast, the discriminator in DG intervenes in generation process, directly. Another difference to previous research is that we train our discriminator without any generator, so discriminator training is stable.

\begin{figure}[t]
	\centering
	\includegraphics[width=\linewidth]{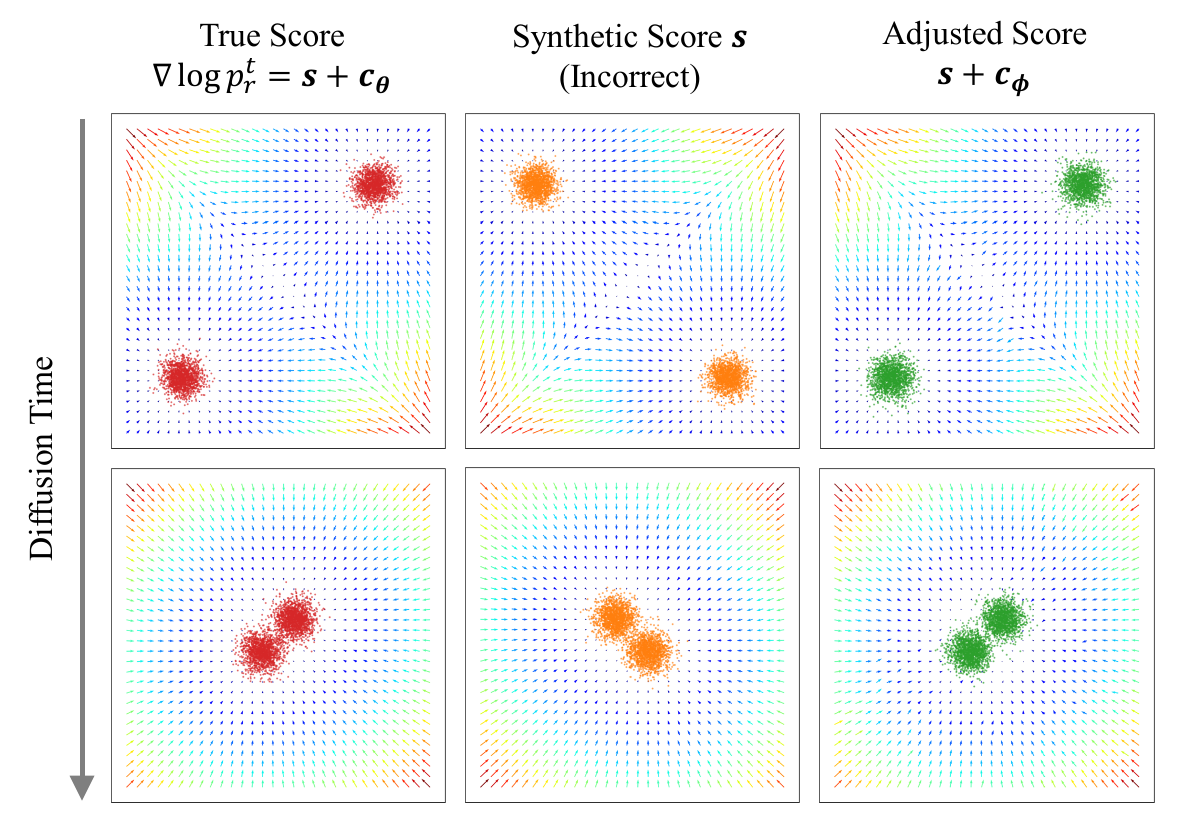}
	\caption{Comparison of the true data score, synthetic score, and adjusted score in a toy 2-dimensional case. We assume $p_{r}(\mathbf{x}):=\frac{1}{2}\mathcal{N}(\mathbf{x};(10,10)^{T},\mathbf{I})+\frac{1}{2}\mathcal{N}(\mathbf{x};(-10,-10)^{T},\mathbf{I})$ as data distribution and $p_{g}(\mathbf{x}):=\frac{1}{2}\mathcal{N}(\mathbf{x};(10,-10)^{T},\mathbf{I})+\frac{1}{2}\mathcal{N}(\mathbf{x};(-10,10)^{T},\mathbf{I})$ as hypothetially synthesized generative distribution with an incorrect score $\mathbf{s}(\mathbf{x}_{t},t):=\nabla\log{p_{g}^{t}(\mathbf{x}_{t})}$. We solve the probability-flow ODE \cite{song2020score} to visualize samples from each score function.}
	\vskip -0.1in
	\label{fig:2d_two_modal}
\end{figure}

Other previous works use the rejection sampler or MCMC to draw samples from a reweighted model distribution. \citet{azadi2018discriminator, che2020your} utilize the rejection sampler to adjust the generator implicit distribution with the discriminator under the likelihood-ratio trick \cite{gutmann2010noise}. Similarly, \citet{turner2019metropolis} make use of likelihood-ratio trick to sample with MCMC. Specifically, \citet{aneja2021contrastive} and \citet{bauer2019resampled} introduce the importance and rejection samplings from the aggregate posterior in VAE, respectively. In diffusion context, Discriminator Guidance maximizes $\frac{p_{r}^{t}}{p_{\bm{\theta}_{\infty}}^{t}}$ in the same spirit of gradient ascent, and it could be interpreted as reweighting the model distribution $p_{\bm{\theta}_{\infty}}^{t}$ with an importance weight of $\frac{p_{r}^{t}}{p_{\bm{\theta}_{\infty}}^{t}}$. Consequently, DG does not require sample rejection and is becoming scalable to high-dimensions.

\section{Experiments}

\subsection{A Toy 2-dimensional Case}

\begin{table}[t]
	\caption{Performance on CIFAR-10.}
	\vskip -0.05in
	\label{tab:cifar10}
	\tiny
	\centering
	\begin{threeparttable}
		\begin{tabular}{l@{\hskip 0.2cm}c@{\hskip 0.2cm}c@{\hskip 0.3cm}c@{\hskip 0.2cm}c@{\hskip 0.cm}c}
			\toprule
			\multirow{2}{*}{Model} & \multirow{2}{*}{\shortstack{Diffusion\\Space}} & \multirow{2}{*}{NFE$\downarrow$} & \multicolumn{2}{@{\hskip 0.1cm}c@{\hskip 0.8cm}}{Unconditional} & \multicolumn{1}{@{\hskip -0.3cm}c}{Conditional} \\
			& & & NLL$\downarrow$ & FID$\downarrow$ & FID$\downarrow$ \\\midrule
			VDM \cite{kingma2021variational} & Data & 1000 & \textbf{2.49} & 7.41 & - \\
			DDPM \cite{ho2020denoising} & Data & 1000 & 3.75 & 3.17 & - \\
			iDDPM \cite{nichol2021improved} & Data & 1000 & 3.37 & 2.90 & - \\
			Soft Truncation \cite{kim2022soft} & Data & 2000 & 2.91 & 2.47 & - \\
			INDM \cite{kim2022maximum} & Latent & 2000 & 3.09 & 2.28 & - \\
			CLD-SGM \cite{dockhorn2021score} & Data & 312 & 3.31 & 2.25 & - \\
			NCSN++ \cite{song2020score} & Data & 2000 & 3.45 & 2.20 & - \\
			LSGM \cite{vahdat2021score} & Latent & 138 & 3.43 & 2.10 & - \\
			NCSN++-G \cite{chao2022denoising} & Data & 2000 & - & - & 2.25 \\
			EDM\tnote{\textdagger}  (random seed) & Data & 39 & 2.60 & 2.03 & 1.82 \\
			EDM (reported, manual seed) & Data & \textbf{35} & 2.60 & 1.97 & 1.79 \\\cdashlinelr{1-6}
			\cc{15}LSGM-G++ & \cc{15}Latent & \cc{15}138 & \cc{15}3.42 & \cc{15}1.94 & \cc{15}- \\
			\cc{15}EDM-G++ & \cc{15}Data & \cc{15}\textbf{35} & \cc{15}2.55 & \cc{15}\textbf{1.77} & \cc{15}\textbf{1.64} \\
			\bottomrule
		\end{tabular}
		\begin{tablenotes}
			\item[\textdagger] We recalculate FID of EDM \cite{karras2022elucidating} under a random seed for a fair comparison with previous research. We report our performances under the random seed by default.
		\end{tablenotes}
	\end{threeparttable}
	\vskip -0.1in
\end{table}

\begin{table}[t]
	\caption{Performance on CelebA/FFHQ 64x64.}
	\vskip -0.05in
	\label{tab:human-face}
	\scriptsize
	\centering
	\begin{tabular}{lccc}
		\toprule
		Model & NFE$\downarrow$ & CelebA & FFHQ\\\midrule
		DDPM++ \cite{song2020score} & 131 & 2.32 & - \\
		Soft Truncation \cite{kim2022soft} & 131 & 1.90 & - \\
		Soft Diffusion \cite{daras2022soft} & 300 & 1.85 & - \\
		INDM \cite{kim2022maximum} & 132 & 1.75 & - \\
		Diffusion StyleGAN2 \cite{wang2022diffusion} & \textbf{1} & 1.69 & - \\
		EDM \cite{karras2022elucidating} & 79 & - & 2.39 \\\cdashlinelr{1-4}
		\cc{15}Soft Truncation-G++ & \cc{15}131 & \cc{15}\textbf{1.34} & \cc{15}- \\
		\cc{15}EDM-G++ & \cc{15}71 & \cc{15}- & \cc{15}\textbf{1.98} \\
		\bottomrule
	\end{tabular}
	\vskip -0.1in
\end{table}

Figure \ref{fig:2d_two_modal} shows the experimental result of a tractable 2-dimensional toy case. We train a 4-layered MLP discriminator with 256 neurons until convergence, and we hypothesize an incorrect score function $\mathbf{s}:=\nabla\log{p_{g}^{t}}$ (for a synthetic generative distribution $p_{g}$) that misfits to the data score. If there is no guidance, the incorrect score $\mathbf{s}$ will generate samples from a wrong distribution as in Figure \ref{fig:2d_two_modal}. On the contrary, $\mathbf{s}+\mathbf{c}_{\bm{\phi}_{\infty}}$ successfully guides $\mathbf{s}$ to $\nabla\log{p_{r}^{t}}$, see Figures \ref{fig:2d_two_gaussian} and \ref{fig:2d_standard_gaussian} for additional visualization. 

\subsection{Image Generation}

We experiment on CIFAR-10, CelebA/FFHQ 64x64, and ImageNet 256x256. We use the pre-trained networks on CIFAR-10 and FFHQ from \citet{karras2022elucidating, vahdat2021score}, CelebA from \citet{kim2022soft} and ImageNet from \citet{dhariwal2021diffusion, peebles2022scalable}. 

\textbf{Discriminator Network.} We use the encoder of U-Net structure\footnote{We tested MLP, ResNet18, and a transformer, but the U-Net performed the best for Discriminator Guidance.} as our discriminator network. For diffusion models on data space, we attach two noise-embedded U-Net encoders: the pre-trained ADM classifier \cite{dhariwal2021diffusion} and an auxiliary (shallow) U-Net encoder. We put $(\mathbf{x}_{t},t)$ to the ADM classifier, and we extract the latent $\mathbf{z}_{t}$ of $\mathbf{x}_{t}$ from the last pooling layer of the pre-trained classifier. Then, we put $(\mathbf{z}_{t},t)$ to the auxiliary U-Net encoder and predict real/fake by its output. We freeze the ADM classifier, and we only fine-tune shallow U-Net encoder as default. Not to mention that fine-tuning save the training cost, fine-tuning performs better or equivalent to training the entire architecture \cite{kato2021non}. For LSGM-G++, we train the U-Net encoder from scratch. For DiT-XL-G++, we train the latent classifier with the same architecture of the ADM classifier except the input dimension, and fine-tuning the shallow U-Net encoder for discriminator. We train a class-conditional discriminator for the class-conditional generation. See Table \ref{tab:configurations} for detailed training configuration.

\begin{table}[t]
	\caption{Performance on ImageNet 256x256.}
	\vskip -0.05in
	\label{tab:ImageNet256}
	\tiny
	\centering
	\begin{tabular}{l@{\hskip 0.05cm}c@{\hskip 0.09cm}c@{\hskip 0.14cm}c@{\hskip 0.13cm}c@{\hskip 0.20cm}c@{\hskip 0.14cm}c@{\hskip 0.14cm}c}
		\toprule
		\multirow{2}{*}{Model} & \multirow{2}{*}{\shortstack{Diffusion\\Space}} & \multirow{2}{*}{FID$\downarrow$} & \multirow{2}{*}{sFID$\downarrow$} & \multirow{2}{*}{IS$\uparrow$} & \multirow{2}{*}{Prec$\uparrow$} & \multirow{2}{*}{Rec$\uparrow$} & \multirow{2}{*}{F1$\uparrow$} \\
		&&&&&&&\\\midrule
		\multicolumn{2}{l}{Validation Data} & 1.68 & 3.67 & 232.21 & 0.75 & 0.66 & 0.70 \\\midrule
		ADM \cite{dhariwal2021diffusion} & Data & 10.94 & 6.02 & 100.98 & 0.69 & 0.63 & 0.66 \\
		DiT-XL/2 \cite{peebles2022scalable} & Latent & 9.62 & 6.85 & 121.50 & 0.67 & 0.67 & 0.67 \\
		ADM-G \cite{dhariwal2021diffusion} & Data & 4.59 & 5.25 & 186.70 & 0.82 & 0.52 & 0.64 \\
		RIN \cite{jabri2022scalable} & Data & 4.51 & - & 161.00 & - & - & - \\
		LDM-4-G \cite{rombach2022high} & Latent & 3.60 & - & 247.67 & \textbf{0.87} & 0.48 & 0.62 \\
		RIN + schedule \cite{chen2023importance} & Data & 3.52 & - & 186.20 & - & - & - \\
		simple diffusion \cite{hoogeboom2023simple} & Data & 2.77 & - & 211.80 & - & - & - \\
		DiT-XL/2-G \cite{peebles2022scalable} & Latent & 2.27 & 4.60 & 278.24 & 0.83 & 0.57 & 0.68 \\\cdashlinelr{1-8}
		\cc{15}ADM-G++ & \cc{15}Data & \cc{15}3.18 & \cc{15}\textbf{4.53} & \cc{15}255.74 & \cc{15}0.84 & \cc{15}0.53 & \cc{15}0.66 \\
		\cc{15}DiT-XL/2-G++ & \cc{15}Latent & \cc{15}\textbf{1.83} & \cc{15}5.16 & \cc{15}\textbf{281.53} & \cc{15}0.78 & \cc{15}\textbf{0.64} & \cc{15}\textbf{0.70} \\
		\bottomrule
	\end{tabular}
\end{table}

\begin{table}[t]
	\caption{Component-wise computational budget.}
	\vskip -0.05in
	\label{tab:budget}
	\scriptsize
	\centering
	\begin{tabular}{lccc}
		\toprule
		\multirow{2}{*}{\shortstack[l]{Dataset (Model)}} & \multirow{2}{*}{\shortstack{Score\\Training}} & \multirow{2}{*}{\shortstack{Sample\\Generation}} & \multirow{2}{*}{\shortstack{Discriminator\\Training}} \\
		&&&\\\midrule
		\multirow{2}{*}{CIFAR-10 (EDM)} & \multirow{2}{*}{\shortstack{200M $\mathbf{s}_{\bm{\theta}_{\infty}}$\\(480hr)}} & \multirow{2}{*}{\shortstack{1.75M $\mathbf{s}_{\bm{\theta}_{\infty}}$\\(1hr)}} & \multirow{2}{*}{\shortstack{1M $d_{\bm{\phi}}$\\(10min)}} \\
		&&&\\\cmidrule(lr){1-1}
		ImageNet 256x256 (ADM) & 2.5T $\mathbf{s}_{\bm{\theta}_{\infty}}$ & 100M $\mathbf{s}_{\bm{\theta}_{\infty}}$ & 25.6M $d_{\bm{\phi}}$ \\
		\bottomrule
	\end{tabular}
\end{table}

\begin{figure}[t]
	\centering
	\begin{subfigure}{0.3\linewidth}
		\centering
		\includegraphics[width=\linewidth]{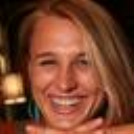}
		\subcaption{Original Data}
	\end{subfigure}
	\hfil
	\begin{subfigure}{0.3\linewidth}
		\centering
		\includegraphics[width=\linewidth]{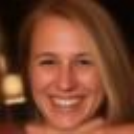}
		\subcaption{EDM}
	\end{subfigure}		
	\hfil
	\begin{subfigure}{0.3\linewidth}
		\centering
		\includegraphics[width=\linewidth]{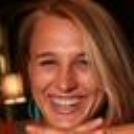}
		\subcaption{EDM-G++}
	\end{subfigure}		
	\caption{Comparison of (a) original sample, and the average of 100 regenerated samples of (b) EDM and (c) EDM-G++ on FFHQ. The regeneration FID is 1.25 (w/o DG) and 1.09 (w/ DG) if we perturb data by the standard Gaussian.}
	\label{fig:comparison_sample}
\end{figure}

\textbf{Quantitative Analysis.} We achieve new SOTA FIDs on all datasets including CIFAR-10, CelebA, FFHQ, and ImageNet. On CIFAR-10, Table \ref{tab:cifar10} shows that Discriminator Guidance is effective in both data diffusion (EDM) and latent diffusion (LSGM) models. Other than Discriminator Guidance, we use all the hyperparameters of EDM and LSGM, so the performance gain purely comes from the discriminator componenet. The gain of Discriminator Guidance is also notable on human-face datasets in Table \ref{tab:human-face}.

On ImageNet 256x256, we present SOTA results in various metrics, including FID, sFID, IS, and recall in Table \ref{tab:ImageNet256}. For reference, we also measure these metrics on the ImageNet 50k validation data. Notably, IS and precision of the validation data are on par with the best models. Thus, optimizing other metrics, such as FID, sFID, and recall, becomes more important once IS and precision of a model reach the level of the validation data. Our experiments show that with Discriminator Guidance, we achieve the strongest performances with respect to FID and recall in DiT-XL/2-G++, indicating that the sample quality and diversity are significantly improved. See Table \ref{tab:configurations} for detailed hyperparameters and Appendix \ref{sec:uncurated} for uncurated samples.

In terms of carbon footprint, Discriminator Guidance requires additional sampling from the pre-trained score model plus discriminator training. Table \ref{tab:budget} summarizes the component-wise neural network evaluation budget. In both CIFAR-10 and ImageNet experiments, Discriminator Guidance requires a lightweight burden compared to score training. Furthermore, we measure the actual elapsed time in GPU (A100) hours, including backpropagation time.

\textbf{Qualitative Analysis.} Figure \ref{fig:comparison_sample} illustrates a comparison between the original sample and the averages of its regenerated samples \cite{meng2021sdedit} on FFHQ. If the score estimation is correct, the average reconstructed image should be approximately equal to the original image when the perturbation noise is small enough. To explain this, suppose the original image is $\mathbf{y}$ and we perturb it with a fixed direction $\bm{\epsilon}$ by $\mathbf{y}+\sigma(t)\bm{\epsilon}$. Then, by putting this perturbed data into the Tweedie's formula \cite{robbins1992empirical, jolicoeur2020adversarial}, the average reconstructed data $\mathbf{x}_{0}$ becomes
\begin{align*}
&\mathbb{E}[\mathbf{x}_{0}\vert\mathbf{x}_{t}=\mathbf{y}+\sigma(t)\bm{\epsilon}]\\
&\quad\quad\quad=\mathbf{y}+\sigma(t)\bm{\epsilon}+\sigma^{2}(t)\nabla\log{p_{r}^{t}\big(\mathbf{y}+\sigma(t)\bm{\epsilon}\big)}.
\end{align*} 

If $\sigma(t)$ is sufficiently small, we obtain $p_{r}^{t}(\mathbf{y}+\sigma(t)\bm{\epsilon})\appropto p(\bm{\epsilon})$, which leads to $\mathbb{E}_{\bm{\epsilon}}[\nabla\log{p_{r}^{t}(\mathbf{y}+\sigma(t)\bm{\epsilon})}]\approx 0$. Therefore, the average reconstructed image of $\mathbf{y}$ with a random direction $\bm{\epsilon}$ would approximately be the original data,
\begin{align*}
\mathbb{E}_{\bm{\epsilon}}\big[\mathbb{E}[\mathbf{x}_{0}\vert\mathbf{x}_{t}=\mathbf{y}+\sigma(t)\bm{\epsilon}]\big]\approx\mathbf{y}.
\end{align*}
In conclusion, the closeness of the average reconstructed image to the original one indirectly diagnose whether the estimated score is accurate because the above argument holds for data score. Figure \ref{fig:comparison_sample}, with a small $\sigma(t)=1$, suggests that the adjusted score provides a more accurate estimation compared to the original model score.

\begin{figure}[t]
	\centering
	\includegraphics[width=0.9\linewidth]{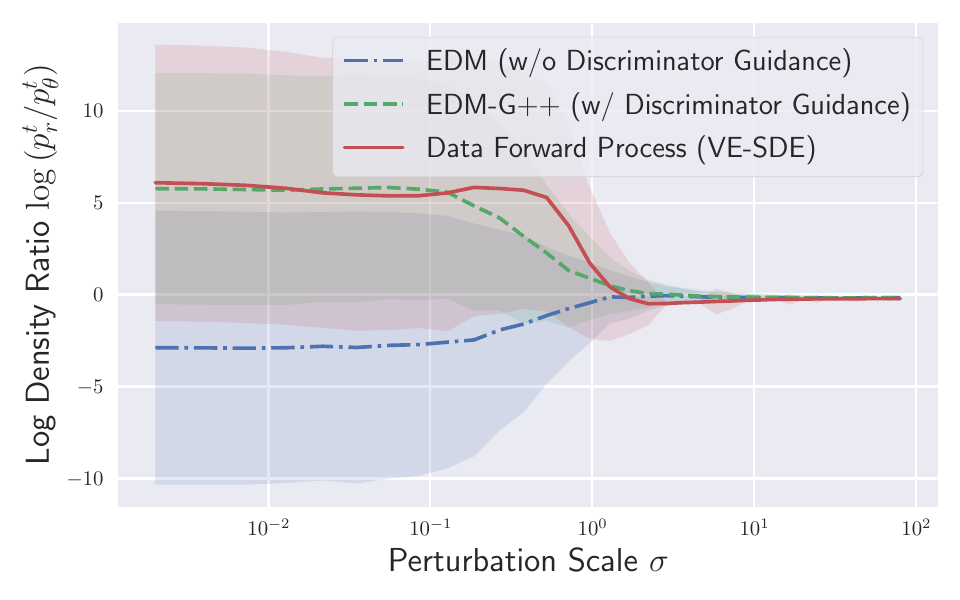}
	\vskip -0.1in
	\caption{Comparison of sample trajectories with respect to the density-ratio on FFHQ.}
	\label{fig:log_density_ratio}
\end{figure}

\begin{figure}[t]
	\vskip -0.1in
	\centering
	\includegraphics[width=0.9\linewidth]{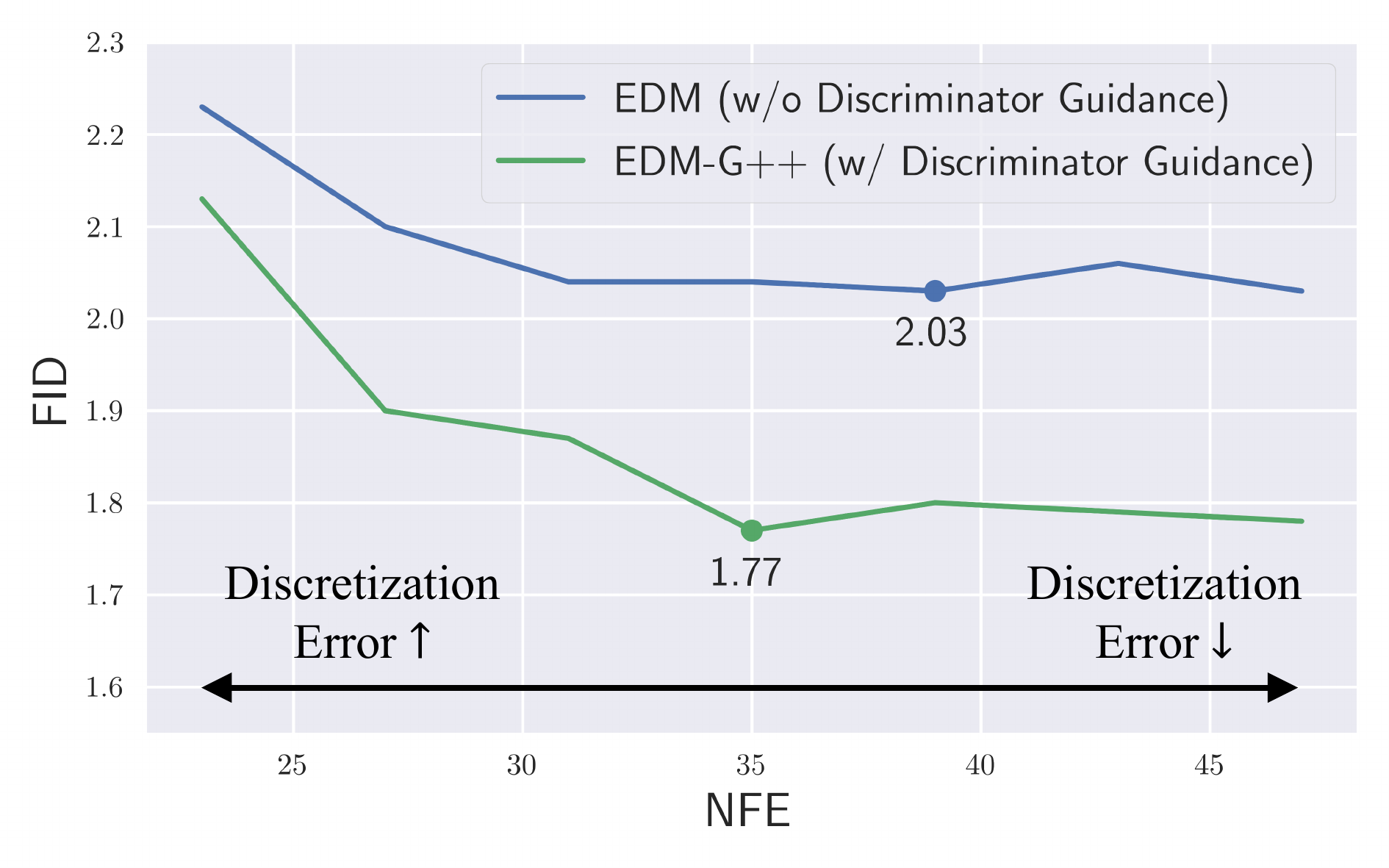}
	\vskip -0.1in
	\caption{Ablation study for NFE on CIFAR-10.}
	\label{fig:NFE}
	\vskip -0.1in
\end{figure}

\begin{figure}[t]
	\centering
	\includegraphics[width=0.9\linewidth]{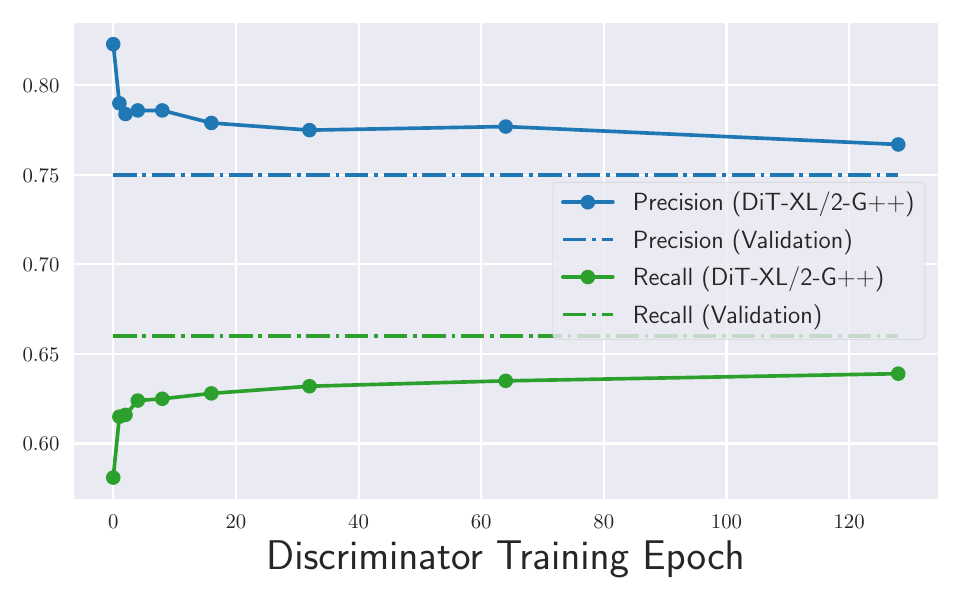}
	\vskip -0.1in
	\caption{Precision/recall by discriminator training.}
	\label{fig:sample_diversity}
\end{figure}

\begin{figure}[t]
	\vskip -0.1in
	\centering
	\includegraphics[width=0.9\linewidth]{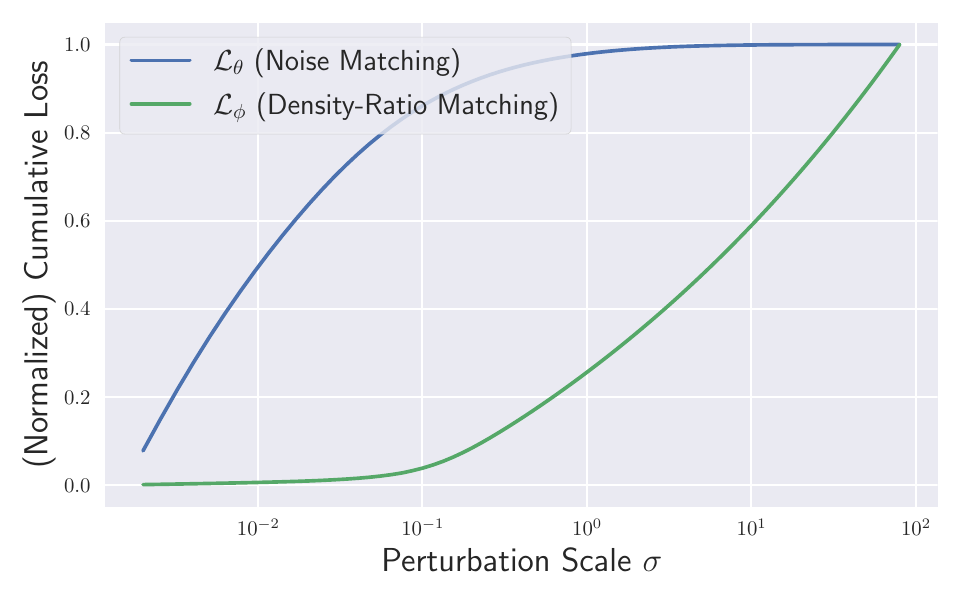}
	\vskip -0.1in
	\caption{Loss contribution by noise scale.}
	\label{fig:loss_contribution}
	\vskip -0.1in
\end{figure}

Figure \ref{fig:log_density_ratio} shows that the trained discriminator is able to accurately distinguish between the diffusion path of real data and the denoising path of the generated sample. In contrast, the discriminator-adjusted denoising path deceives the discriminator, resulting in the density-ratio curve of the adjusted generative SDE (Eq. \ref{eq:dg}) closely approximating that of the data forward SDE.

Figure \ref{fig:NFE} illustrates the effect of Discriminator Guidance with respect to the sampling NFEs on CIFAR-10. As NFE decreases, the discretization error dominates the sampling error \cite{de2022convergence}, and the gain from Discriminator Guidance becomes suboptimal. We leave it as a future work to fit Discriminator Guidance on samplers with extremely small NFEs. See Appendix \ref{sec:ablation} for more ablation studies.

Figure \ref{fig:sample_diversity} shows the precision and recall curve by discriminator training. At the zero-th epoch, before discriminator training, we observe that precision/recall for vanilla DiT-XL-G are higher/lower than those of the validation data, respectively. This is because Classifier Guidance generates overconfident samples in terms of the classifier. However, Discriminator Guidance significantly mitigates this precision-recall trade-off of Classifier Guidance. 

Figure \ref{fig:loss_contribution} displays the normalized cumulative objective loss by noise scale. To ensure a fair comparison, we utilize the same weighting function $\xi(t)=\lambda(t)$ to evaluate the discriminator loss $\mathcal{L}_{\bm{\phi}}$ and the score loss $\mathcal{L}_{\bm{\theta}}$. The results reveal that the discriminator can capture the estimation error, particularly at a large diffusion time that determines the sample diversity. This finding highlights the potential of Discriminator Guidance as a supplementary approach to address the problem of poor estimation at large time \cite{kim2022soft} in the score matching framework.

\begin{figure*}[t]
	\centering
	\begin{subfigure}{0.61\linewidth}
		\centering
		\includegraphics[width=\linewidth]{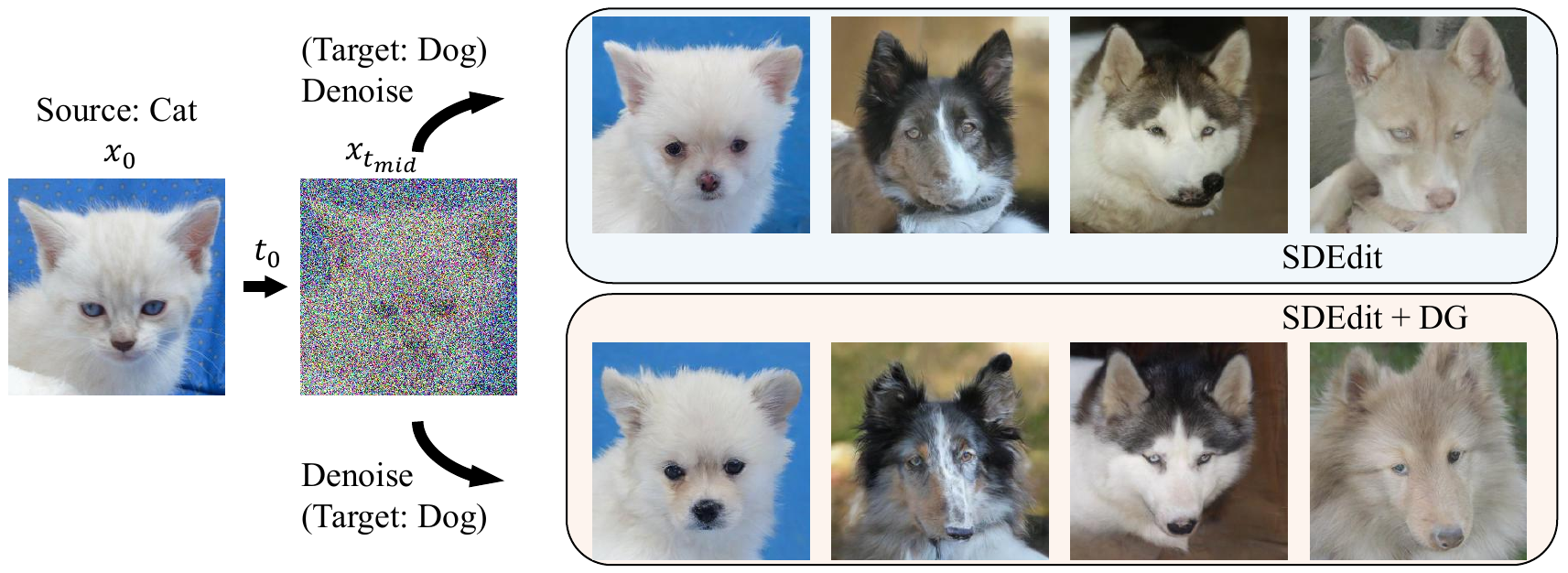}
		\subcaption{SDEdit+DG improves translation quality}
	\end{subfigure}
	\hfil
	\begin{subfigure}{0.36\linewidth}
		\centering
		\includegraphics[width=\linewidth]{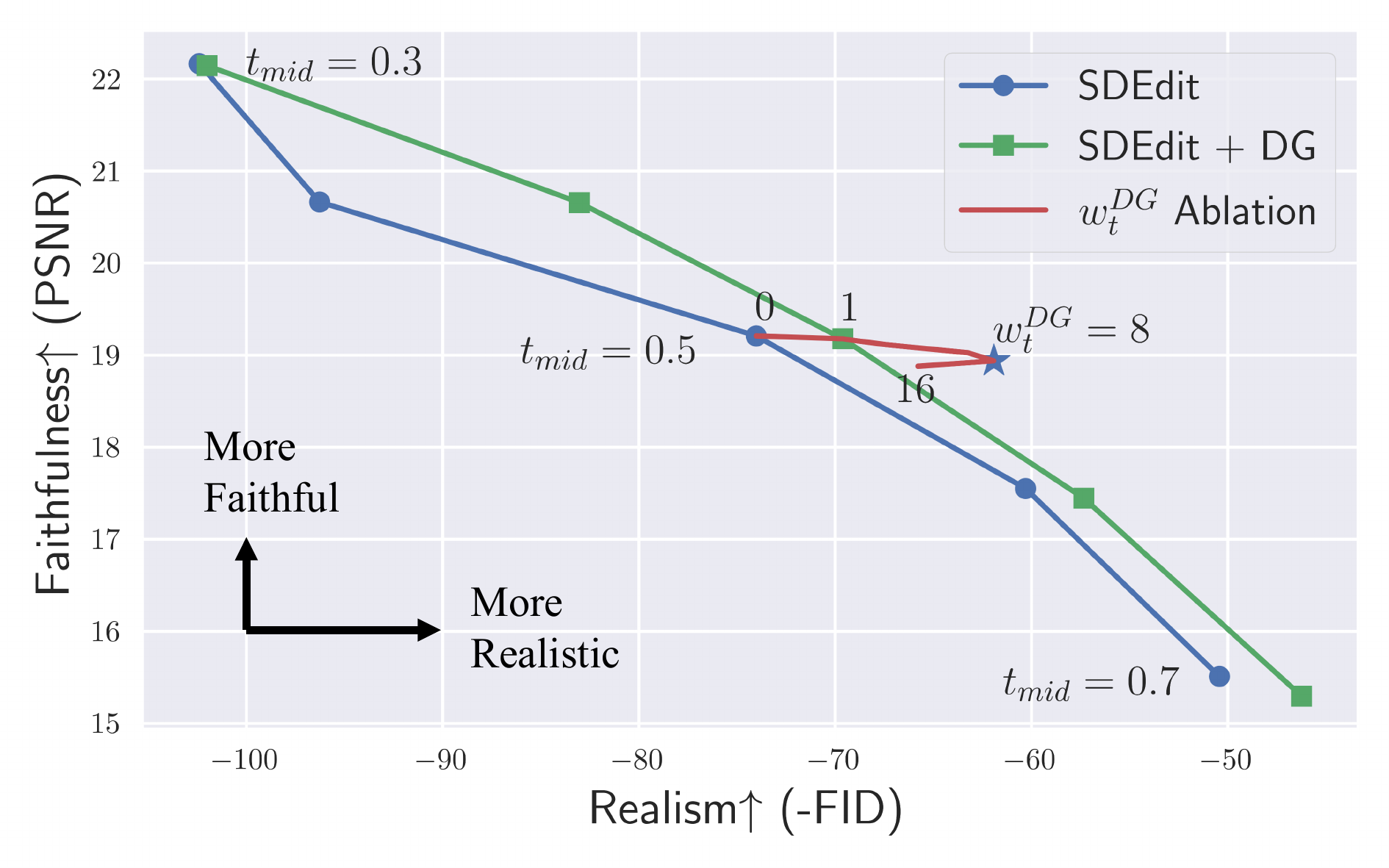}
		\subcaption{Realism vs. Faithfulness}
	\end{subfigure}
	\caption{I2I translation task. (a) The discriminator-guided translated samples from SDEdit are more realistic than non-guided samples from SDEdit given the same translation starting time $t_{mid}=0.5$. (b) DG partially mitigates the trade-off between realism and faithfulness. Also, the weighted DG improves sample realism without hurting faithfulness.}
	\label{fig:I2I}
\end{figure*}

\subsection{Image-2-Image Translation}

Discriminator Guidance could be applied to the Image-2-Image (I2I) translation task. I2I \cite{meng2021sdedit} denoise a perturbed source image with a score network trained on the target domain. For discriminator training, we first translate source training images, and then we aggregate these translated images with source images to a fake dataset $\mathcal{G}$. By specifying $\mathcal{D}$ as target images, we follow Algorithm \ref{alg:discriminator} to train the discriminator. Discriminator Guidance avoids going neither the \textit{source} domain nor the \textit{translated} domain, and it leads to the \textit{target} domain. Empirically, the curve in Figure \ref{fig:I2I}-(b) shows that our approach remedies the trade-off between \textit{realism} (to target) and \textit{faithfulness} (with source). We leave Appendix \ref{sec:I2I_detail} for details.

\section{Discussion}\label{sec:discussion}

\begin{table}[t]
	\vskip -0.1in
	\centering
	\caption{Ablation study for $h$-function on CIFAR-10.}
	\label{tab:bregman}
	\scriptsize
	\centering
	\begin{tabular}{lcc}
		\toprule
		\multirow{2}{*}{Model} & \multirow{2}{*}{\shortstack{Density-Ratio Matching\\$D_{h}(r_{\bm{\theta}_{\infty}}\Vert r_{\bm{\phi}})$}} & \multirow{2}{*}{FID} \\
		&&\\\midrule
		EDM & - & 2.03 \\\cdashlinelr{1-3}
		\multirow{5}{*}[-.4em]{EDM-G++} & \multirow{2}{*}{\shortstack{UKL \cite{nguyen2010estimating}\\$h(r)=r\log{r}-r$}} & \multirow{2}{*}{1.84} \\
		&&\\\cdashlinelr{2-3}
		& \multirow{2}{*}{\shortstack{LSIF \cite{kanamori2009least}\\$h(r)=(r-1)^{2}/2$}} & \multirow{2}{*}{1.84} \\
		&&\\\cdashlinelr{2-3}
		& \multirow{2}{*}{\shortstack{BCE ($\mathcal{L}_{\bm{\phi}}$ of Eq. \ref{eq:discriminator_loss})\\$h(r)=r\log{r}-(r+1)\log{(1+r)}$}} & \multirow{2}{*}{1.77} \\
		&&\\
		\bottomrule
	\end{tabular}
	\vskip -0.05in
\end{table}

This section discusses two possible avenues for the future development of Discriminator Guidance. The first direction involves rewriting our approach in terms of Bregman divergence, while the second direction explores the simultaneous training of score and discriminator networks. In the first direction, the density-ratio $r_{\bm{\theta}_{\infty}}^{t}=\frac{p_{r}^{t}}{p_{\bm{\theta}_{\infty}}^{t}}$ is the target of the discriminator, and BCE loss $\mathcal{L}_{\bm{\phi}}$ can be generalized into the family of $h$-Bregman divergence \cite{sugiyama2012density}
	\begin{align*}
	D_{h}(r_{\bm{\theta}_{\infty}}\Vert r_{\bm{\phi}})&=\int \lambda(t) \mathbb{E}_{p_{\bm{\theta}_{\infty}}^{t}(\mathbf{x}_{t})}\big[h(r_{\bm{\theta}_{\infty}}^{t}(\mathbf{x}_{t}))-h(r_{\bm{\phi}}^{t}(\mathbf{x}_{t}))\\
	&\quad-\partial h(r_{\bm{\phi}}^{t}(\mathbf{x}_{t}))(r_{\bm{\theta}_{\infty}}^{t}(\mathbf{x}_{t})-r_{\bm{\phi}}^{t}(\mathbf{x}_{t}))\big]\diff t,
	\end{align*}
	where $r_{\bm{\phi}}=\frac{d_{\bm{\phi}}}{1-d_{\bm{\phi}}}$. It is worth noting that the BCE loss is the unique divergence that belongs to both the $h$-Bregman divergence and the $f$-divergence \cite{amari2009alpha}. Therefore, our approach suggests a new divergence family for score estimation, see Appendix \ref{sec:Bregman} for more discussion. Table \ref{tab:bregman} presents an experiment on Bregmena divergence.

\begin{figure}[t]
	\vskip -0.1in
	\centering
	\includegraphics[width=0.9\linewidth]{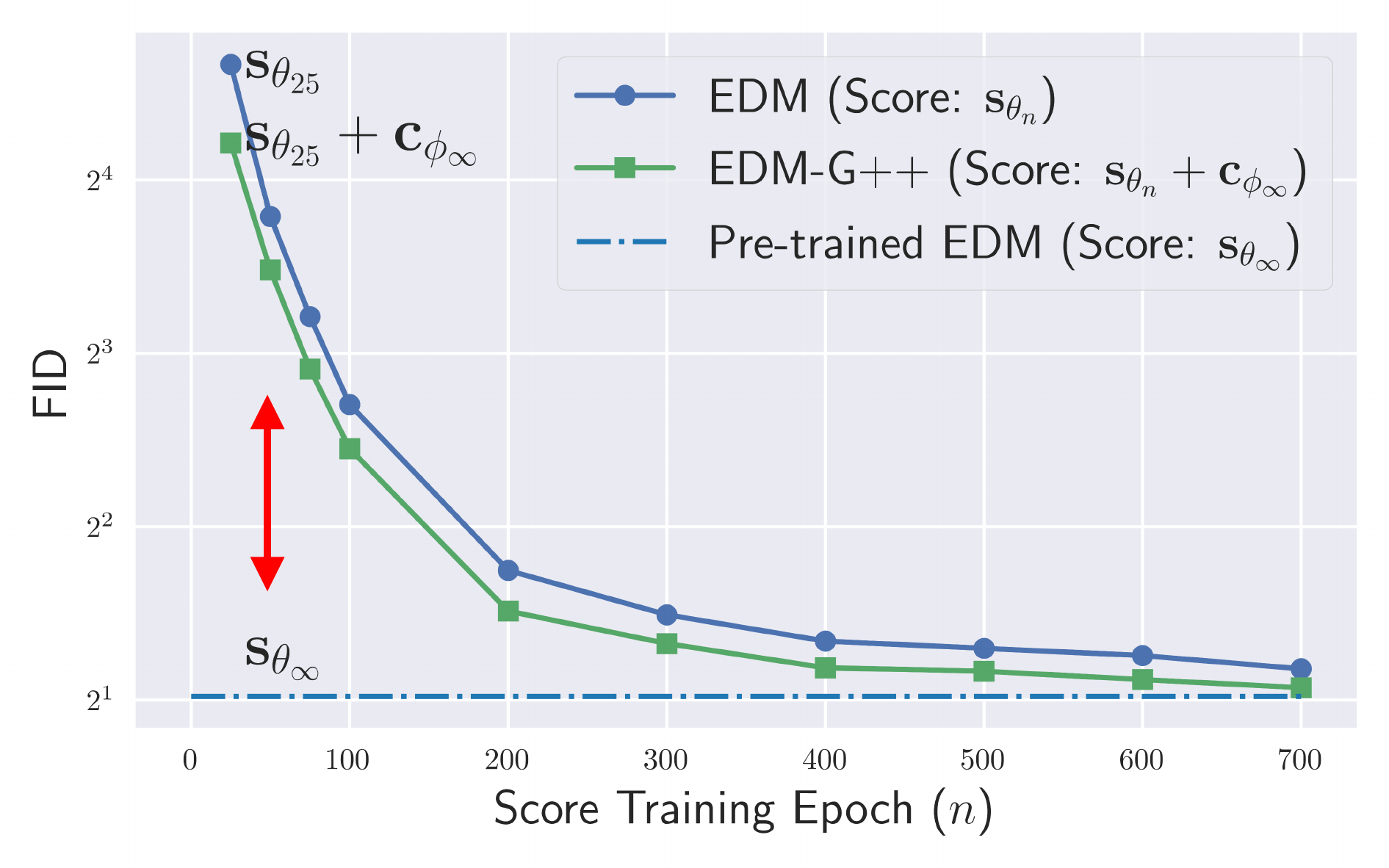}
	\vskip -0.1in
	\caption{FID of simultaneous training on CIFAR-10.}
	\label{fig:score_training}
	\vskip -0.1in
\end{figure}

Another potential direction is the simultaneous training of discriminator and score networks, which could be more appealing than GAN as it is a min-min problem instead of a mini-max GAN problem. However, the loss functions of $\mathcal{L}_{\bm{\theta}}$ and $\mathcal{L}_{\bm{\phi}}$ are independent, so their joint interplay would be restricted. The discriminator only marginally enhances the score accuracy, as shown in Figure \ref{fig:score_training}.

As an alternative, we could modify the score loss to an $f$-divergence \cite{song2021maximum}
\begin{align*}
D_{f}(p_{r}(\mathbf{x}_{0})\Vert p_{\bm{\theta}}(\mathbf{x}_{0}))=D_{f}(p_{r}^{T}(\mathbf{x}_{T})\Vert \pi(\mathbf{x}_{T}))+E_{\bm{\theta},\bm{\phi}}^{f},
\end{align*}
where $E_{\bm{\theta},\bm{\phi}}^{f}=\frac{1}{2}\int_{0}^{T}g^{2}(t)\mathbb{E}\big[ f''\big(\frac{p_{r}^{t}(\mathbf{x}_{t})}{p_{\bm{\theta}}^{t}(\mathbf{x}_{t})}\big)\frac{p_{r}^{t}(\mathbf{x}_{t})}{p_{\bm{\theta}}^{t}(\mathbf{x}_{t})}\Vert\nabla\log{p_{r}^{t}(\mathbf{x}_{t})}-\mathbf{s}_{\bm{\theta}}(\mathbf{x}_{t},t)\Vert_{2}^{2} \big]\diff t$ given assumptions of Theorem \ref{thm:1}. This $f$-divergence estabilishes a connection between discriminator and score losses when using a neural ratio to approximate the true likelihood-ratio. The $f$-divergence allows better score estimation in perceptually plausible region by weighting the score matching more on the spatial domain with high density-ratio $\frac{p_{r}^{t}}{p_{\bm{\theta}}^{t}}$, which could be an advantage over KL divergence training. We leave it as future work.

\section{Conclusion}

This paper refines the denoising process with an adjusted score estimation. With the proposed method, we could further optimize the divergence of a pre-trained score model. Empirical results demonstrate that our approach achieves new SOTA FIDs on all datasets. The deepfake images are one of the potential risks of the negative usage of this work.

\section*{Acknowledgements}

This work was supported by Institute of Information $\&$ communications Technology Planning $\&$ Evaluation (IITP) grant funded by the Korea government(MSIT) (NO. 2022-0-00077, AI Technology Development for Commonsense Extraction, Reasoning, and Inference from Heterogeneous Data). Also, this work was supported by the National Research Foundation of Korea (NRF) grant funded by the Korea government(MSIT) (NRF-2019R1A5A1028324). We thank to Jiaming Song for his sincere advice to our work. We also thank to Byunghu Na and Seungjae Shin for their warm and keen feedbacks on our manuscript.

\bibliography{references}
\bibliographystyle{icml2023}

\newpage
\appendix
\onecolumn
\section{Proofs and More Analysis}\label{sec:proofs}

\subsection{Proof of Theorem \ref{thm:1}}

Throughout the section, we assume that the assumptions made in Appendix A of \citet{song2021maximum} hold.

\begingroup
\renewcommand\thetheorem{1}
\begin{theorem}
	Suppose $p_{\bm{\theta}}$ be the solution of the time-reversal generative process of Eq. \eqref{eq:generative_process}. Let $p_{r}^{t}$ and $p_{\bm{\theta}}^{t}$ be the marginal densities (at $t$) of the forward-time SDE $\diff\mathbf{x}_{t}=\mathbf{f}(\mathbf{x}_{t},t)\diff t+g(t)\diff \mathbf{w}_{t}$ starting from $p_{r}$ and $p_{\bm{\theta}}$, respectively. If $\mathbf{s}_{\bm{\theta}}(\mathbf{x},T)=\nabla\log{\pi(\mathbf{x})}$, where $\pi$ is the prior distribution, and the log-likelihood $\log{p_{\bm{\theta}}}$ equals its evidence lower bound $\mathcal{L}_{\bm{\theta}}$, then
	\begin{align*}
	\diff\mathbf{x}_{t}=\big[\mathbf{f}(\mathbf{x}_{t},t)-g^{2}(t)\nabla\log{p_{r}^{t}(\mathbf{x}_{t})}\big]\diff \bar{t}+g(t)\diff\bar{\mathbf{w}}_{t},
	\end{align*}
	coincides with a diffusion process with adjusted score,
	\begin{align*}
	\diff\mathbf{x}_{t}&=\big[\mathbf{f}(\mathbf{x}_{t},t)-g^{2}(t)(\mathbf{s}_{\bm{\theta}}+\mathbf{c}_{\bm{\theta}})(\mathbf{x}_{t},t)\big]\diff\bar{t}+g(t)\diff\bar{\mathbf{w}}_{t},
	\end{align*}
	for $\mathbf{c}_{\bm{\theta}}(\mathbf{x}_{t},t):=\nabla\log{\frac{p_{r}^{t}(\mathbf{x}_{t})}{p_{\bm{\theta}}^{t}(\mathbf{x}_{t})}}$.
\end{theorem}
\endgroup

To prove the theorems, we define a family of rotation-free score functions by $\mathbf{S}_{sol}$ in Definition \ref{def:1}.

\begin{definition}[Definition 1 of \citet{kim2022maximum}]\label{def:1}
	Let $\mathbf{S}_{div}=\{\mathbf{v}:\mathbb{R}^{d}\rightarrow\mathbb{R}^{d}\vert\mathbf{v}=\nabla\log{p}\text{ for some probability }p\}$ be the family of rotation-free score functions. Define $\mathbf{S}_{sol}$ be a family of time-conditioned score network $\mathbf{s}$ that satisfies the following: there exists $q_{0}$ and $q_{t}$ such that $\mathbf{s}(\mathbf{x},t)=\nabla\log{q_{t}(\mathbf{x})}$ almost everywhere, where $q_{t}$ is the marginal density at time $t$ of $\diff\mathbf{x}_{t}=\mathbf{f}(\mathbf{x}_{t},t)\diff t+g(t)\diff\mathbf{w}_{t}$, starting from $\mathbf{x}_{0}\sim q_{0}$.
\end{definition}

With Definition \ref{def:1}, \citet{kim2022maximum} introduces the necessary and sufficient condition for $\mathbf{s}_{\bm{\theta}}\in\mathbf{S}_{sol}$.

\begin{lemma}[Theorem 2 of \citet{kim2022maximum}]\label{lemma:kim}
	Suppose $\mathbf{s}_{\bm{\theta}}$ is twice continuously differentiable with respect to $\mathbf{x}$. Then,
	\begin{align*}
	D_{KL}(p_{r}\Vert p_{\bm{\theta}})=\frac{1}{2}\int_{0}^{T}g^{2}(t)\mathbb{E}\big[\Vert\mathbf{s}_{\bm{\theta}}(\mathbf{x}_{t},t)-\nabla\log{p_{r}^{t}(\mathbf{x}_{t})}\Vert_{2}^{2}\big]\diff t+D_{KL}(p_{r}^{T}\Vert\pi)
	\end{align*}
	holds if and only if $\mathbf{s}_{\bm{\theta}}\in\mathbf{S}_{sol}$.
\end{lemma}

\begin{lemma}\label{lemma:iff}
	The log-likelihood equals the evidence lower bound if and only if
	\begin{align*}
	D_{KL}(p_{r}\Vert p_{\bm{\theta}})=\frac{1}{2}\int_{0}^{T}g^{2}(t)\mathbb{E}\big[\Vert\mathbf{s}_{\bm{\theta}}(\mathbf{x}_{t},t)-\nabla\log{p_{r}^{t}(\mathbf{x}_{t})}\Vert_{2}^{2}\big]\diff t+D_{KL}(p_{r}^{T}\Vert\pi).
	\end{align*}
\end{lemma}

Combining Lemmas \ref{lemma:kim} and \ref{lemma:iff}, we yield that $\mathbf{s}_{\bm{\theta}}\in\mathbf{S}_{sol}$ if and only if the log-likelihood equals the evidence lower bound. Now, we provide the proof.

\begin{proof}[Proof of Theorem \ref{thm:1}]
	From Lemmas \ref{lemma:kim} and \ref{lemma:iff} combined, we obtain $\mathbf{s}_{\bm{\theta}}\in\mathbf{S}_{sol}$. Therefore, there exists $q_{0}$ such that $q_{t}$ is the marginal density of $\diff\mathbf{x}_{t}=\mathbf{f}(\mathbf{x}_{t},t)\diff t+g(t)\diff\mathbf{w}_{t}$ and satisfies $\mathbf{s}_{\bm{\theta}}(\mathbf{x}_{t},t)=\nabla\log{q_{t}(\mathbf{x}_{t})}$ almost everywhere. Then, the solution of the generative process
	\begin{align}\label{eq:generative_process_appendix}
	\begin{split}
	\diff\mathbf{x}_{t}&=\big[\mathbf{f}(\mathbf{x}_{t},t)-g^{2}(t)\mathbf{s}_{\bm{\theta}}(\mathbf{x}_{t},t)\big]\diff \bar{t}+g(t)\diff\bar{\mathbf{w}}_{t}\\
	&=\big[\mathbf{f}(\mathbf{x}_{t},t)-g^{2}(t)\nabla\log{q_{t}(\mathbf{x}_{t})}\big]\diff \bar{t}+g(t)\diff\bar{\mathbf{w}}_{t},
	\end{split}
	\end{align}
	starting from $q_{T}$ at $T$ attains $q_{t}$ as its marginal density at $t$. As $\mathbf{s}_{\bm{\theta}}(\mathbf{x}_{T},T)=\nabla\log{\pi(\mathbf{x}_{T})}$, the marginal density $q_{T}$ at $T$ becomes the prior distribution $\pi$, and we get $q_{0}=p_{\bm{\theta}}$.
	
	Hence, $p_{\bm{\theta}}^{t}$ is the marginal density of $\diff\mathbf{x}_{t}=\mathbf{f}(\mathbf{x}_{t},t)\diff t+g(t)\diff\mathbf{w}_{t}$, starting from $\mathbf{x}_{0}\sim p_{\bm{\theta}}=q_{0}$. From the uniqueness, we conclude that $p_{\bm{\theta}}^{t}=q_{t}$ for all $t$ and thus $\mathbf{s}_{\bm{\theta}}(\mathbf{x}_{t},t)=\nabla\log{p_{\bm{\theta}}^{t}(\mathbf{x}_{t})}$, which leads the desired result.
\end{proof}

\subsection{Proof of Theorem \ref{thm:2}}

\begingroup
\renewcommand\thetheorem{2}
\begin{theorem}
	If the assumptions of Theorem \ref{thm:1} hold, then
	\begin{align*}
	&D_{KL}(p_{r}\Vert p_{\bm{\theta}})=D_{KL}(p_{r}^{T}\Vert \pi)+E_{\bm{\theta}},\\
	&D_{KL}(p_{r}\Vert p_{\bm{\theta},\bm{\phi}})\le D_{KL}(p_{r}^{T}\Vert\pi)+E_{\bm{\theta},\bm{\phi}},
	\end{align*}
	where 
	\begin{align*}
	&E_{\bm{\theta}}=\frac{1}{2}\int_{0}^{T}g^{2}(t)\mathbb{E}_{p_{r}^{t}}\big[\Vert\nabla\log{p_{r}^{t}}-\nabla\log{p_{\bm{\theta}}^{t}}\Vert_{2}^{2}\big],\\
	&E_{\bm{\theta},\bm{\phi}}=\frac{1}{2}\int_{0}^{T}g^{2}(t)\mathbb{E}_{p_{r}^{t}}\big[\Vert \mathbf{c}_{\bm{\theta}}-\mathbf{c}_{\bm{\phi}}\Vert_{2}^{2}\big]\diff t.
	\end{align*}
\end{theorem}
\endgroup

\begin{proof}[Proof of Theorem \ref{thm:2}]
	We have
	\begin{align*}
	\diff\mathbf{x}_{t}&=\big[\mathbf{f}(\mathbf{x}_{t},t)-g^{2}(t)(\mathbf{s}_{\bm{\theta}}+\mathbf{c}_{\bm{\phi}})\big]\diff \bar{t}+g(t)\diff\bar{\mathbf{w}}_{t}\\	
	&=\Big[\mathbf{f}(\mathbf{x}_{t},t)-g^{2}(t)\Big(\mathbf{s}_{\bm{\theta}}+\nabla\log{\frac{d_{\bm{\phi}}}{1-d_{\bm{\phi}}}}\Big)\Big]\diff \bar{t}+g(t)\diff\bar{\mathbf{w}}_{t}\\
	&=\Big[\mathbf{f}(\mathbf{x}_{t},t)-g^{2}(t)\Big(\nabla\log{p_{g}^{t}}+\nabla\log{\frac{d_{\bm{\phi}}}{1-d_{\bm{\phi}}}}\Big)\Big]\diff \bar{t}+g(t)\diff\bar{\mathbf{w}}_{t}\\
	&=\bigg[\mathbf{f}(\mathbf{x}_{t},t)-g^{2}(t)\Big(\nabla\log{p_{g}^{t}}+\nabla\log{\frac{d_{\bm{\phi}_{*}}}{1-d_{\bm{\phi}_{*}}}}-\nabla\log{\frac{d_{\bm{\phi}_{*}}}{1-d_{\bm{\phi}_{*}}}}+\nabla\log{\frac{d_{\bm{\phi}}}{1-d_{\bm{\phi}}}}\Big)\bigg]\diff \bar{t}+g(t)\diff\bar{\mathbf{w}}_{t}\\
	&=\bigg[\mathbf{f}(\mathbf{x}_{t},t)-g^{2}(t)\Big(\nabla\log{p_{g}^{t}}+\nabla\log{\frac{p_{r}^{t}}{p_{g}^{t}}}-\nabla\log{\frac{d_{\bm{\phi}_{*}}}{1-d_{\bm{\phi}_{*}}}}+\nabla\log{\frac{d_{\bm{\phi}}}{1-d_{\bm{\phi}}}}\Big)\bigg]\diff \bar{t}+g(t)\diff\bar{\mathbf{w}}_{t}\\
	&=\bigg[\mathbf{f}(\mathbf{x}_{t},t)-g^{2}(t)\Big(\nabla\log{p_{r}^{t}}-\nabla\log{\frac{d_{\bm{\phi}_{*}}}{1-d_{\bm{\phi}_{*}}}}+\nabla\log{\frac{d_{\bm{\phi}}}{1-d_{\bm{\phi}}}}\Big)\bigg]\diff \bar{t}+g(t)\diff\bar{\mathbf{w}}_{t}.
	\end{align*}
	Applying the Girsanov theorem to this generative SDE with the reverse-time data SDE with the data-processing inequality, we get
	\begin{align*}
	D_{KL}(p_{r}\Vert p_{\bm{\theta},\bm{\phi}})\le D_{KL}(p_{r}^{T}\Vert \pi)+\frac{1}{2}\int_{0}^{T}g^{2}(t)\mathbb{E}_{p_{r}^{t}}\bigg[\Big\Vert\nabla\log{\frac{d_{\bm{\phi}_{*}}}{1-d_{\bm{\phi}_{*}}}}-\nabla\log{\frac{d_{\bm{\phi}}}{1-d_{\bm{\phi}}}}\Big\Vert_{2}^{2}\bigg]\diff t.
	\end{align*}
	Also, the equality of $D_{KL}(p_{r}\Vert p_{\bm{\theta}})=D_{KL}(p_{r}^{T}\Vert\pi)+\frac{1}{2}\int_{0}^{T}g^{2}(t)\mathbb{E}_{p_{r}^{t}}\big[\Vert\nabla\log{p_{r}^{t}}-\nabla\log{p_{\bm{\theta}}^{t}}\Vert_{2}^{2}\big]\diff t$ holds by Lemma \ref{lemma:iff}.
\end{proof}

\subsection{Validity of Assumption in Theorem \ref{thm:1}}\label{sec:assumption}

\begin{wrapfigure}{r}{0.4\textwidth}
	\vskip -0.3in
	\centering
	\includegraphics[width=\linewidth]{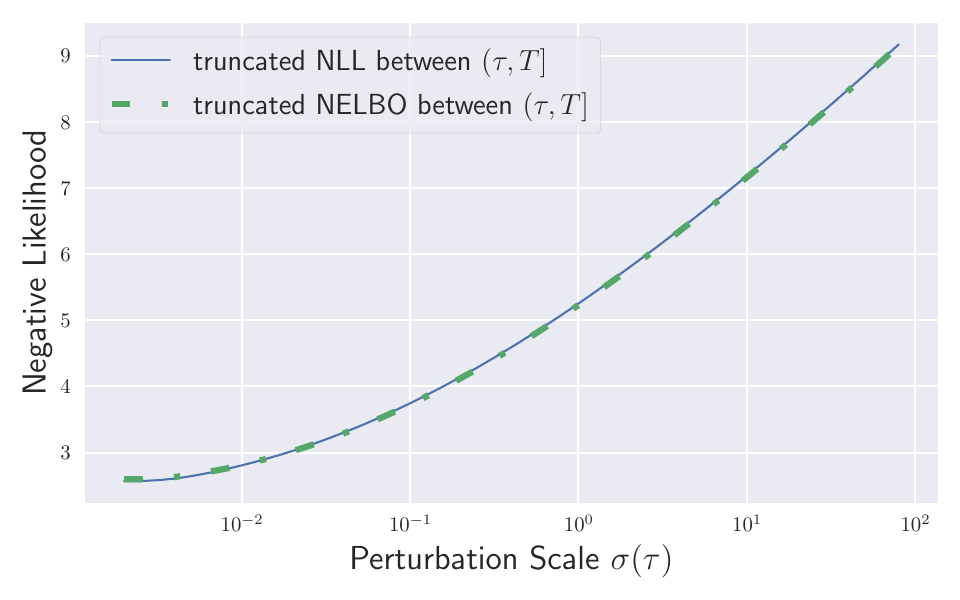}
	\caption{NELBO is uniformly close to NLL on CIFAR-10.}
	\vskip -0.1in
	\label{fig:NLL_NELBO}
\end{wrapfigure}
It now remains to show if the assumptions of Theorem \ref{thm:1} holds in practice. Figure \ref{fig:NLL_NELBO} compares the NLL and NELBO curve of $\mathbf{s}_{\bm{\theta}_{\infty}}$. Figure \ref{fig:NLL_NELBO} shows that the equality condition of NLL and NELBO holds approximately on a wide range of time horizon, in practice.

Figure \ref{fig:NLL_NELBO} is obtained from the EDM checkpoint \cite{karras2022elucidating}. For the NLL and NELBO computation, we follow \citet{kim2022maximum, kim2022soft}: we estimate the \textit{truncated} NLL and NELBO for the purpose of thorough investigation throughout timescales. The truncated NLL and NELBO assumes that the score network is given only on $(\tau,T]$ with $\tau$ a truncation bound. Then, the truncated NLL is the right-hand-side of
\begin{align*}
\mathbb{E}_{\mathbf{x}_{0}\sim p_{r}}[-\log{p_{g}(\mathbf{x}_{0})}]\le\mathbb{E}_{\mathbf{x}_{\tau}\sim p_{r}^{\tau}}[-\log{p_{\bm{\theta},\tau}(\mathbf{x}_{\tau})}]+R_{\tau}(\bm{\theta}),
\end{align*}
where $R_{\tau}(\bm{\theta})=\mathbb{E}_{\mathbf{x}_{0}\sim p_{r}}\big[\int p_{0\tau}(\mathbf{x}_{\tau}\vert\mathbf{x}_{0})\log{\frac{p_{0\tau}(\mathbf{x}_{\tau}\vert\mathbf{x}_{0})}{p_{\bm{\theta}}(\mathbf{x}_{0}\vert\mathbf{x}_{\tau})}}\diff\mathbf{x}_{\tau}\big]$ and $p_{\bm{\theta},\tau}$ is the marginal density at $\tau$ of the generative process. We evaluate $p_{\bm{\theta},\tau}(\mathbf{x}_{\tau})$ by solving the instantaneous change-of-variable formula
\begin{align*}
\frac{\diff}{\diff t}\log{p_{\bm{\theta},t}(\mathbf{x}_{t})}=-\text{tr}\bigg(\nabla\Big[\mathbf{f}(\mathbf{x}_{t},t)-\frac{1}{2}g^{2}(t)\mathbf{x}_{\bm{\theta}}(\mathbf{x}_{t},t)\Big]\bigg)
\end{align*}
from $t=\tau$ to $T$, of which probability flow ODE is
\begin{align*}
\diff\mathbf{x}_{t}=\bigg[\mathbf{f}(\mathbf{x}_{t},t)-\frac{1}{2}g^{2}(t)\mathbf{s}_{\bm{\theta}}(\mathbf{x}_{t},t)\bigg]\diff t.
\end{align*}

Analogously, we evaluate the truncated evidence lower bound as the right-hand-side of
\begin{align*}
\mathcal{L}([0,T])\le \mathcal{L}([\tau,T])+R_{\tau}(\bm{\theta}),
\end{align*}
where
\begin{align*}
\mathcal{L}([\tau,T])&=\frac{1}{2}\int_{\tau}^{T}\mathbb{E}\Big[g^{2}(t)\Vert \nabla\log{p_{0t}(\mathbf{x}_{t}\vert\mathbf{x}_{0})}-\mathbf{s}_{\bm{\theta}}(\mathbf{x}_{t},t)\Vert_{2}^{2}-g^{2}(t)\Vert\nabla\log{p_{0t}(\mathbf{x}_{t}\vert\mathbf{x}_{0})}\Vert_{2}^{2}-2\text{div}(\mathbf{f})\Big]\diff t-\mathbb{E}\big[\log{\pi(\mathbf{x}_{T})}\big].
\end{align*}

We utilize the importance sampling technique (for the integration with respect to $t$ in $\mathcal{L}$) to minimize the estimation variance of the evidence lower bound. To derive the closed-form importance-weighted time, first, observe that the EDM forward diffusion is given by $\sigma^{2}(t)=(\sigma_{min}^{1/\rho}+t(\sigma_{max}^{1/\rho}-\sigma_{min}^{1/\rho}))$. Then, with the importance weight of $\frac{g^{2}(t)}{\sigma^{2}(t)}$, if we define $F(t)=\frac{1}{Z}\int_{\tau}^{t}\frac{g^{2}(s)}{\sigma^{2}(s)}\diff s$ to be the cumulative distribution function of the importance sampler, the importance-weighted time becomes $t=F^{-1}(u)$ for uniformly sampled $u$ on $[0,T]$.

Now, the antiderivative of the importance weight becomes
\begin{align*}
\mathcal{F}(t)&=\int \frac{g^{2}(t)}{\sigma^{2}(t)}\diff t\\
&=\int 2\rho\frac{\sigma_{max}^{1/\rho}-\sigma_{min}^{1/\rho}}{\sigma_{min}^{1/\rho}+t(\sigma_{max}^{1/\rho}-\sigma_{min}^{1/\rho})}\diff t\\
&=2\rho\log{(\sigma_{min}^{1/\rho}+t(\sigma_{max}^{1/\rho}-\sigma_{min}^{1/\rho}))},
\end{align*}
and the normalizing constant becomes
\begin{align*}
Z&=\int_{\tau}^{T}\frac{g^{2}(t)}{\sigma^{2}(t)}\diff t\\
&=\mathcal{F}(T)-\mathcal{F}(\tau)\\
&=2\rho\log{\frac{\sigma_{min}^{1/\rho}+T(\sigma_{max}^{1/\rho}-\sigma_{min}^{1/\rho})}{\sigma_{min}^{1/\rho}+\tau(\sigma_{max}^{1/\rho}-\sigma_{min}^{1/\rho})}}.
\end{align*}
Therefore, we have
\begin{align*}
&t=F^{-1}(u)\\
&\iff u=F(t)=\frac{1}{Z}\int_{\tau}^{t}\frac{g^{2}(s)}{\sigma^{2}(s)}\diff s=\frac{1}{Z}\big(\mathcal{F}(t)-\mathcal{F}(\tau)\big)\\
&\iff u\log{\frac{\sigma_{min}^{1/\rho}+T(\sigma_{max}^{1/\rho}-\sigma_{min}^{1/\rho})}{\sigma_{min}^{1/\rho}+\tau(\sigma_{max}^{1/\rho}-\sigma_{min}^{1/\rho})}}=\log{\frac{\sigma_{min}^{1/\rho}+t(\sigma_{max}^{1/\rho}-\sigma_{min}^{1/\rho})}{\sigma_{min}^{1/\rho}+\tau(\sigma_{max}^{1/\rho}-\sigma_{min}^{1/\rho})}}\\
&\iff \bigg(\frac{\sigma_{min}^{1/\rho}+T(\sigma_{max}^{1/\rho}-\sigma_{min}^{1/\rho})}{\sigma_{min}^{1/\rho}+\tau(\sigma_{max}^{1/\rho}-\sigma_{min}^{1/\rho})}\bigg)^{u}=\frac{\sigma_{min}^{1/\rho}+t(\sigma_{max}^{1/\rho}-\sigma_{min}^{1/\rho})}{\sigma_{min}^{1/\rho}+\tau(\sigma_{max}^{1/\rho}-\sigma_{min}^{1/\rho})}\\
&\iff \big(\sigma_{min}^{1/\rho}+\tau(\sigma_{max}^{1/\rho}-\sigma_{min}^{1/\rho})\big)\bigg(\frac{\sigma_{min}^{1/\rho}+T(\sigma_{max}^{1/\rho}-\sigma_{min}^{1/\rho})}{\sigma_{min}^{1/\rho}+\tau(\sigma_{max}^{1/\rho}-\sigma_{min}^{1/\rho})}\bigg)^{u}=\sigma_{min}^{1/\rho}+t(\sigma_{max}^{1/\rho}-\sigma_{min}^{1/\rho})\\
&\iff t=\bigg(\big(\sigma_{min}^{1/\rho}+\tau(\sigma_{max}^{1/\rho}-\sigma_{min}^{1/\rho})\big)\bigg(\frac{\sigma_{min}^{1/\rho}+T(\sigma_{max}^{1/\rho}-\sigma_{min}^{1/\rho})}{\sigma_{min}^{1/\rho}+\tau(\sigma_{max}^{1/\rho}-\sigma_{min}^{1/\rho})}\bigg)^{u}-\sigma_{min}^{1/\rho}\bigg)\bigg/(\sigma_{max}^{1/\rho}-\sigma_{min}^{1/\rho}).
\end{align*}

\subsection{Why $p_{\bm{\theta}}^{t}$ is defined as a forward marginal rather than a generative marginal}\label{sec:forward_instead_of_generative}

Unintuitively, we define $p_{\bm{\theta}}^{t}$ as a forward-time marginal density rather than a reverse-time generative marginal. We design $p_{\bm{\theta}}^{t}$ as a forward-time marginal by two reasons. First, it saves memory. If it was the reverse-time generative marginal, the generated dataset $\mathcal{G}$ should contain the whole sample trajectories to optimize the time-embedded discriminator, and this could be prohibitive when we run $\sim 1000$ steps to generate a sample. Instead, if we use the forward-time marginal, we could only save the final sample as $\mathcal{G}$, and we optimize the discriminator by diffusing it with \textit{arbitrary} diffusion time with \textit{arbitrary} diffusion noise. Here comes the second advantage: as we could update the discriminator with arbitrary diffusion noise, it attains the data augmentation effect in training the discriminator network, and it could prevent the overfitting issue at large time. On the other hand, if it was the generative marginal, the discriminator guidance needs more sample data to prevent overfitting.

\begin{figure}[p]
	\centering
	\includegraphics[width=\linewidth]{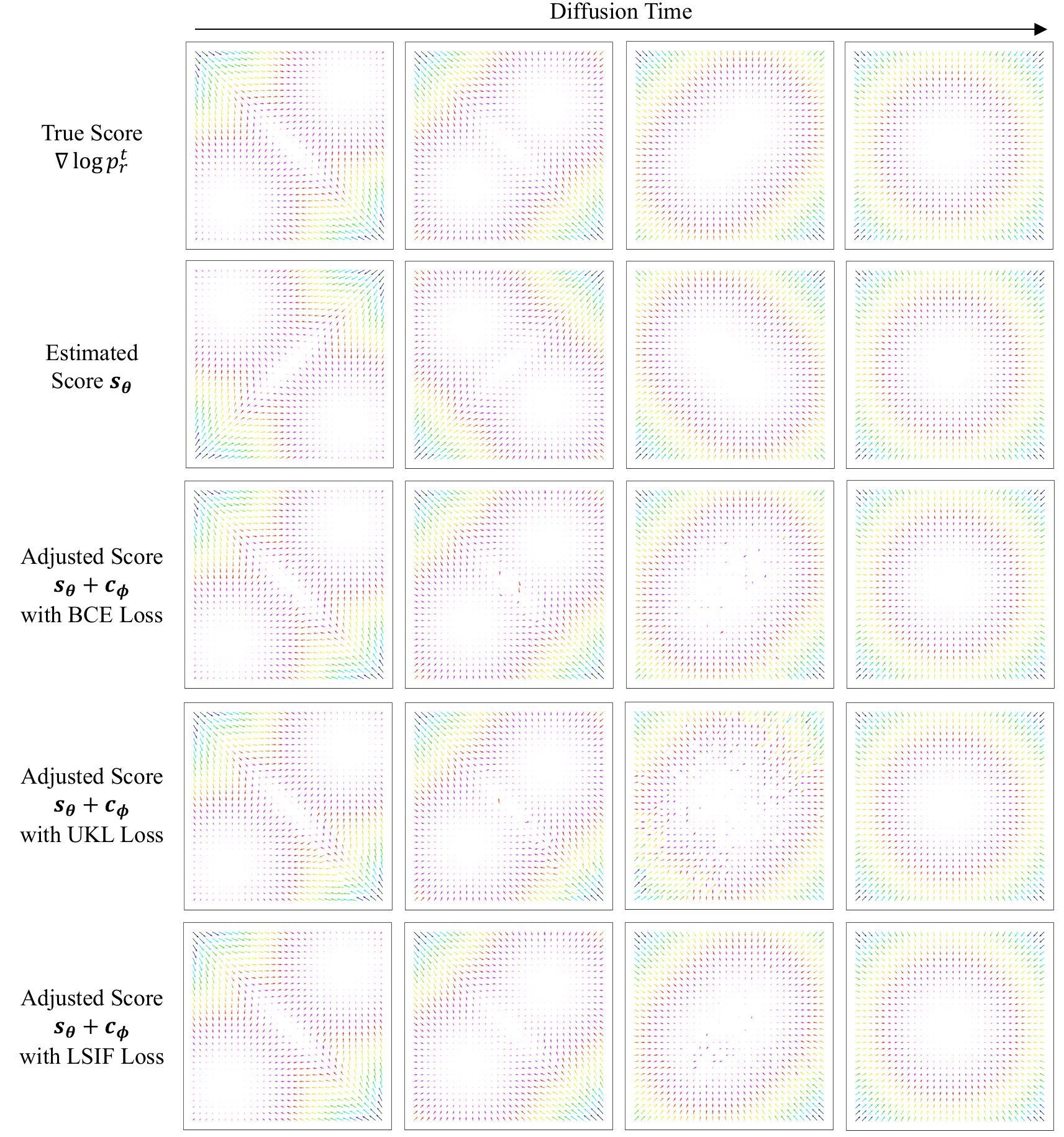}
	\caption{A 2-dimentionsal toy case with a bimodal Gaussian data distribution. In this experiment, we visualize the adjusted scores for BKL/UKL/LSIF losses under the assumption that the estimated score is misleadingly capturing the location of bimodalities.}
	\label{fig:2d_two_gaussian}
\end{figure}	

\begin{figure}[p]
	\centering
	\includegraphics[width=\linewidth]{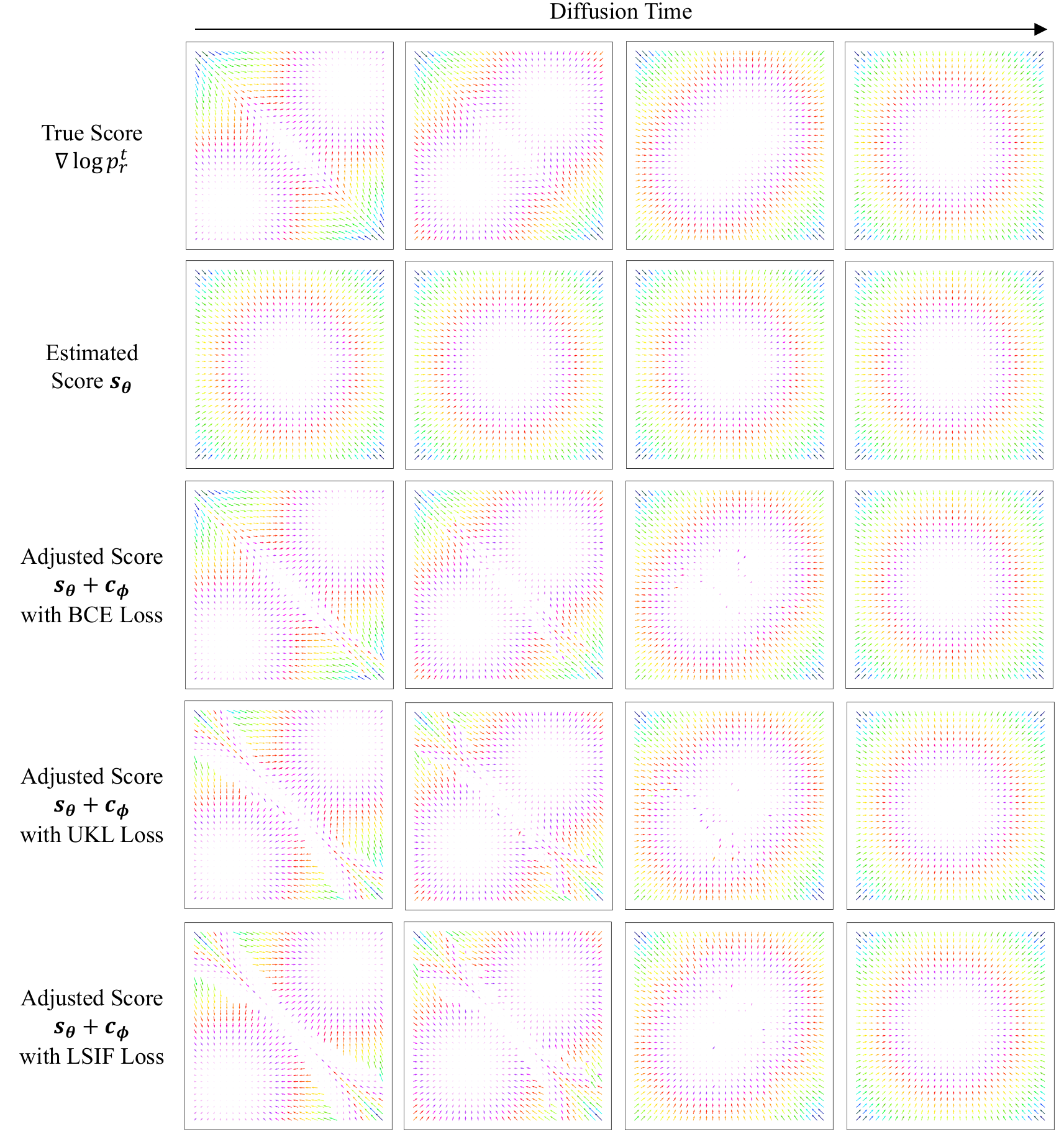}
	\caption{A 2-dimentionsal toy case with a bimodal Gaussian data distribution. In this experiment, we visualize the adjusted scores for BKL/UKL/LSIF losses under the assumption that the estimated score is completely blind to the data distribution, which means that the estimated score $\mathbf{s}_{\bm{\theta}}$ is simply a score function of a standard Gaussian distribution.}
	\label{fig:2d_standard_gaussian}
\end{figure}	

\begin{figure*}[t]
	\centering
	\begin{subfigure}{0.48\linewidth}
		\centering
		\includegraphics[width=0.87\linewidth]{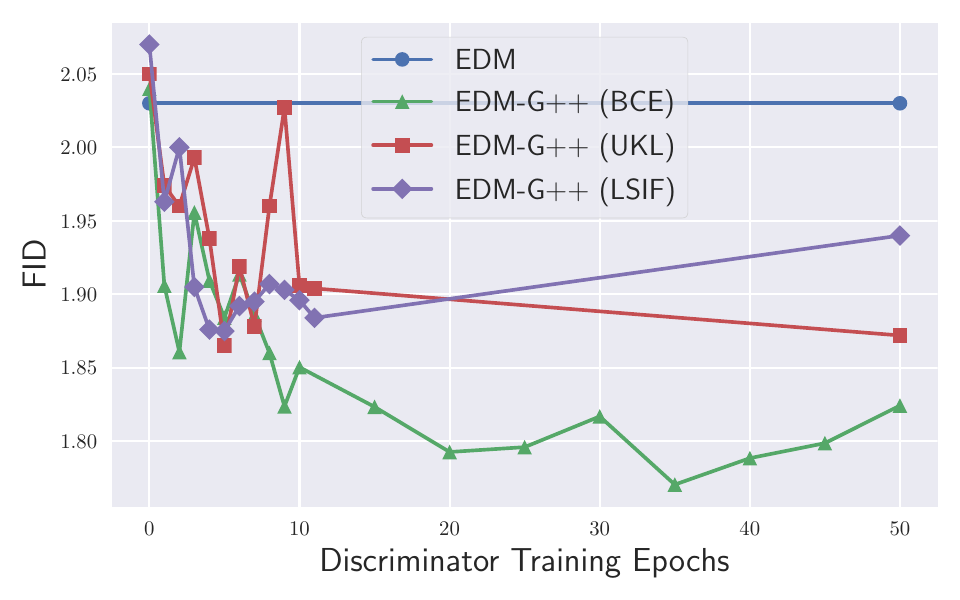}
		\subcaption{FID by Discriminator Training}
	\end{subfigure}
	\hfil
	\begin{subfigure}{0.48\linewidth}
		\centering
		\includegraphics[width=0.87\linewidth]{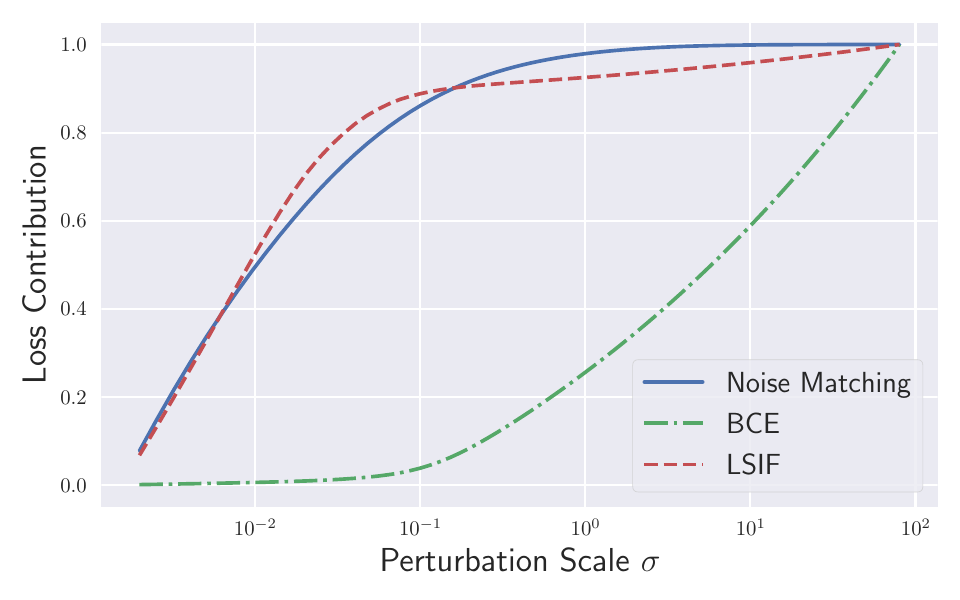}
		\subcaption{Loss Contribution}
	\end{subfigure}
	\caption{Study of EDM on CIFAR-10 with respect to Bregman divergences. (a) illustrates how sample quality is improved by discriminator training with various loss functions, and (b) shows the mechanism of such different FID.}
	\label{fig:Bregman_div}
\end{figure*}

\section{More on Bregman Divergences}\label{sec:Bregman}

The most general framework for learning the density-ratio $\frac{p_{r}^{t}}{p_{\bm{\theta}}^{t}}$ is using the Bregman divergence \cite{sugiyama2012density}. In this section, we investigate the effect of representative Bregman divergences. To begin with, let us define the Bregman divergence in an abstract form. Suppose $r^{*}(\mathbf{x})=\frac{p_{nu}(\mathbf{x})}{p_{de}(\mathbf{x})}$ be the target density-ratio to be estimated with $r_{\bm{\phi}}$, parametrized by $\bm{\phi}$. Then,
\begin{align*}
D_{h}(r^{*}\Vert r_{\bm{\phi}})&=\int p_{de}(\mathbf{x})B_{h}(r^{*}(\mathbf{x})\Vert r_{\bm{\phi}}(\mathbf{x}))\diff\mathbf{x}\\
&=\int p_{de}(\mathbf{x})\Big(h\big(r^{*}(\mathbf{x})\big)-h\big(r_{\bm{\phi}}(\mathbf{x})\big)-\partial h\big(r_{\bm{\phi}}(\mathbf{x})\big)\big(r^{*}(\mathbf{x})-r_{\bm{\phi}}(\mathbf{x})\big)\Big)\diff\mathbf{x},
\end{align*}
where $B_{h}$ is the data-level Bregman divergence. For a twice continuously differentiable convex function $h$ with a bounded derivative $\partial h$, the Bregman divergence quantifies the discrepancy between two density-ratios. Subtracting a constant term $\int p_{de}(\mathbf{x})h(r^{*}(\mathbf{x}))\diff x$, we obtain (up to a constant)
\begin{align*}
D_{h}(r^{*}\Vert r_{\bm{\phi}})=\int p_{de}(\mathbf{x})\Big[\partial h\big(r_{\bm{\phi}}(\mathbf{x})\big)r_{\bm{\phi}}(\mathbf{x})-h\big(r_{\bm{\phi}}(\mathbf{x})\big)\Big]\diff\mathbf{x}-\int p_{nu}(\mathbf{x})\Big[\partial h\big(r_{\bm{\phi}}(\mathbf{x})\big)\Big]\diff\mathbf{x}.
\end{align*}

A few non-exhaustive examples of the Bregman divergence are Least-Squared Importance Fitting (LSIF) \cite{kanamori2009least}, Binary Cross Entropy (BCE) \cite{hastie2009elements}, and Kullback-Leibler Importance Estimation Procedure (KLIEP) \cite{nguyen2010estimating}. LSIF is equivalent to
\begin{align*}
BD_{f_{LSIF}}(r^{*}\Vert r_{\bm{\phi}})&=\frac{1}{2}\int p_{de}(\mathbf{x})\big(r^{*}(\mathbf{x})-r_{\bm{\phi}}(\mathbf{x})\big)^{2}\diff \mathbf{x}\\
&=\frac{1}{2}\int p_{de}(\mathbf{x})r_{\bm{\phi}}^{2}(\mathbf{x})\diff\mathbf{x}-\int p_{nu}(\mathbf{x})r_{\bm{\phi}}(\mathbf{x})\diff\mathbf{x},
\end{align*}
with $h_{LSIF}(r)=(r-1)^{2}/2$. BCE is also widely denoted as Binary Kullback-Leibler (BKL), and is defined with $h_{BKL}(r)=r\log{r}-(1+r)\log{(1+r)}$. KLIEP is also known as the Unbounded Kullback-Leibler (UKL) with $h_{UKL}(r)=r\log{r}-r$, and is a Lagrangian of the constrained optimization problem of
\begin{align*}
D_{KL}(p_{nu}\Vert p_{nu,\bm{\phi}})\text{ subject to }\int p_{nu,\bm{\phi}}(\mathbf{x})\diff\mathbf{x}=1,
\end{align*}
where $p_{nu,\bm{\phi}}(\mathbf{x}):=p_{de}(\mathbf{x})r_{\bm{\phi}}(\mathbf{x})$. The UKL Bregman divergence is defined as the Lagrangian of the above problem by
\begin{align*}
BD_{f_{UKL}}(r^{*}\Vert r_{\bm{\phi}})=\int p_{de}(\mathbf{x})r_{\bm{\phi}}(\mathbf{x})\diff\mathbf{x}-\int p_{nu}(\mathbf{x})\log{r_{\bm{\phi}}(\mathbf{x})}\diff\mathbf{x}.
\end{align*}

Now, we consider the time-dependent Bregman divergence to train the discriminator network. The time-dependent Bregman divergence is defined by
\begin{align*}
\int\lambda(t)D_{h}\bigg(\frac{p_{r}^{t}(\cdot)}{p_{\bm{\theta}}^{t}(\cdot)}\Big\Vert \frac{1-d_{\bm{\phi}}(\cdot,t)}{d_{\bm{\phi}}(\cdot,t)}\bigg)\diff t,
\end{align*}
where $\lambda(t)$ is a time-weighting function. We train the discriminator network with this time-dependent Bregman divergence of LSIF, BCE, and UKL. In a toy 2-dimensional case, Figures \ref{fig:2d_standard_gaussian} and \ref{fig:2d_two_gaussian} visualizes the adjusted score by discriminators trained by aforementioned Bregman divergences: Figure \ref{fig:2d_standard_gaussian} for the case where the estimated score is completely blind to the data score, and Figure \ref{fig:2d_two_gaussian} for the case where the estimated score is confusing on the exact location of bi-modalities of the data distribution. Figures \ref{fig:2d_standard_gaussian} and \ref{fig:2d_two_gaussian} illustrate that BCE estimates the correction term $\mathbf{c}_{\bm{\theta}}$ most robust. 

\begin{wrapfigure}{R}{0.4\textwidth}
	\centering
	\includegraphics[width=\linewidth]{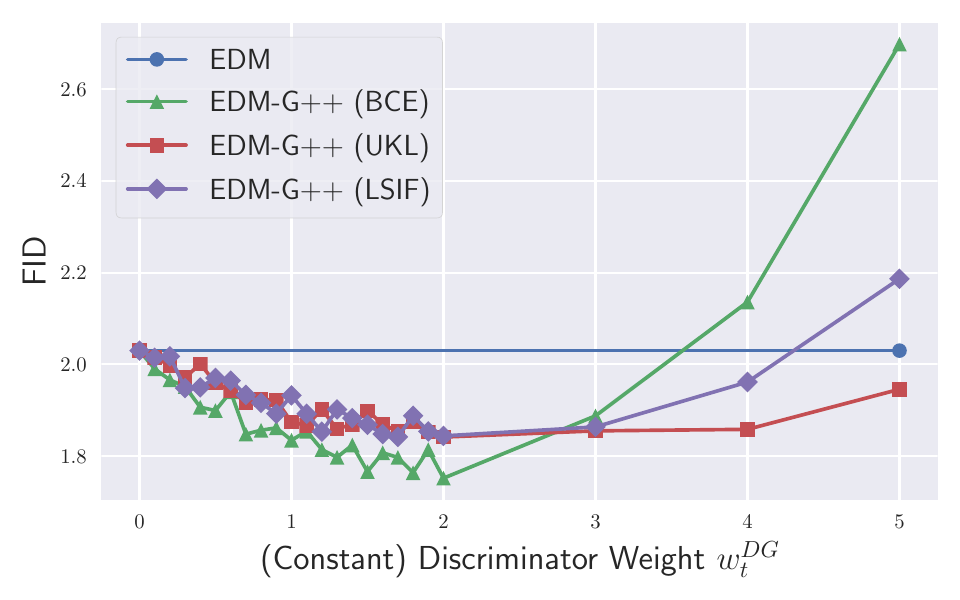}
	\caption{Study of $w_{t}^{DG}$ for various Bregman divergences on CIFAR-10.}
	\label{fig:bregm}
\end{wrapfigure}
Figure \ref{fig:Bregman_div}-(a) shows that the listed divergences have distinctive characteristics in sample quality. Similar to the 2-dimensional case, BCE performs the best. Figure \ref{fig:Bregman_div}-(b) illustrates why such a behavioral difference occurs. We visualize the (normalized) cumulative loss to the perturbation scale. For the noise matching, most loss contribution is on the range of small diffusion scale \cite{kim2022soft}. On the contrary, BCE is qualitatively different from the score loss: most of loss is concentrated on the range of large diffusion scale, and it would be presumably the reason for the arguments in Section \ref{sec:discussion}. Figure \ref{fig:Bregman_div}-(b) shows that LSIF is similar to the noise matching loss. For the UKL loss, we do not plot the cumulative loss because the loss is neither strictly positive nor strictly negative for every diffusion scale.

Figure \ref{fig:bregm} conducts the ablation study of $w_{t}^{DG}$ for various Bregman divergences on CIFAR-10. It illustrates that BCE loss performs the best but UKL loss is the most robust one.

\section{Details on Image-to-Image Translation}\label{sec:I2I_detail}

\begin{figure*}[p]
	\centering
		\includegraphics[width=0.9\linewidth]{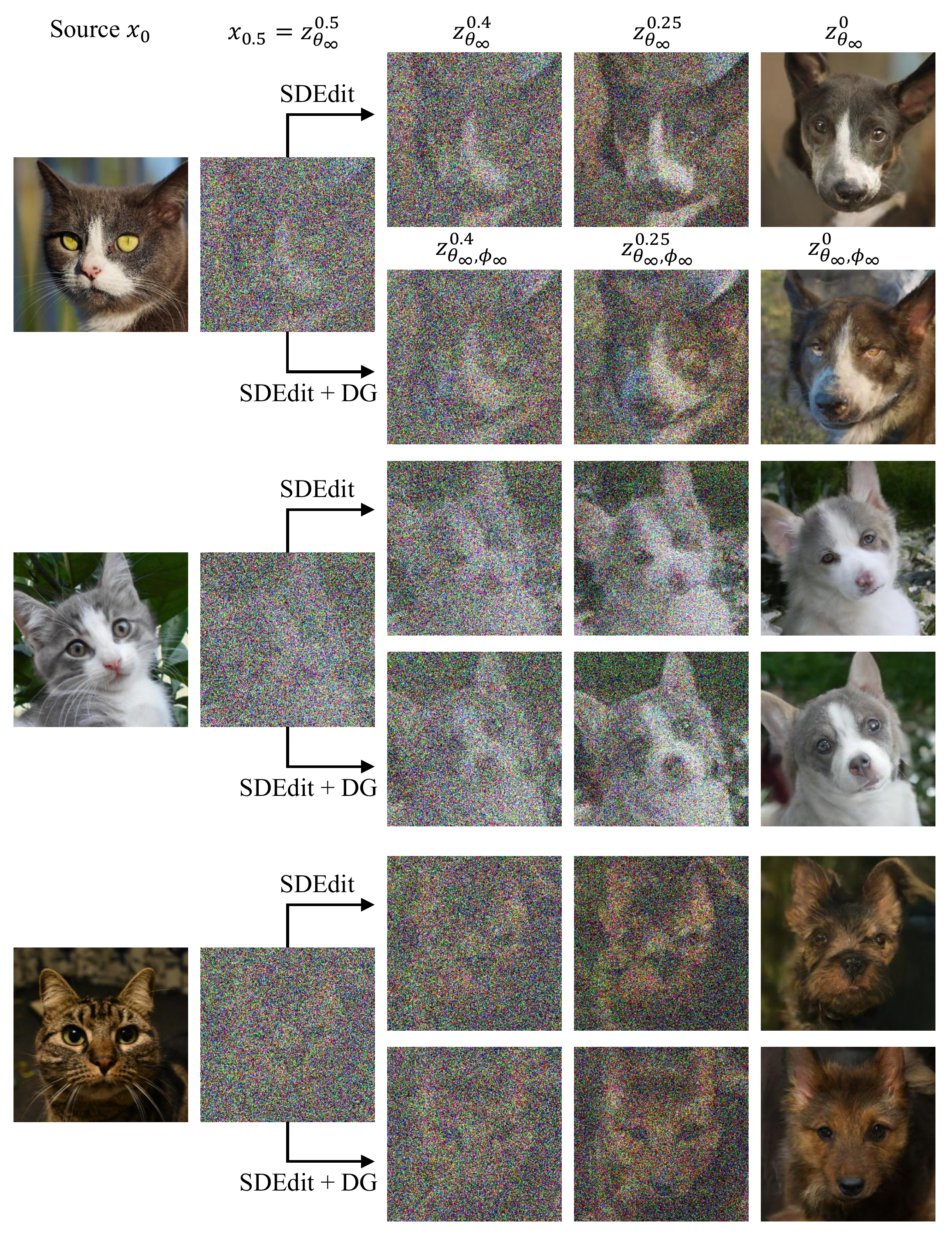}
	\caption{Step-by-step denoising illustration of image-to-image translation.}
	\label{fig:I2I_denoising}
\end{figure*}

Figure \ref{fig:I2I_denoising} illustrates the step-by-step denoising process of I2I translation task. Suppose $\mathbf{x}_{0}\sim\mathcal{S}$ is the source image, and $\mathbf{y}_{0}\sim\mathcal{T}$ is the target image. Then, following SDEdit \cite{meng2021sdedit}, we start denoising from $\mathbf{x}_{\tau}$ with
\begin{align*}
\diff\mathbf{y}_{t}=[\mathbf{f}(\mathbf{y}_{t},t)-g^{2}(t)\mathbf{s}_{\bm{\theta}}(\mathbf{y}_{t},t)]\diff\bar{t}+g(t)\diff\mathbf{\bar{w}}_{t},
\end{align*}
where $\mathbf{s}_{\bm{\theta}}(\mathbf{y}_{t},t)$ is the score network trained on the target domain. Suppose we define $\mathbf{z}_{\bm{\theta}}^{t}$ be the random variable of the solution process: $\mathbf{z}_{\bm{\theta}}^{t_{mid}}=\mathbf{x}_{t_{mid}}$ is the starting variable and $\mathbf{z}_{\bm{\theta}}^{0}$ is the translated variable. Suppose $q_{\bm{\theta}}^{t}$ be the probability distribution of $\mathbf{z}_{\bm{\theta}}^{t}$, $p_{\bm{\theta}}=q_{\bm{\theta}}^{0}$ be the probability distribution of $\mathbf{z}_{\bm{\theta}}^{0}$, and $p_{\bm{\theta}}^{t}$ be the probability distribution of the diffused variable by the forward SDE, starting from $\mathbf{z}_{\bm{\theta}}^{0}$. For the discriminator training, we sample $\mathbf{z}_{t}\sim \alpha p_{\mathcal{S}}^{t}(\mathbf{z}_{t})+(1-\alpha)p_{\bm{\theta}}^{t}(\mathbf{z}_{t})$ for the fake data and $\mathbf{y}_{t}\sim p_{\mathcal{T}}^{t}(\mathbf{y}_{t})$ for the real data. Then, our discriminator guidance becomes
\begin{align*}
\diff\mathbf{y}_{t}=\bigg[\mathbf{f}(\mathbf{y}_{t},t)-g^{2}(t)\Big(\mathbf{s}_{\bm{\theta}}(\mathbf{y}_{t},t)+\nabla\log{\frac{p_{\mathcal{T}}^{t}(\mathbf{y}_{t})}{\alpha p_{\mathcal{S}}^{t}(\mathbf{y}_{t})+(1-\alpha)p_{\bm{\theta}}^{t}(\mathbf{y}_{t})}}\Big)\bigg]\diff\bar{t}+g(t)\diff\mathbf{\bar{w}}_{t}.
\end{align*}
When $t\approx t_{mid}$, we have $p_{\mathcal{S}}^{t}(\mathbf{z}_{t})\gg p_{\bm{\theta}}^{t}(\mathbf{z}_{t})$ for $\mathbf{z}_{t}\sim q_{\bm{\theta}}^{t}$ and the adjusted generative process is approximately
\begin{align*}
\diff\mathbf{y}_{t}\approx\bigg[\mathbf{f}(\mathbf{y}_{t},t)-g^{2}(t)\Big(\mathbf{s}_{\bm{\theta}}(\mathbf{y}_{t},t)+\nabla\log{\frac{p_{\mathcal{T}}^{t}(\mathbf{y}_{t})}{p_{\mathcal{S}}^{t}(\mathbf{y}_{t})}}\Big)\bigg]\diff\bar{t}+g(t)\diff\mathbf{\bar{w}}_{t},
\end{align*}
so the discriminator guidance gives a direct signal to avoid from $p_{\mathcal{S}}^{t}$ when $t\approx t_{mid}$. When $t\approx 0$, we have $p_{\mathcal{S}}^{t}(\mathbf{z}_{t})\ll p_{\bm{\theta}}^{t}(\mathbf{z}_{t})$ for $\mathbf{z}_{t}\sim q_{\bm{\theta}}^{t}$ and the adjusted generative process is approximately
\begin{align*}
\diff\mathbf{y}_{t}\approx\bigg[\mathbf{f}(\mathbf{y}_{t},t)-g^{2}(t)\Big(\mathbf{s}_{\bm{\theta}}(\mathbf{y}_{t},t)+\nabla\log{\frac{p_{\mathcal{T}}^{t}(\mathbf{y}_{t})}{p_{\bm{\theta}}^{t}(\mathbf{y}_{t})}}\Big)\bigg]\diff\bar{t}+g(t)\diff\mathbf{\bar{w}}_{t},
\end{align*}
and the discriminator is guiding the sample denoising toward the desired destination density $p_{\mathcal{T}}^{t}$, rather than the original destination density $p_{\bm{\theta}}^{t}$.

\begin{figure*}[t]
	\centering
		\includegraphics[width=0.9\linewidth]{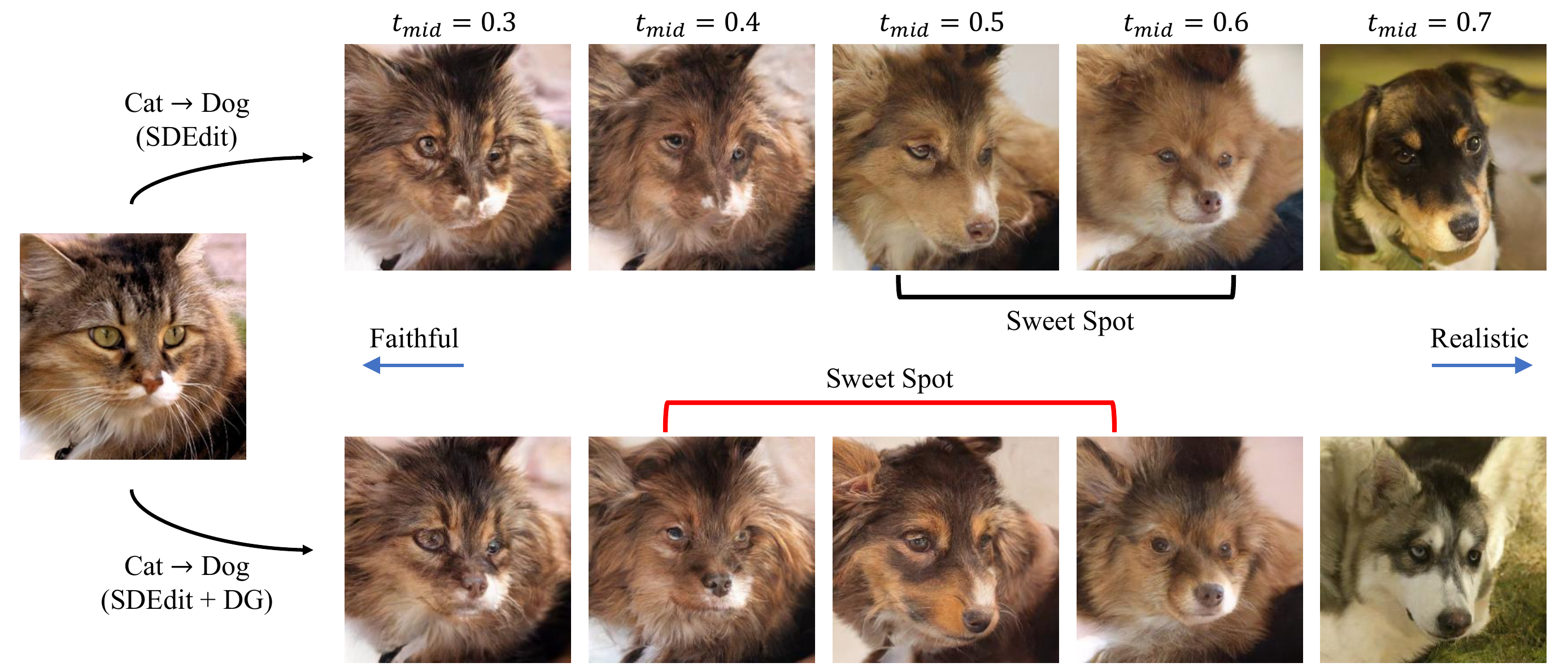}
	\caption{The discriminator guidance swifts the sweet spot to the range of small $t_{0}$.}
	\label{fig:sweet_spot}
\end{figure*}

By applying the discriminator guidance, more realistic samples are generated with relatively small $t_{mid}$, compared to SDEdit, so we could lift the sweet spot of the image-to-image translation to a range of small $t_{mid}$ as in Figure \ref{fig:sweet_spot}.

\section{Experimental Details}

\subsection{Training and Sampling Details}\label{sec:sampling}

\begin{table*}[t]
	\caption{Training and sampling configurations.}
	\label{tab:configurations}
	\scriptsize
	\centering
	\begin{tabular}{lccccccc}
		\toprule
		& \multicolumn{3}{c}{CIFAR-10} & CelebA & FFHQ & \multicolumn{2}{c}{ImageNet256} \\\cmidrule(lr){2-4}\cmidrule(lr){5-5}\cmidrule(lr){6-6}\cmidrule(lr){7-8}
		& LSGM-G++ & \multicolumn{2}{c}{EDM-G++} & Soft Truncation-G++ & EDM-G++ & ADM-G++ & DiT-G++ \\\midrule
		\multicolumn{8}{l}{\textbf{Pre-trained Score Network}}\\
		Model & LSGM & EDM & EDM & Soft Truncation & EDM & ADM & DiT \\
		Class condition & \xmark & \xmark & \cmark & \xmark & \xmark & \cmark & \cmark\\\midrule
		\multicolumn{8}{l}{\textbf{Discriminator Training}}\\
		SDE & LVP & LVP & LVP & CVP & CVP & LVP & LVP \\
		Class condition & \xmark & \xmark & \cmark & \xmark & \xmark & \cmark & \cmark \\
		Time sampling & Importance & Importance & Importance & Importance & Importance & Importance & Importance \\
		$\lambda$ & $\frac{g^{2}}{\sigma^{2}}$ & $\frac{g^{2}}{\sigma^{2}}$ & $\frac{g^{2}}{\sigma^{2}}$ & $\frac{g^{2}}{\sigma^{2}}$ & $\frac{g^{2}}{\sigma^{2}}$ & $\frac{g^{2}}{\sigma^{2}}$ & $\frac{g^{2}}{\sigma^{2}}$ \\
		Minimum diffusion time & 0.01 & $10^{-5}$ & $10^{-5}$ & $10^{-5}$ & $10^{-5}$ & $10^{-5}$ & $10^{-5}$ \\
		EMA & \xmark & \xmark & \xmark & \xmark & \xmark & \xmark & \xmark \\
		Batch size & 128 & 128 & 128 & 128 & 128 & 512 & 512 \\
		$\#\mathcal{D}$ & 50,000 & 50,000 & 50,000 & 10,000 & 60,000 & 1,281,167 & 1,281,167 \\
		$\#\mathcal{G}$ & 50,000 & 25,000 & 50,000 & 10,000 & 60,000 & 400,000 & 1,281,167 \\
		$\#$ Epochs & 280 & 60 & 250 & 150 & 250 & 10 & 7 \\
		GPUs & 1x V100 & 1x V100 & 1x V100 & 1x V100 & 1x V100 & 1x A100 & 1x A100 \\\midrule
		\multicolumn{8}{l}{\textbf{Pre-trained Classifier}}\\
		Model & No classifier & ADM & ADM & ADM & ADM & ADM & ADM \\
		Input shape (data dimension) & \xmark & (B,32,32,3) & (B,32,32,3) & (B,64,64,3) & (B,64,64,3) & (B,256,256,3) & (B,32,32,4) \\
		Output shape (latent dimension) & \xmark & (B,8,8,512) & (B,8,8,512) & (B,8,8,512) & (B,8,8,512) & (B,8,8,512) & (B,8,8,384) \\\midrule
		\multicolumn{8}{l}{\textbf{Shallow U-Net Architecture}}\\
		Input shape (latent dimension) & (B,16,16,180) & (B,8,8,512) & (B,8,8,512) & (B,8,8,512) & (B,8,8,512) & (B,8,8,512) & (B,8,8,384) \\
		Output shape & (B,1) & (B,1) & (B,1) & (B,1) & (B,1) & (B,1) & (B,1)\\
		Minimum value of discriminator (clip) & $10^{-5}$ & 0 & 0 & 0 & 0 & $10^{-5}$ & $10^{-5}$ \\
		Maximum value of discriminator (clip) & $1-10^{-5}$ & 1 & 1 & 1 & 1 & $1-10^{-5}$ & $10^{-5}$ \\
		Class condition & \xmark & \xmark & \cmark & \xmark & \xmark & \cmark & \cmark \\
		$\#$ Resnet blocks & 5 & 4 & 4 & 6 & 6 & 4 & 4 \\
		$\#$ Attention blocks & 3 & 3 & 3 & 5 & 5 & 3 & 3 \\
		Attention resolutions & 16, 8 & 8 & 8 & 8 & 8 & 8 & 8 \\
		Model channel & 128 & 128 & 128 & 128 & 128 & 128 & 128 \\
		Channel multiplier & (1,2) & 1 & 1 & 1 & 1 & 1 & 1 \\\midrule
		\multicolumn{8}{l}{\textbf{Sampling}}\\
		SDE & LVP & WVE & WVE & LVP & WVE & LVP & LVP \\
		Class condition & \xmark & \xmark & \cmark & \xmark & \xmark & \cmark & \cmark \\
		Minimum value of discriminator (clip) & $10^{-3}$ & $10^{-5}$ & $10^{-5}$ & $10^{-5}$ & $10^{-5}$ & $10^{-5}$ & $10^{-5}$ \\
		Maximum value of discriminator (clip) & $1-10^{-3}$ & $1-10^{-5}$ & $1-10^{-5}$ & $1-10^{-5}$ & $1-10^{-5}$ & $1-10^{-5}$ & $10^{-5}$ \\
		Solver & PFODE & PFODE & PFODE & PFODE & PFODE & DDPM & DDPM \\
		Solver accuracy of $\mathbf{s}_{\bm{\theta}}$ & $1^{\text{st}}$-order & $2^{\text{nd}}$-order & $2^{\text{nd}}$-order & $1^{\text{st}}$-order & $2^{\text{nd}}$-order & $1^{\text{st}}$-order & $1^{\text{st}}$-order \\
		Solver type of $\mathbf{s}_{\bm{\theta}}$ & RK45 & Heun & Heun & RK45 & Heun & Euler (DDPM) & Euler (DDPM) \\
		Solver accuracy of $\mathbf{c}_{\bm{\phi}}$ & $1^{\text{st}}$-order & $1^{\text{st}}$-order & $1^{\text{st}}$-order & $1^{\text{st}}$-order & $1^{\text{st}}$-order & $1^{\text{st}}$-order & $1^{\text{st}}$-order \\
		Solver type of $\mathbf{c}_{\bm{\phi}}$ & RK45 & Euler & Euler & RK45 & Euler & Euler & Euler \\
		$t_{mid}$ & 0.1 & 0.01 & 0.01 & 0.01 & 0.01 & 0.1 & 0.1 \\
		NFE & 138 & 35 & 35 & 131 & 71 & 250 & 250 \\
		Classifier Guidance & \xmark & \xmark & \xmark & \xmark & \xmark & \cmark & \cmark \\
		EDS & \xmark & \xmark & \xmark & \xmark & \xmark & \cmark & \xmark \\
		$w_{t}^{CG}$ & 0 & 0 & 0 & 0 & 0 & Adaptive & Adaptive \\
		$w_{t}^{DG}$ & 2 & 2 & Adaptive & Adaptive & Adaptive & Adaptive & 1 \\
		\bottomrule
	\end{tabular}
\end{table*}

\begin{wrapfigure}{R}{0.4\textwidth}
	\vskip -0.3in
	\centering
	\includegraphics[width=\linewidth]{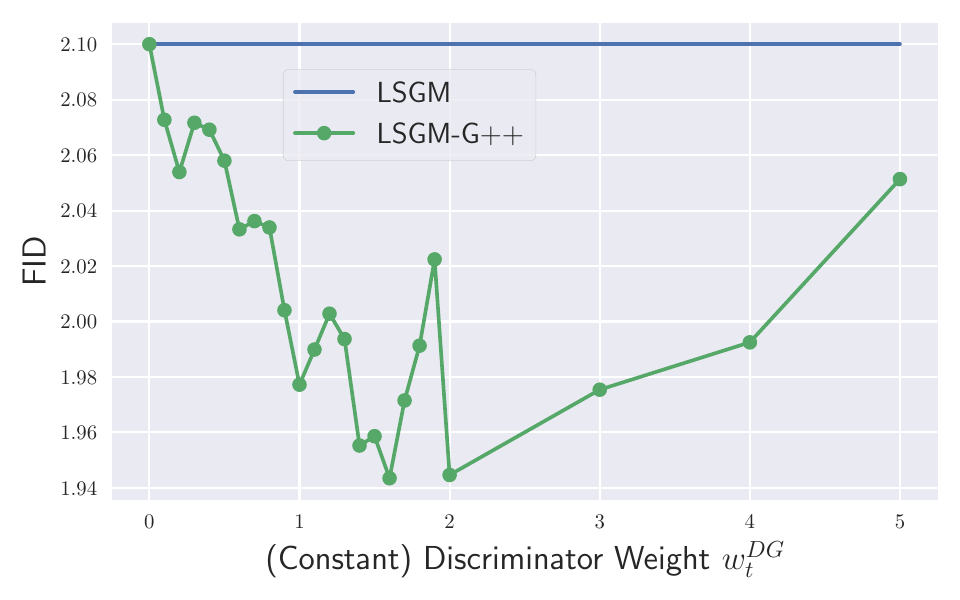}
	\caption{Ablation of $w_{t}^{DG}$ in LSGM-G++ on CIFAR-10.}
	\label{fig:LSGM-G++_dg_weight}
	\vskip -0.25in
\end{wrapfigure}
Table \ref{tab:configurations} presents the experimental configuration for Tables \ref{tab:cifar10}, \ref{tab:human-face}, and \ref{tab:ImageNet256}. Except for ImageNet 256x256, we solve the Probability-Flow ODE (PFODE) \cite{song2020score} for sampling. We use the adjusted PFODE
\begin{align*}
\frac{\diff\mathbf{x}_{t}}{\diff t}=\mathbf{f}(\mathbf{x}_{t},t)-\frac{1}{2}g^{2}(t)(\mathbf{s}_{\bm{\theta}_{\infty}}+w_{t}^{DG}\mathbf{c}_{\bm{\phi}_{\infty}})(\mathbf{x}_{t},t)
\end{align*}
on $t\in[t_{mid},T]$ and the unadjusted PFODE
\begin{align*}
\frac{\diff\mathbf{x}_{t}}{\diff t}=\mathbf{f}(\mathbf{x}_{t},t)-\frac{1}{2}g^{2}(t)\mathbf{s}_{\bm{\theta}_{\infty}}(\mathbf{x}_{t},t)
\end{align*}
on $t\in[0,t_{mid}]$.	

For LSGM-G++, we borrow the pre-trained CIFAR-10 checkpoint of LSGM with best FID \cite{vahdat2021score} at \url{https://github.com/NVlabs/LSGM}. We do not use any of the pre-trained classifier for this latent-diffusion model, and we train a discriminator from scratch. We solve the unadjusted/adjusted PFODEs with explicit Runge-Kutta solver of order 5(4) \cite{dormand1980family}. We find $w_{t}^{DG}=2$ works the best in practice and report the value in Table \ref{tab:cifar10}. Figure \ref{fig:LSGM-G++_dg_weight} presents the ablation study on DG weight.

\begin{figure*}[t]
	\centering
	\begin{subfigure}{0.44\linewidth}
		\centering
		\includegraphics[width=0.87\linewidth]{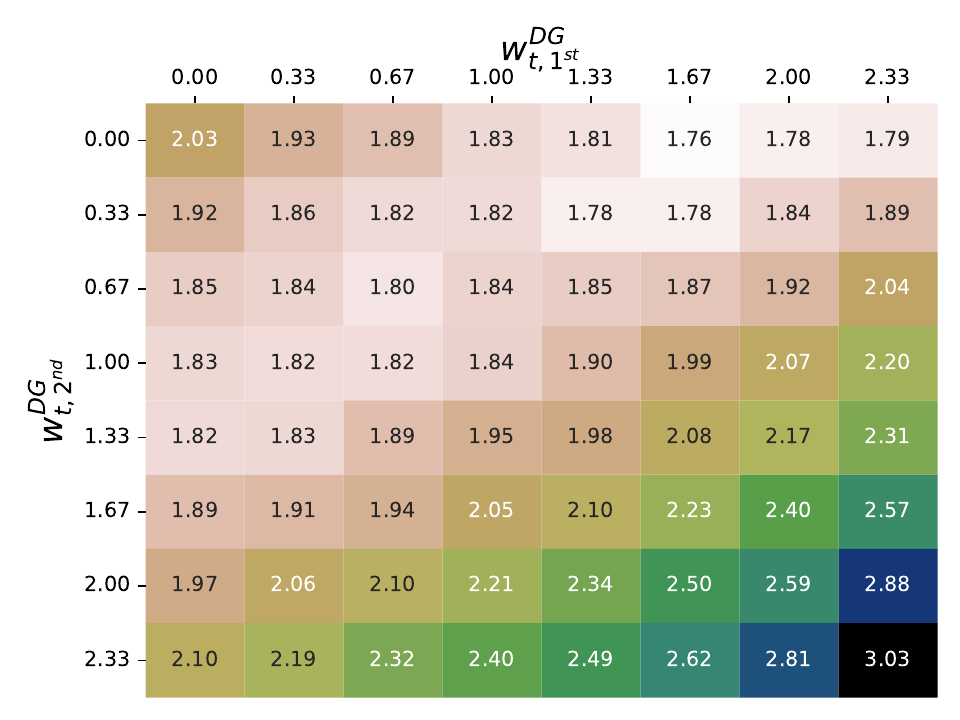}
		\subcaption{Ablation of $w_{t,1^{\text{st}}}^{DG}$ and $w_{t,2^{\text{nd}}}^{DG}$}
	\end{subfigure}
	\hfil
	\begin{subfigure}{0.53\linewidth}
		\centering
		\includegraphics[width=0.87\linewidth]{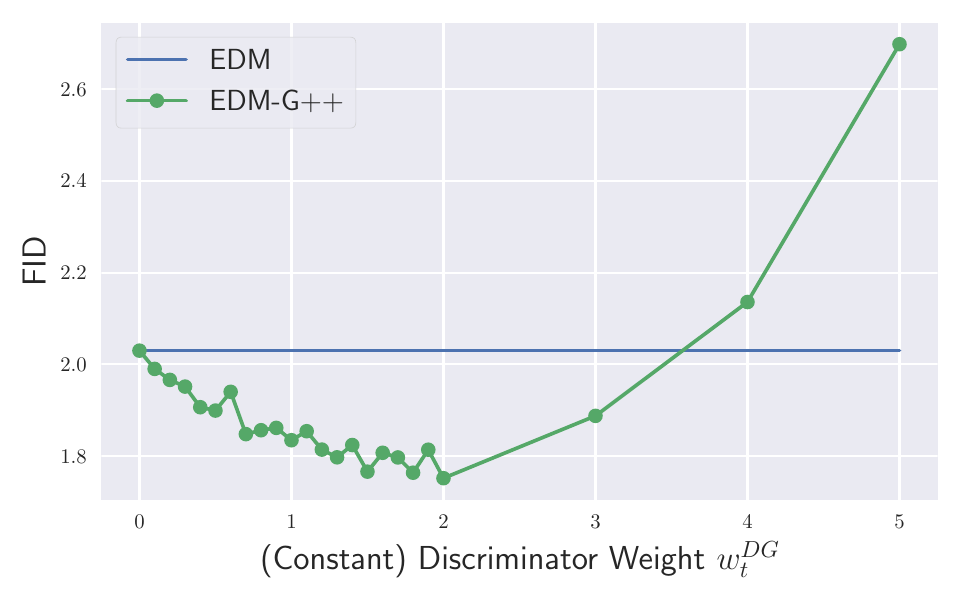}
		\subcaption{Detailed ablation of $w_{t,1^{\text{st}}}^{DG}$}
	\end{subfigure}		
	\caption{Study of Heun solver for EDM on unconditional CIFAR-10.}
	\label{fig:EDM-G++_dg_weight}
\end{figure*}

For EDM-G++, we experiment with unconditional and conditional CIFAR-10, as well as FFHQ 64x64. We use the pre-trained score model of EDM \cite{karras2022elucidating} at \url{https://github.com/NVlabs/edm}. For the classifier, we borrow a pre-trained classifier from ADM \cite{dhariwal2021diffusion} at \url{https://github.com/openai/guided-diffusion} on 64x64 FFHQ, and we train a 32x32 classifier (and freeze it at discriminator training phase) for CIFAR-10 experiment. We use the ImageNet dataset \cite{russakovsky2015imagenet} to train the 32x32 classifier. We follow the identical setting of \citet{dhariwal2021diffusion} to train the 32x32 classifier, except the dataset resolution. We solve the unadjusted/adjusted PFODEs with a Heun solver \cite{ascher1998computer} with pre-designated timesteps that is determined by NFE. As the Heun solver is a 2$^{\text{nd}}$-order numerical solver, we divide the DG weight $w_{t}^{DG}$ into $w_{t,1^{\text{st}}}^{DG}$ and $t_{t,2^{\text{nd}}}^{DG}$, where $w_{t,1^{\text{st}}}^{DG}$ is for the $1^{\text{st}}$-order and $t_{t,2^{\text{nd}}}^{DG}$ is for the $2^{\text{nd}}$-order. For denoising, we construct an intermediate state 
\begin{align*}
\mathbf{\tilde{x}}_{t-\Delta t}=\mathbf{x}_{t}-\Delta t\Big[f(\mathbf{x}_{t},t)-\frac{1}{2}g^{2}(t)(\mathbf{s}_{\bm{\theta}_{\infty}}+w_{t,1^{\text{st}}}^{DG}\mathbf{c}_{\bm{\phi}_{\infty}})(\mathbf{x}_{t},t)\Big].
\end{align*}
Then, we denoise $\mathbf{x}_{t}$ by
\begin{align*}
\mathbf{x}_{t-\Delta t}&=\mathbf{x}_{t}-\Delta t\bigg[\frac{1}{2}\Big(\mathbf{f}(\mathbf{x}_{t},t)-\frac{1}{2}g^{2}(t)(\mathbf{s}_{\bm{\theta}_{\infty}}+w_{t,1^{\text{st}}}^{DG}\mathbf{c}_{\bm{\phi}_{\infty}})(\mathbf{x}_{t},t)\Big)\\
&\quad\quad\quad\quad\quad+\frac{1}{2}\Big(\mathbf{f}(\mathbf{\tilde{x}}_{t-\Delta t},t-\Delta t)-\frac{1}{2}g^{2}(t-\Delta t)(\mathbf{s}_{\bm{\theta}_{\infty}}+w_{t,2^{\text{nd}}}^{DG}\mathbf{c}_{\bm{\phi}_{\infty}})(\mathbf{\tilde{x}}_{t-\Delta t},t-\Delta t)\Big)\bigg]\\
&=\mathbf{x}_{t}-\Delta t\bigg[\frac{1}{2}\Big(\mathbf{f}(\mathbf{x}_{t},t)+\mathbf{f}(\mathbf{\tilde{x}}_{t-\Delta t},t-\Delta t)\Big)-\frac{1}{4}\Big(g^{2}(t)\mathbf{s}_{\bm{\theta}_{\infty}}(\mathbf{x}_{t},t)+\mathbf{s}_{\bm{\theta}_{\infty}}(\mathbf{\tilde{x}}_{t-\Delta t},t-\Delta t)\Big)\\
&\quad\quad\quad\quad\quad-\frac{1}{4}\Big(w_{t,1^{\text{st}}}^{DG}g^{2}(t)\mathbf{c}_{\bm{\phi}_{\infty}}(\mathbf{x}_{t},t)+w_{t,1^{\text{nd}}}^{DG}g^{2}(t-\Delta t)\mathbf{c}_{\bm{\phi}_{\infty}}(\mathbf{\tilde{x}}_{t-\Delta t},t-\Delta t)\bigg].
\end{align*}
Figure \ref{fig:EDM-G++_dg_weight}-(a) illustrates the FID heatmap with respect to $w_{t,1^{\text{st}}}^{DG}$ and $w_{t,2^{\text{nd}}}^{DG}$. It shows that the effect of $w_{t,1^{\text{st}}}^{DG}$ is inverse proportional to that of $w_{t,2^{\text{nd}}}^{DG}$. Also, the line $w_{t,1^{\text{st}}}^{DG}+w_{t,2^{\text{nd}}}^{DG}=2$ performs the best, which is consistent to our intuition. In particular, we emphasize that the best performance happens at $(w_{t,1^{\text{st}}}^{DG}, w_{t,2^{\text{nd}}}^{DG})=(1.67, 0)$, which implies that DG does not has to be applied on the $2^{\text{nd}}$-order correction stage of the Heun solver. This reduces the computational burden of calculating $\mathbf{c}_{\bm{\phi}_{\infty}}(\mathbf{\tilde{x}}_{t-\Delta t},t-\Delta t)$ at every denoising step. Figure \ref{fig:EDM-G++_dg_weight}-(b) shows the detailed abalation with respect to $w_{t,1^{\text{st}}}^{DG}$ with $w_{t,2^{\text{nd}}}^{DG}=0$. From Figure \ref{fig:EDM-G++_dg_weight}-(b), we set $w_{t,1^{\text{st}}}^{DG}=2$ and $w_{t,2^{\text{nd}}}^{DG}=0$ by default for the Heun solver. From Figure \ref{fig:EDM-G++_dg_weight}, we denote $w_{t,1^{\text{st}}}^{DG}$ by $w_{t}^{DG}$ if no confusion arises.

\begin{figure*}[t]
	\centering
	\begin{subfigure}{0.44\linewidth}
		\centering
		\includegraphics[width=0.87\linewidth]{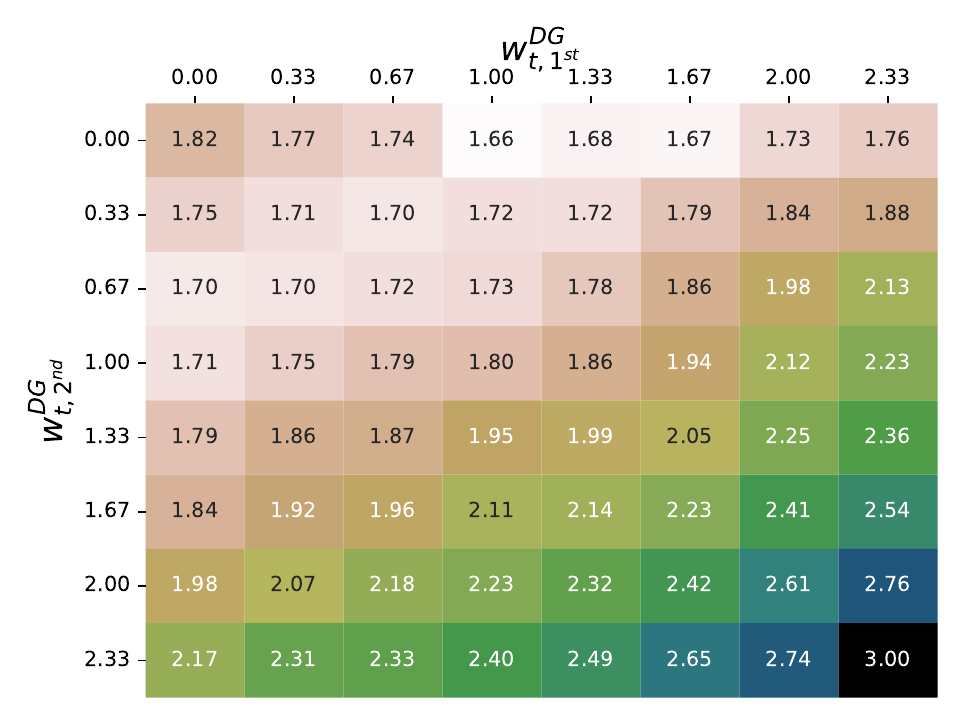}
		\subcaption{Ablation of $w_{t,1^{\text{st}}}^{DG}$ and $w_{t,2^{\text{nd}}}^{DG}$}
	\end{subfigure}
	\hfil
	\begin{subfigure}{0.53\linewidth}
		\centering
		\includegraphics[width=0.87\linewidth]{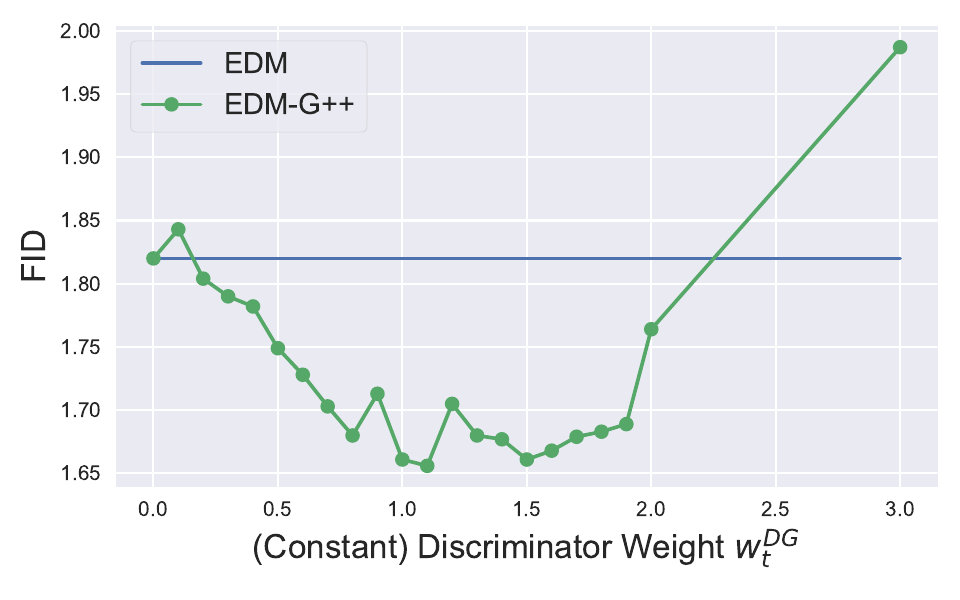}
		\subcaption{Detailed ablation of $w_{t,1^{\text{st}}}^{DG}$}
	\end{subfigure}		
	\caption{Study of Heun solver for EDM on conditional CIFAR-10.}
	\label{fig:EDM-G++_dg_weight_conditional}
\end{figure*}
Similar to the unconditional CIFAR-10, conditional CIFAR-10 has a powerful FID improvement. The optimal weight strategy is at $(w_{t,1^{\text{st}}}^{DG},w_{t,2^{\text{nd}}}^{DG})=(1,0)$ from Figure \ref{fig:EDM-G++_dg_weight_conditional}-(a), and it has FID gain of 1.66. Figure \ref{fig:EDM-G++_dg_weight_conditional}-(b) ablates $w_{t,1^{\text{st}}}^{DG}$ with a fixed $w_{t,2^{\text{nd}}}^{DG}=0$. The best performance is 1.66 when $w_{t,1^{\text{st}}}^{DG}=1.1$. However, if we give $w_{t,1^{\text{st}}}^{DG}=2$ for samples with density-ratio less than 0 in every odd denoising steps and apply $w_{t,1^{\text{st}}}^{DG}=1$ otherwise, we get a better FID of 1.64. For such samples with density-ratio less than 0 in the odd steps, we also make $S_{churn}=4$ to give small stochasticity to avoid local optimum points. There might be better hyperparameter settings because $S_{churn}=4$ is set manually without thorough investigation. We call this approach as Adaptive in Table \ref{tab:configurations}.

Adaptive DG is also effective on FFHQ. With $(w_{t,1^{\text{st}}}^{DG},w_{t,2^{\text{nd}}}^{DG})=(2,0)$, EDM-G++ performs 2.04 in FID, but it drops to 1.98 if we apply the Adaptive DG weight of which adaptive strategy is identical to the conditional CIFAR-10 case.

It is worth to note that the score checkpoint of EDM and classifier checkpoint of ADM are trained under different diffusion strategies. The use of distinctive SDEs leads merging two pre-trained models in one sampler being infeasible. To clarify the difference of diffusion mechanisms, we define Weighted VE (WVE) SDE by $\mathbf{x}_{t}\sim\mathcal{N}(\mathbf{x}_{0},\sigma_{WVE}^{2}(t))$ with $\sigma_{WVE}(t)=\big(\sigma_{min}^{\frac{1}{\rho}}+t(\sigma_{max}^{\frac{1}{\rho}}-\sigma_{min}^{\frac{1}{\rho}})\big)^{\rho}$ for $\sigma_{max}=80$ and $\sigma_{min}=0.002$, which is introduced in \citet{karras2022elucidating}. On the other hand, \citet{ho2020denoising} introduce Linear VP (LVP) with a linear scheduling of $\beta(t)=\beta_{min}+t(\beta_{max}-\beta_{min})$ and \citet{nichol2021improved} propose Cosine VP (CVP) with a cosine scheduling of $\beta(t)$. All of EDM checkpoints are trained under WVE, and the classifier checkpoints are trained with LVP for 32x32 and CVP for 64x64. For the brevity, we only consider the case of LVP.

The key to merging two checkpoints from different diffusion strategies is observing that VE/VP-style SDEs are indeed equivalent under scale translations. Concretely, suppose $p_{t}^{VE}$ and $p_{t}^{VP}$ are the marginal densities of VE/VP SDEs, respectively. Then, it satisfies that $p_{t}^{VE}(\mathbf{x}_{t})=p_{\tau(t)}^{VP}(\nu_{\tau(t)}\mathbf{x}_{t})$ for 
\begin{align}\label{eq:tau}
\begin{split}
&\tau=\frac{-\beta_{min}+\sqrt{\beta_{min}^{2}+2(\beta_{max}-\beta_{min})\log{1+\sigma_{WVE}^{2}(t)}}}{\beta_{max}-\beta_{min}},\\
&\nu_{\tau(t)}=e^{-\frac{1}{2}\int_{0}^{\tau(t)}\beta(s)\diff s}.
\end{split}
\end{align}
With this relations, we put $(\nu_{\tau_{t_{i}}}\mathbf{x}_{t_{i}},\tau_{t_{i}})=(\nu_{\tau(t_{i})}\mathbf{x}_{t_{i}},\tau(t_{i}))$ to the discriminator at the $i$-th denoising step of our sampler in our implementation. We reflect this actual implementation in Algorithm \ref{alg:sampler}. 

For Soft Truncation-G++ with CelebA, we utilize the pre-trained score model from \citet{kim2022soft} at \url{https://github.com/kim-dongjun/soft-truncation} and the 64x64 pre-trained classifier model of ADM. We solve the unadjusted/adjusted PFODEs with explicit Runge-Kutta solver of order 5(4) \cite{dormand1980family}. Similar to the aforementioned checkpoint mismatch issue, we transform time to align the pre-trained classifier on CVPSDE and the pre-trained score on LVPSDE.

\begin{table}[t]
	\caption{Performance on ImageNet 256x256 with ADM.}
	\label{tab:ImageNet256_}
	\tiny
	\centering
	\begin{tabular}{lccccccccc}
		\toprule
		Model & CG & DG & EDS \cite{zheng2022entropy} & FID$\downarrow$ & sFID$\downarrow$ & IS$\uparrow$ & Prec$\uparrow$ & Rec$\uparrow$ & F1$\uparrow$ \\\midrule
		Validation Data & - & - & - & 1.68 & 3.67 & 232.21 & 0.75 & 0.66 & 0.70 \\\midrule
		ADM \cite{dhariwal2021diffusion} & \xmark & \xmark & \xmark & 10.94 & 6.02 & 100.98 & 0.69 & \textbf{0.63} & 0.66 \\
		\cc{15}ADM-G++ (cfg=0.10) & \cc{15}\cmark & \cc{15}\cmark & \cc{15}\cmark & \textbf{4.45}\cc{15} & \textbf{5.38}\cc{15} & \textbf{190.71}\cc{15} & \textbf{0.76}\cc{15} & 0.60\cc{15} & \cc{15}\textbf{0.67} \\\cdashlinelr{1-10}
		ADM-G (cfg=1.50) & \cmark & \xmark & \xmark & 4.59 & 5.25 & 186.70 & 0.82 & 0.52 & 0.64 \\
		ADM-G (cfg=0.75) & \cmark & \xmark & \cmark & 4.01 & 5.15 & 217.25 & 0.82 & 0.53 & 0.64 \\
		\cc{15}ADM-G++ (cfg=0.25) & \cc{15}\cmark & \cc{15}\cmark & \cc{15}\cmark & \cc{15}3.73 & \cc{15}5.03 & \cc{15}204.49 & \cc{15}0.78 & \cc{15}\textbf{0.59} & \cc{15}\textbf{0.67} \\
		\cc{15}ADM-G++ (cfg=0.75) & \cc{15}\cmark & \cc{15}\cmark & \cc{15}\cmark & \cc{15}\textbf{3.18} & \cc{15}\textbf{4.53} & \cc{15}255.74 & \cc{15}0.84 & \cc{15}0.53 & \cc{15}0.65 \\
		\bottomrule
	\end{tabular}
\end{table}

\begin{table}[t]
	\centering
	\caption{Study on $w_{t}^{DG}$ and $w_{t}^{CG}$ in ADM-G++ on ImageNet.}
	\label{tab:ADM-G++_weight}
	\tiny
	\centering
	\begin{tabular}{cccccccccccccc}
		\toprule
		Cases & Implications & $w_{t\ge t_{0}}^{DG}$ & $w_{t<t_{0}}^{DG}$ & $w_{t\ge t_{0}}^{CG}$ & $w_{t<t_{0}}^{CG}$ & $t_{0}$ & EDS & Scaling & FID & sFID & IS & precision & recall \\\midrule
		\multirow{2}{*}{(a)} & \multirow{2}{*}{+ DG} & 0 & 0 & 1.5 & 1.5 & 0 & \xmark & \xmark & 4.59 & 5.25 & 186.70 & 0.82 & 0.52 \\
		& & \cc{15}1 & \cc{15}1 & 1.5 & 1.5 & 0 & \xmark & \xmark & 4.18 & 4.81 & 199.94 & 0.82 & 0.53 \\\cdashlinelr{1-14}
		\multirow{2}{*}{(b)} & \multirow{2}{*}{+ EDS to CG} & 0 & 0 & 1.5 & 1.5 & 0 & \xmark & \xmark & 4.59 & 5.25 & 186.70 & 0.82 & 0.52 \\
		& & 0 & 0 & \cc{15}0.75 & \cc{15}0.75 & 0 & \cc{15}\cmark & \xmark & 4.01 & 5.15 & 217.25 & 0.82 & 0.53 \\\cdashlinelr{1-14}
		\multirow{2}{*}{(c)} & \multirow{2}{*}{+ DG on CG-EDS} & 0 & 0 & 0.75 & 0.75 & 0 & \cmark & \xmark & 4.01 & 5.15 & 217.25 & 0.82 & 0.53 \\
		& & \cc{15}1 & \cc{15}1 & 0.75 & 0.75 & 0 & \cmark & \xmark & 3.69 & 4.75 & 215.32 & 0.82 & 0.54 \\\cdashlinelr{1-14}
		\multirow{2}{*}{(d-1)} & \multirow{2}{*}{+ Adaptive CG} & 1 & 1 & 0.75 & 0.75 & 0 & \cmark & \xmark & 3.69 & 4.75 & 215.32 & 0.82 & 0.54 \\
		& & 1 & 1 & 0.75 & \cc{15}1.5 & \cc{15}650 & \cmark & \xmark & 3.42 & 4.62 & 239.64 & 0.82 & 0.53 \\\cdashlinelr{1-2}
		\multirow{2}{*}{(d-2)} & \multirow{2}{*}{+ Adaptive DG} & 1 & 1 & 0.75 & 1.5 & 650 & \cmark & \xmark & 3.42 & 4.62 & 239.64 & 0.82 & 0.53 \\
		& & 1 & \cc{15}1/0.75 & 0.75 & 1.5 & 650 & \cmark & \cc{15}\cmark & 3.18 & 4.53 & 255.74 & 0.84 & 0.53 \\\cdashlinelr{1-14}
		\multirow{5}{*}{(e)} & \multirow{5}{*}{\shortstack{Final Results}} & 0 & 0 & 0.75 & 1.5 & 650 & \cmark & \xmark & 3.93 & 5.03 & 252.56 & 0.85 & 0.49\\
		& & \cc{15}1 & \cc{15}1/0.75 & 0.75 & 1.5 & 650 & \cmark & \cmark & 3.18 & 4.53 & 255.74 & 0.84 & 0.53 \\
		& & \cc{15}1/3 & 1/0.75 & \cc{15}0.25 & 1.5 & 650 & \cmark & \cmark & 3.73 & 5.03 & 204.49 & 0.78 & 0.59 \\
		& & 1/3 & 1/0.75 & 0.25 & 1.5 & \cc{15}600 & \cmark & \cmark & 4.66 & 5.52 & 183.31 & 0.76 & 0.61 \\
		& & \cc{15}1/5 & \cc{15}1/0.5 & \cc{15}0.1 & \cc{15}2 & 600 & \cmark & \cmark & 4.45 & 5.38 & 190.71 & 0.76 & 0.60 \\
		\bottomrule
	\end{tabular}
\end{table}	
For ADM-G++, we apply the pre-trained checkpoint from \citet{dhariwal2021diffusion} at \url{https://github.com/openai/guided-diffusion}. We exclude the upsampling method from the comparison baseline in order to solely compare models synthesized in the same dimension. In this case, as drawing 1,281,167 samples from ADM is too expensive given our computational budget, we train our discriminator with 400,000 samples. Given our budget, it took nearly 1 day to train 2 epochs. In total, we train 10 epochs, and still we achieve SOTA performance with such a small training budget. Instead of PFODE, we sample data in the same way of DDPM \cite{ho2020denoising}.

\begin{figure*}[t]
	\centering
	\begin{subfigure}{0.48\linewidth}
		\centering
		\includegraphics[width=0.87\linewidth]{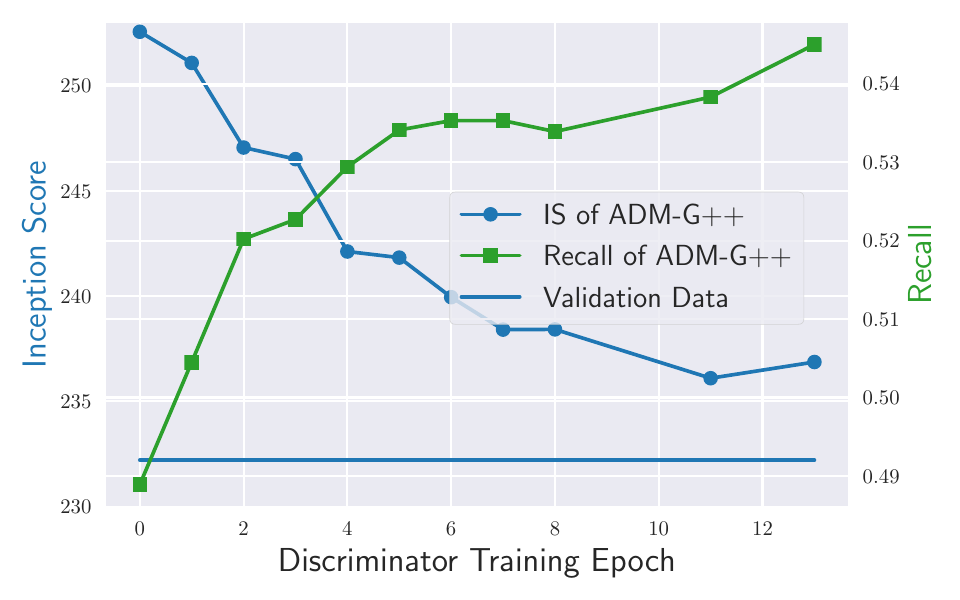}
		\subcaption{Discriminator epoch ablation}
	\end{subfigure}
	\hfil
	\begin{subfigure}{0.48\linewidth}
		\centering
		\includegraphics[width=0.87\linewidth]{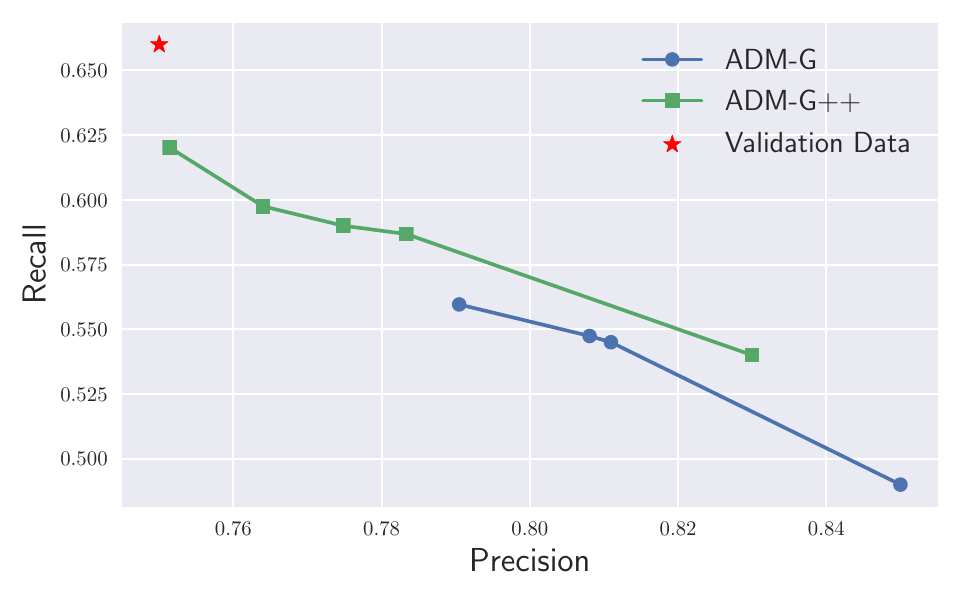}
		\subcaption{Precision/Recall trade-off}
	\end{subfigure}
	\caption{Ablation study of ADM-G++ on ImageNet 256x256.}
	\label{fig:ADM-G++}
\end{figure*}

Table \ref{tab:ADM-G++_weight} shows experimental results of ADM in the ImageNet dataset. Cases of (a), (c), and (e) demonstrate the efficacy of the discriminator guidance. Cases of (b) and (d-1,2) are for achieving the SOTA performance with the discriminotor guidance. For (d-2), we multiply to DG weight by the norm ratio of the classifier guidance and the discriminator guidance, so to balance the guidance scale. Also, for the samples of density-ratio less than 0 in every odd denoising steps, we set $w_{t<t_{0}}^{DG}$ by 0.75. Otherwise, $w_{t<t_{0}}^{DG}=1$. In (e), we apply Adaptive CG/DG to report the performance of ADM-G++. As in (d-2), 1/3 of (e) experiment represents that we boost the DG weight by a factor of 3 for samples of density-ratio less than 0 in every odd denoising steps. 

Figure \ref{fig:ADM-G++}-(a) shows the IS-recall curve by discriminator training epochs. We fix all the other hyperparameters except the discriminator network. Figure \ref{fig:ADM-G++}-(a) illustrates that the discriminator guidance has an effect of increasing the recall metric, which implies that a well-trained discriminator facilitates a diverse generation. Due to the diverse generation, the IS metric is decreased. Similarly, Figure \ref{fig:ADM-G++}-(b) shows that ADM-G++ significantly improves the sample diversity with a sacrifice in the precision metric.

For DiT-G++, we apply the pre-trained checkpoint from \citet{peebles2022scalable} at \url{https://github.com/facebookresearch/DiT}. We draw 1,281,167 number of samples from DiT-XL/2-G to train the discriminator, with the classifier guidance scale of 1.5. In sample generation of DiT-XL/2-G++, we choose $w_{t}^{CG}=1.25$ for $t<t_{0}$ with $t_{0}=200$ and $w_{t}^{CG}=3.0$ otherwise. We use $w_{t}^{DG}=1$ and do not ablate the weight scale.

\subsection{Further Score Training}\label{sec:score_training}

\begin{wrapfigure}{R}{0.4\textwidth}
	\vskip -0.2in
	\centering
	\includegraphics[width=\linewidth]{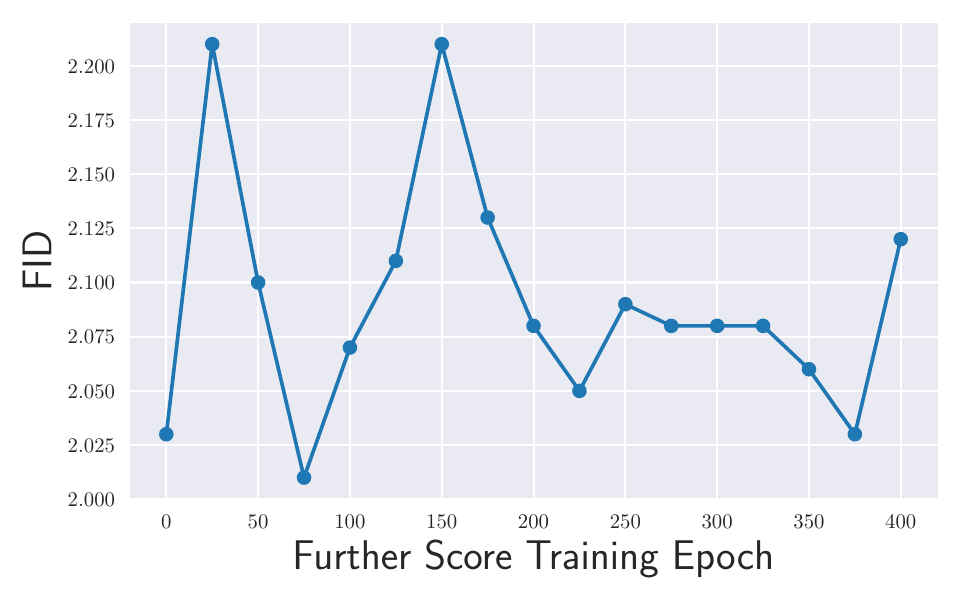}
	\caption{Further score training of a pre-trained score model.}
	\vskip -0.1in
	\label{fig:further_score_training}
\end{wrapfigure}

\subsection{More Ablation Studies}\label{sec:ablation}

\subsubsection{Discriminator Training}

\begin{figure*}[t]
	\centering
	\begin{subfigure}{0.48\linewidth}
		\centering
		\includegraphics[width=0.87\linewidth]{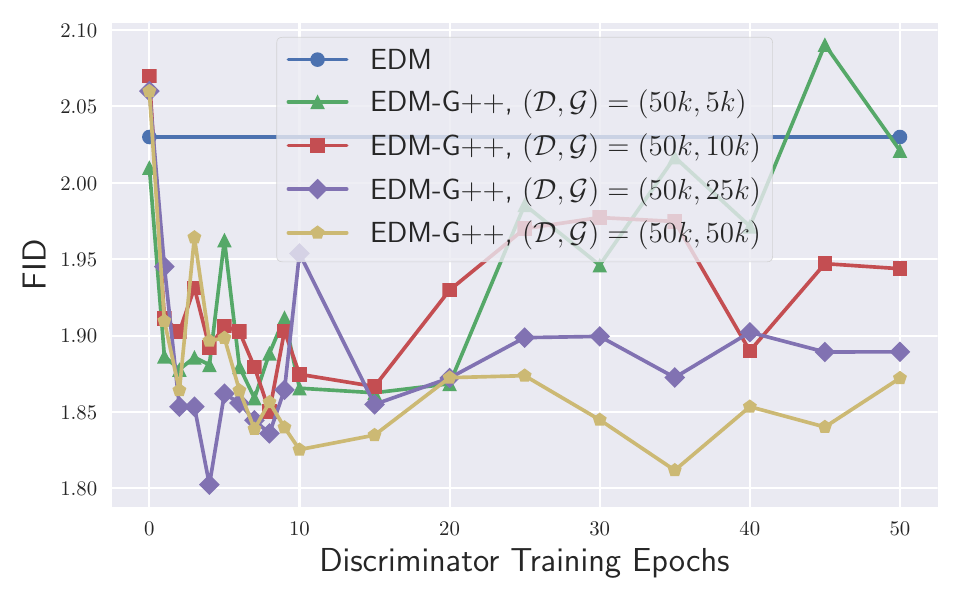}
		\subcaption{Full data and partial sample ablation}
	\end{subfigure}
	\hfil
	\begin{subfigure}{0.48\linewidth}
		\centering
		\includegraphics[width=0.87\linewidth]{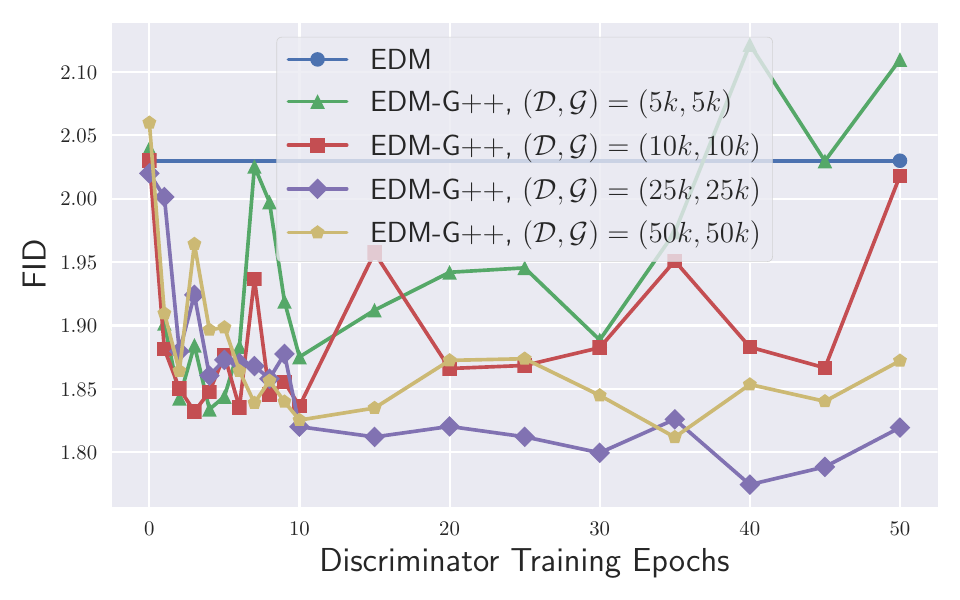}
		\subcaption{Partial data and partial sample ablation}
	\end{subfigure}
	\caption{Study of EDM on CIFAR-10 with respect to the number of training data. (a) is the case with full real data $\mathcal{D}$ and partial sample data $\mathcal{G}$ and (b) is the case with partial $\mathcal{D}$ and partial $\mathcal{G}$.}
	\label{fig:num_data}
\end{figure*}

\textbf{Ablation of $\#$ Training Data} Figure \ref{fig:num_data} shows the ablation study on the number of training data on CIFAR-10. The discriminator guidance requires many samples from $p_{\bm{\theta}_{\infty}}$ to learn the density-ratio, so this could arise the computational issue in some cases. For that, we train the discriminator with the full set of real data and a partial set of sample data in Figure \ref{fig:num_data}-(a). In other words, the case of $(\mathcal{D},\mathcal{G})=(50k, 25k)$ represents for the case of the full use of real data as $\mathcal{D}$ and the half number of sample data as $\mathcal{G}$ in Algorithm \ref{alg:discriminator}. Figure \ref{fig:num_data}-(b), on the other hand, uses the same number of real data for discriminator training. For both experiments, the discriminator suffers from the overffiting issue as the number of generated data decreases. 

Analogous to Figure \ref{fig:num_data}-(a), Figure \ref{fig:num_data}-(b) shows the another ablation study on the number of training data. The situation is a bit different: in this case, we also reduce the size of real data as well as the sample data. Comparing Figure \ref{fig:num_data}-(b) with Figure \ref{fig:num_data}-(a), it is generally not recommendable in any cases. The overfitting arises faster, leading the performance gets worsened faster. 

\begin{figure*}[t]
	\centering
	\begin{subfigure}{0.48\linewidth}
		\centering
		\includegraphics[width=0.87\linewidth]{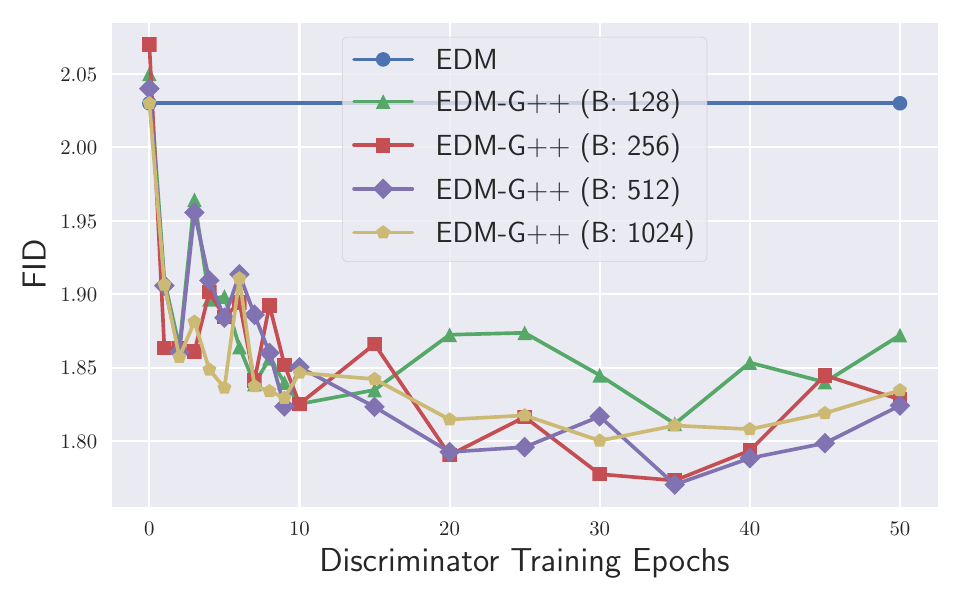}
		\subcaption{Batch size ablation}
	\end{subfigure}
	\hfil
	\begin{subfigure}{0.48\linewidth}
		\centering
		\includegraphics[width=0.87\linewidth]{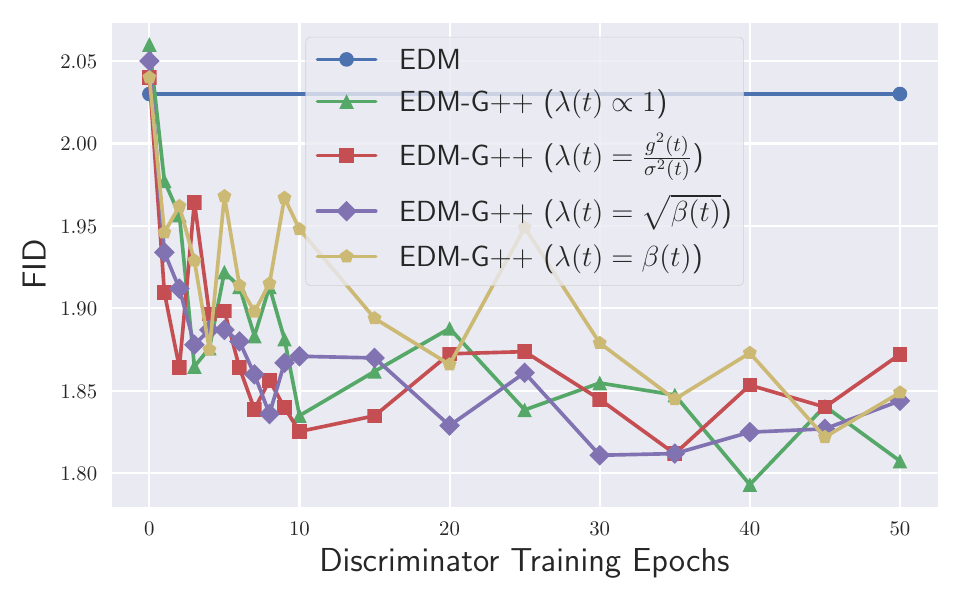}
		\subcaption{Temporal weight $\lambda$ ablation}
	\end{subfigure}
	\caption{Study of EDM on CIFAR-10 with respect to (a) the number of training batch size and (b) the temporal weight $\lambda$.}
	\label{fig:batch_size}
\end{figure*}

\begin{figure*}[t]
	\centering
	\begin{subfigure}{0.48\linewidth}
		\centering
		\includegraphics[width=0.87\linewidth]{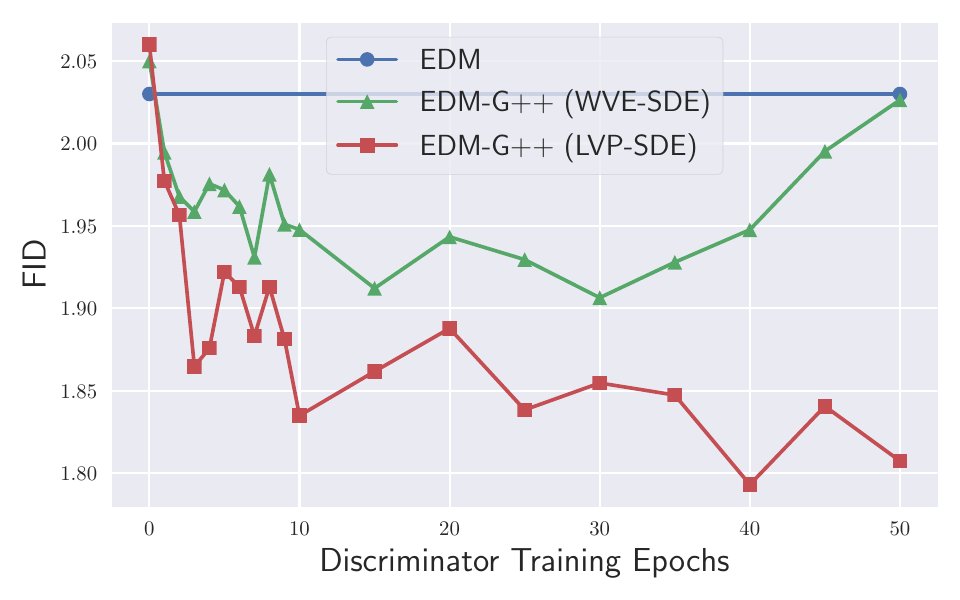}
		\subcaption{Forward SDE ablation}
	\end{subfigure}	
	\hfil	
	\begin{subfigure}{0.48\linewidth}
		\centering
		\includegraphics[width=0.87\linewidth]{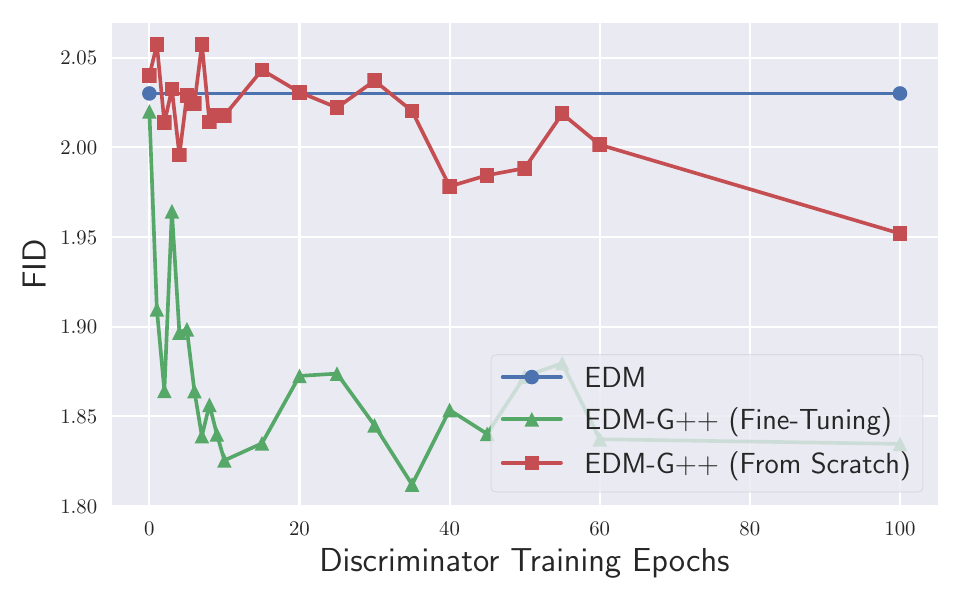}
		\subcaption{Initialization ablation}
	\end{subfigure}
	\caption{Study of EDM on CIFAR-10 with respect to (a) the discriminator diffusing method and (b) the discriminator parameter initialization.}
	\label{fig:scratch}
\end{figure*}

\textbf{Ablation of $\#$ Batch Size} Figure \ref{fig:batch_size}-(a) illustrates the ablation study on the number of training batch size on CIFAR-10. Figure \ref{fig:batch_size}-(a) implies that the number of batch size is not a crucial factor for final performance of discriminator guidance.

\textbf{Ablation of Temporal Weight $\lambda$} Figure \ref{fig:batch_size}-(b) shows indistinguishable FID by the variance of temporal weighting function $\lambda$ for $\mathcal{L}_{\bm{\phi}}$ of Eq. \eqref{eq:discriminator_loss}. We experiment with a uniform distribution $\lambda(t)\propto 1$, an importance-weighted \cite{song2021maximum} distribution $\lambda(t)=\frac{g^{2}(t)}{\sigma^{2}(t)}$ that is one of a common practice, and two $\beta$-related distributions of $\lambda(t)=\sqrt{\beta(t)}$ and $\lambda(t)=\beta(t)$. We train the discriminator with LVP-SDE. Figure \ref{fig:batch_size}-(b) demonstrates that there is no significant difference between the choice of $\lambda$.

\textbf{Ablation of SDE} Figure \ref{fig:scratch}-(a) empirically shows that using LVP-SDE as discriminator SDE performs better than using WVE-SDE. This indicates an important fact: while score training is beneficial with WVE-SDE, discriminator training best fits with LVP-SDE. We, therefore, train the discriminator with LVP-SDE as default.

\textbf{Ablation of Parameter Initialization} Figure \ref{fig:scratch}-(b) shows that the fine-tuning performs strictly better than the training from scratch. For the setting of fine-tuning, we fix the pre-trained classifier parameters, and only train the shallow U-Net for discriminator. In contrast, we train all the parameters including that of the latent extractor, and we set fine-tuning as default in our paper.

Additionally, we find that Exponential Moving Average (EMA) has nearly no effect on the discriminator training, and we do not apply EMA of our discriminator training.

\subsubsection{After Discriminator Training}

\begin{wrapfigure}{R}{0.4\textwidth}
	\vskip -0.2in
	\centering
	\includegraphics[width=\linewidth]{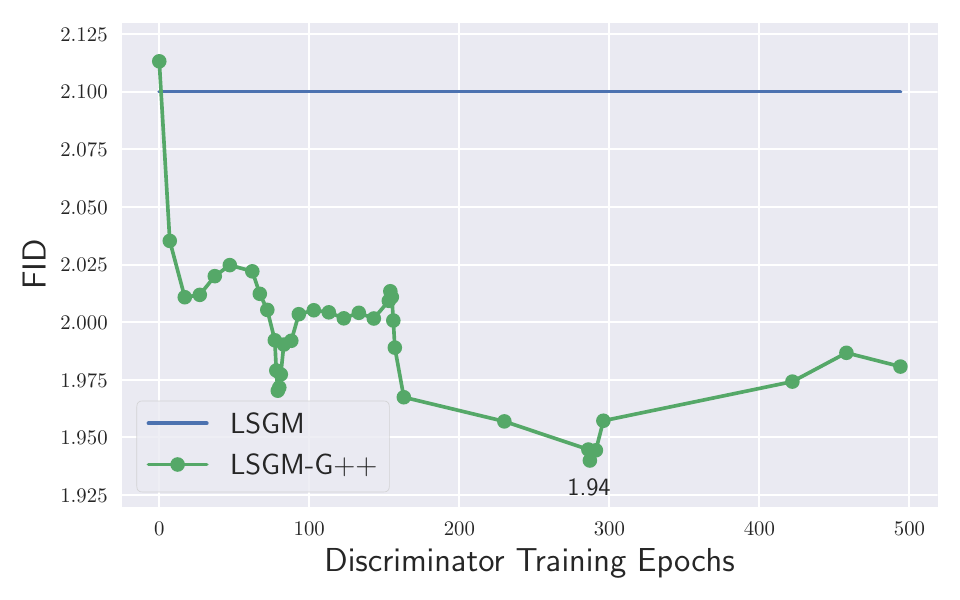}
	\caption{Discriminator epoch ablation.}
	\vskip -0.1in
	\label{fig:LSGM}
\end{wrapfigure}
\textbf{Ablation of Discriminator Training in Latent Diffusion on CIFAR-10} Figure \ref{fig:LSGM} additionally studies LSGM on CIFAR-10. Figure \ref{fig:LSGM} shows FID by discriminator training. The performance saturates after nearly 300 epochs, and it is because we do not use a pre-trained latent extractor for LSGM. \citet{kim2022maximum} clarify that latent diffusion models in general have no diffusion process in the pixel space as long as models use auto-encoder structure to map data to latent. We could detour this problem by defining the diffused data as a decoded diffused latent $\mathbf{x}_{t}=Dec(\mathbf{z}_{t})$, but there is no pre-trained classifier for generic diffusion strategy nor latent diffusion space. We leave it as future work in this direction. In our implementation, we train a U-Net encoder from scratch. This takes a long time to saturate, but FID improves immediately after the discriminator training.

\begin{figure*}[t]
	\centering
	\begin{subfigure}{0.48\linewidth}
		\centering
		\includegraphics[width=0.87\linewidth]{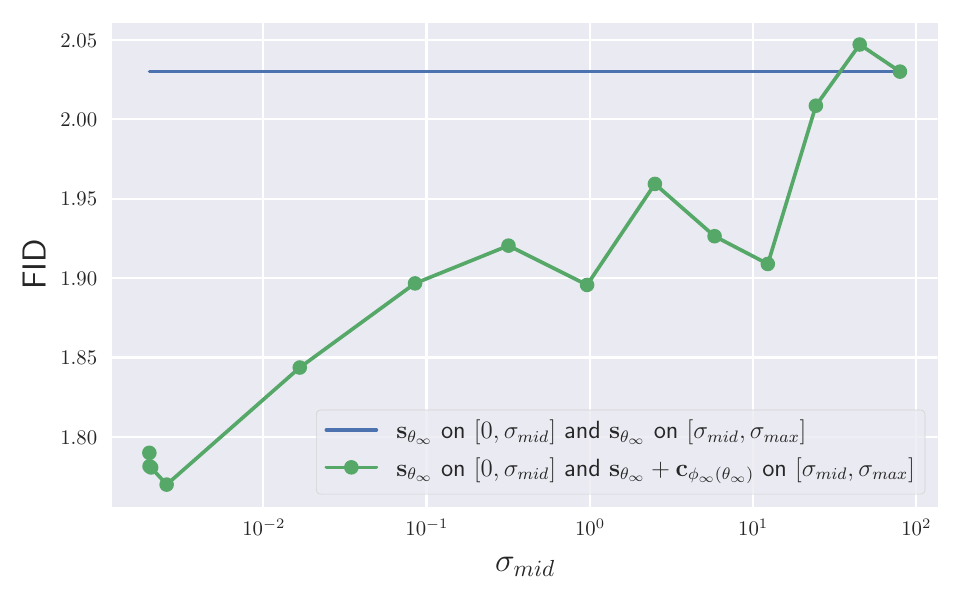}
		\subcaption{Minimum DG scale ablation}
	\end{subfigure}
	\hfil
	\begin{subfigure}{0.48\linewidth}
		\centering
		\includegraphics[width=0.87\linewidth]{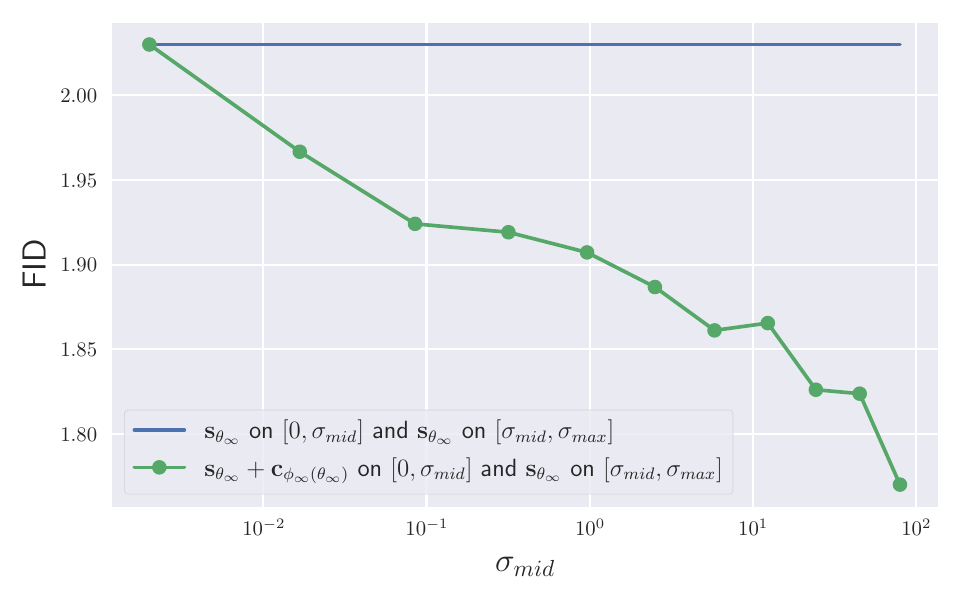}
		\subcaption{Maximum DG scale ablation}
	\end{subfigure}
	\caption{Study of EDM on CIFAR-10 for optimal scales to apply DG: (a) minimum scale, (b) maximum scale.}
	\label{fig:scale_ablation}
\end{figure*}

\textbf{Ablation of optimal scales on CIFAR-10} Figure \ref{fig:scale_ablation} studies the minimum/maximum diffusion scales to apply DG. For the experiment in Figure \ref{fig:scale_ablation}-(a), we divide the diffusion scales by $\sigma_{mid}$, and denoise with the reverse-time generative process
\begin{align*}
\diff\mathbf{x}_{t}=\big[\mathbf{f}(\mathbf{x}_{t},t)-g^{2}(t)(\mathbf{s}_{\bm{\theta}_{\infty}}+\mathbf{c}_{\bm{\phi}_{\infty}})(\mathbf{x}_{t},t)\big]\diff \bar{t}+g(t)\diff\bar{\mathbf{w}}_{t},
\end{align*}	
on the large range $[\sigma_{mid},\sigma_{max}]$, and denoise with
\begin{align*}
\diff\mathbf{x}_{t}=\big[\mathbf{f}(\mathbf{x}_{t},t)-g^{2}(t)\mathbf{s}_{\bm{\theta}_{\infty}}(\mathbf{x}_{t},t)\big]\diff \bar{t}+g(t)\diff\bar{\mathbf{w}}_{t},
\end{align*}

on the small range $[0,\sigma_{mid}]$. With this ablation study, we could find the optimal minimum stopping scale to apply the diffusion scale. Figure \ref{fig:scale_ablation}-(a) illustrates that the optimal stopping scale near $\sigma_{mid}^{*}=0.0025$. This strictly positive optimal scale implies that either the score adjustment becomes inaccurate or the discretization error matters at the range of extremely low scale. In the density-ratio community, \textit{density-chasm problem} is a well-known problem \cite{rhodes2020telescoping, kato2021non} that depicts a poor density-ratio estimation when the two densities have distinctive supports. It arises from the training-test mismatch: the discriminator perfectly classifies real/fake, but the middle area between the real data support and the fake data support remain unoptimized. Therefore, the density-chasm problem is one of the reason for such a strictly positive optimal scale.

Figure \ref{fig:scale_ablation}-(b) studies the maximum diffusion scale to apply DG. Analogous to the experimental setting of Figure \ref{fig:scale_ablation}-(a), we divide the diffusion scales by $\sigma_{mid}$, and denoise with the reverse-time generative process
\begin{align*}
\diff\mathbf{x}_{t}=\big[\mathbf{f}(\mathbf{x}_{t},t)-g^{2}(t)\mathbf{s}_{\bm{\theta}_{\infty}}(\mathbf{x}_{t},t)\big]\diff \bar{t}+g(t)\diff\bar{\mathbf{w}}_{t},
\end{align*}	
on the large range $[\sigma_{mid},\sigma_{max}]$, and denoise with
\begin{align*}
\diff\mathbf{x}_{t}=\big[\mathbf{f}(\mathbf{x}_{t},t)-g^{2}(t)(\mathbf{s}_{\bm{\theta}_{\infty}}+\mathbf{c}_{\bm{\phi}_{\infty}})(\mathbf{x}_{t},t)\big]\diff \bar{t}+g(t)\diff\bar{\mathbf{w}}_{t},
\end{align*}
on the small range $[0,\sigma_{mid}]$. Contrastive to the minimum scale ablation study, the larger the scale DG applied, the better the performance. This means that the actual effect of DG lies in constructing the global shape of the generation, rather than denoising fine-details. In the community of diffusion models, there are only a few works that systematically divide the context generation ability and fine-detail capturing ability of diffusion models. Figure \ref{fig:scale_ablation} clarify that DG effectively adjusts the context generation ability of diffusion models, rather than cleansing the fine-dust in images.

\begin{wrapfigure}{R}{0.4\textwidth}
	\vskip -0.2in
	\centering
	\includegraphics[width=\linewidth]{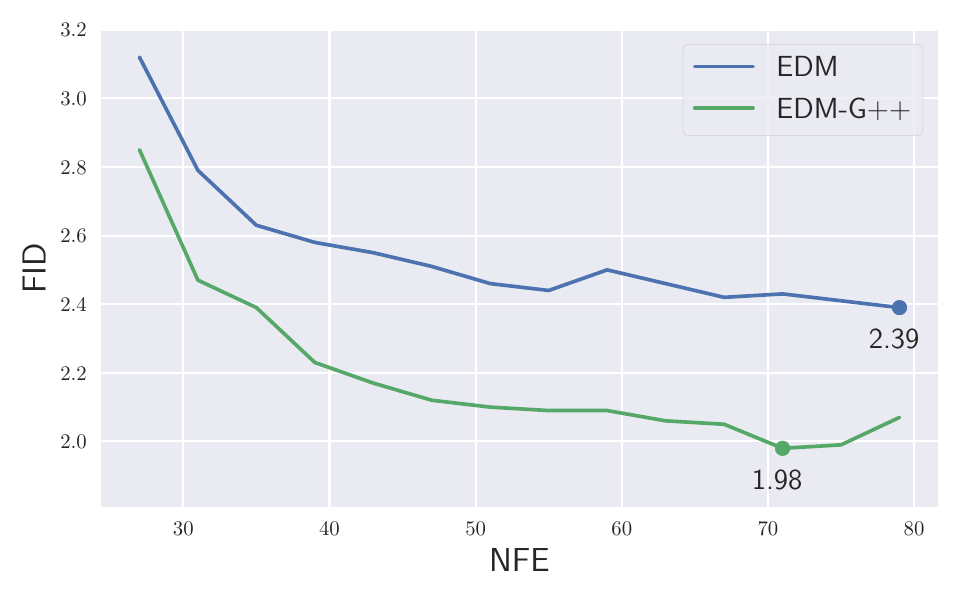}
	\caption{NFE ablation on FFHQ.}
	\vskip -0.3in
	\label{fig:FFHQ}
\end{wrapfigure}
\textbf{Ablation of NFE on FFHQ} Figure \ref{fig:FFHQ} illustrates FID by NFE after discriminator training. For visualization purpose, we select the best hyperparameters to experiment with, except NFE. Similar to NFE ablation on CIFAR-10, the discriminator guidance keeps enhancing FID throughout NFEs. 

\subsection{Uncurated Samples}\label{sec:uncurated}

\begin{figure}[t]
	\centering
	\begin{subfigure}{0.48\linewidth}
		\centering
		\includegraphics[width=\linewidth]{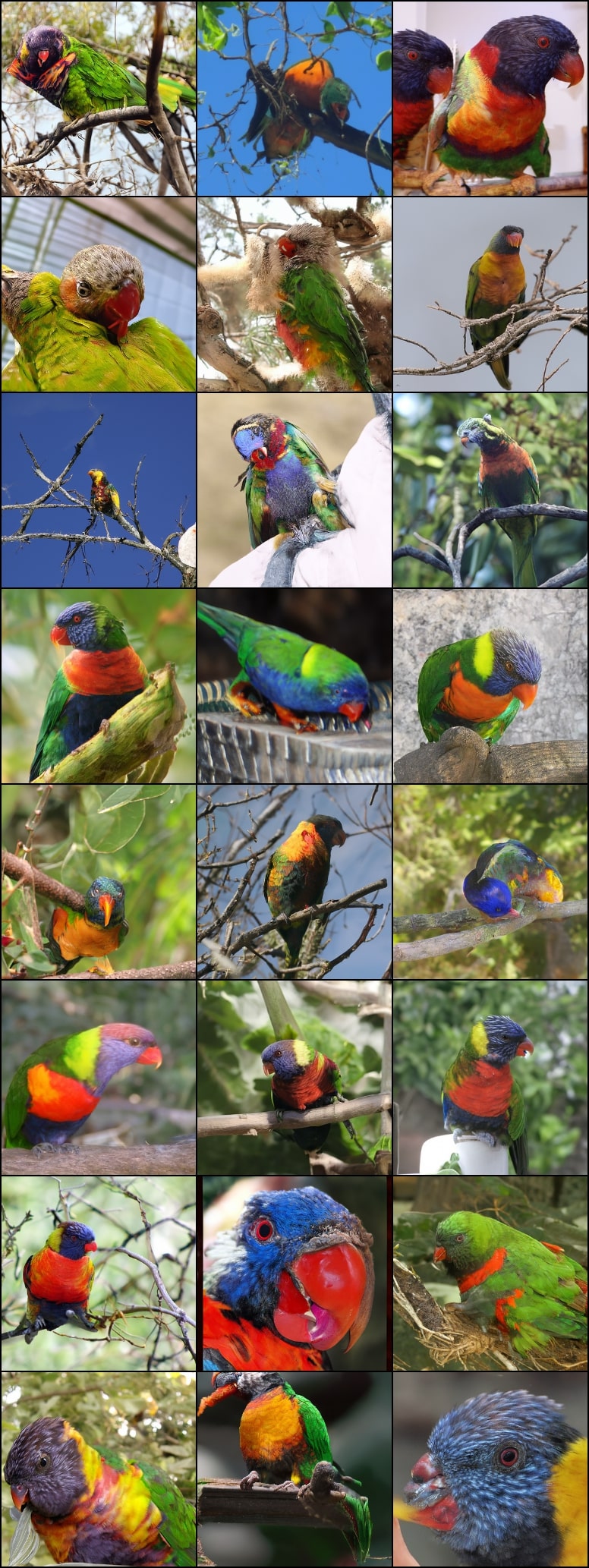}
		\subcaption{ADM (FID 10.94 recall 0.63)}
	\end{subfigure}
	\hfil
	\begin{subfigure}{0.48\linewidth}
		\centering
		\includegraphics[width=\linewidth]{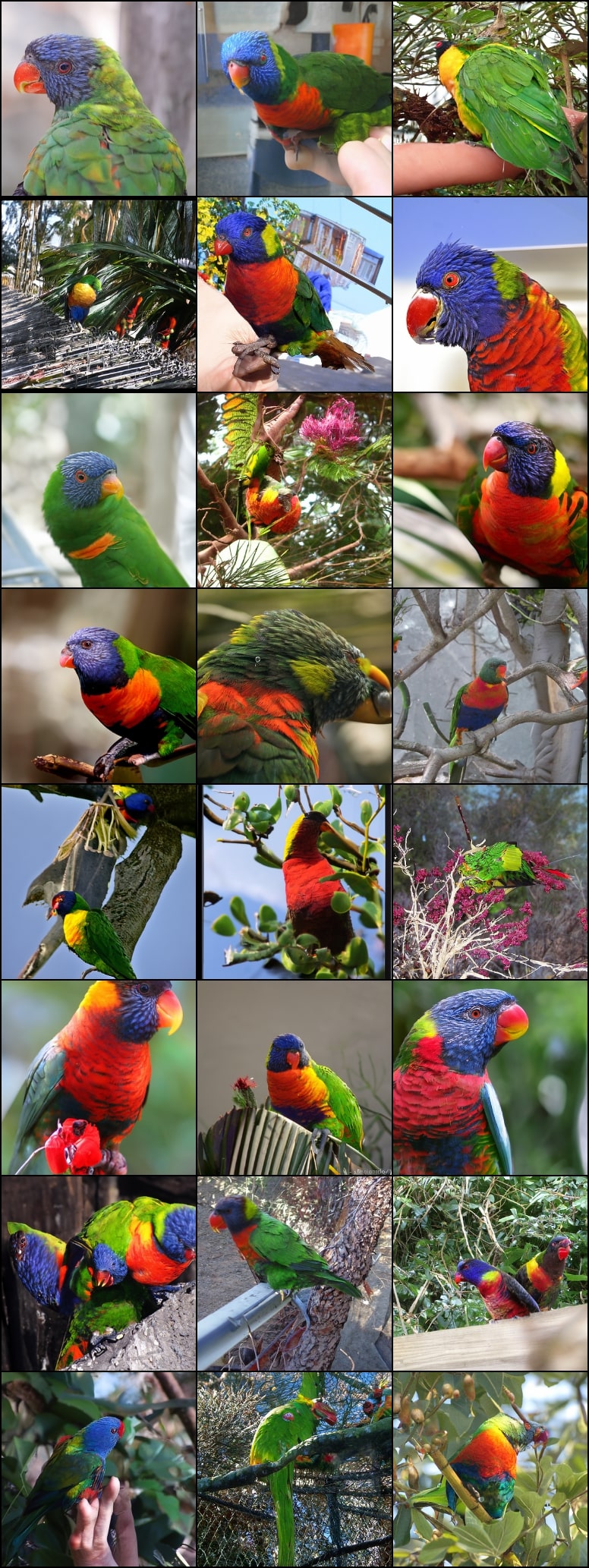}
		\subcaption{ADM-G++ (FID 4.45 recall 0.60)}
	\end{subfigure}
	\caption{Uncurated random samples from lorikeet class (90) (a) ADM with poor FID (10.94) and good recall (0.63), (b) ADM-G++ with good FID (4.45) and good recall (0.60).}
	\label{fig:ImageNet256_recall_90}
\end{figure}

\begin{figure}[t]
	\centering
	\begin{subfigure}{0.48\linewidth}
		\centering
		\includegraphics[width=\linewidth]{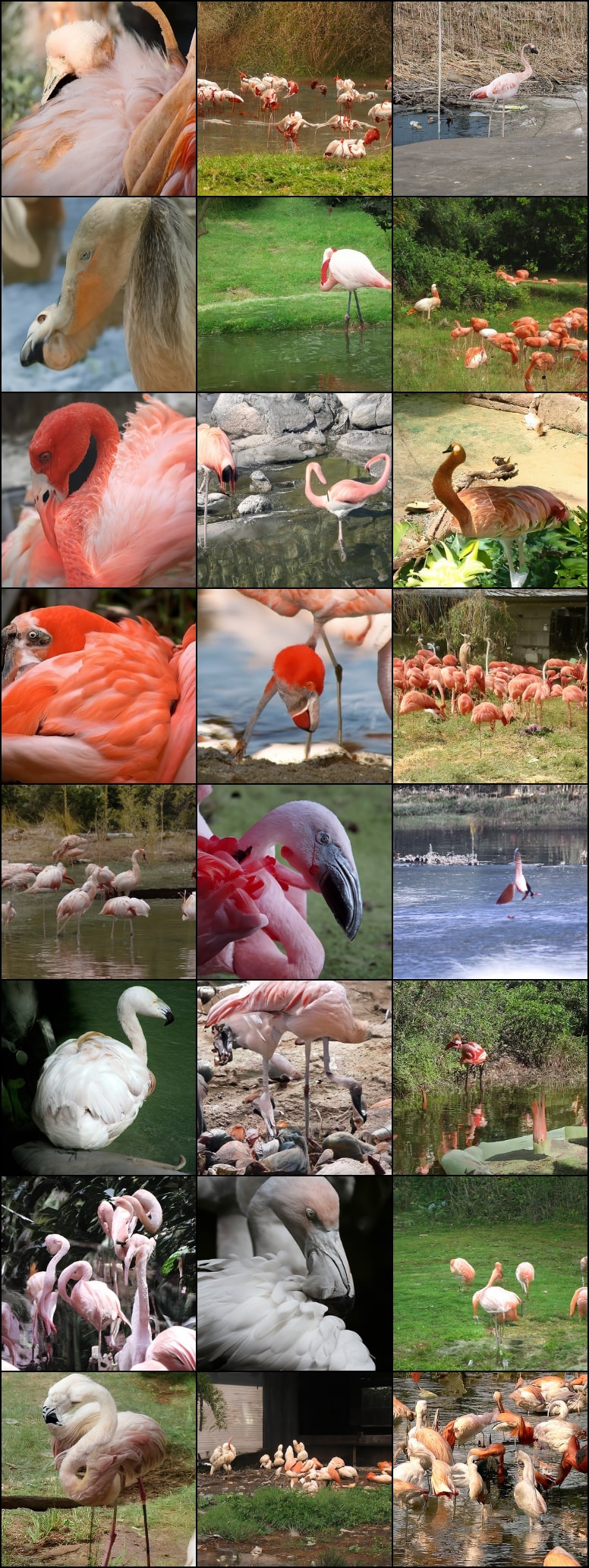}
		\subcaption{ADM (FID 10.94 recall 0.63)}
	\end{subfigure}
	\hfil
	\begin{subfigure}{0.48\linewidth}
		\centering
		\includegraphics[width=\linewidth]{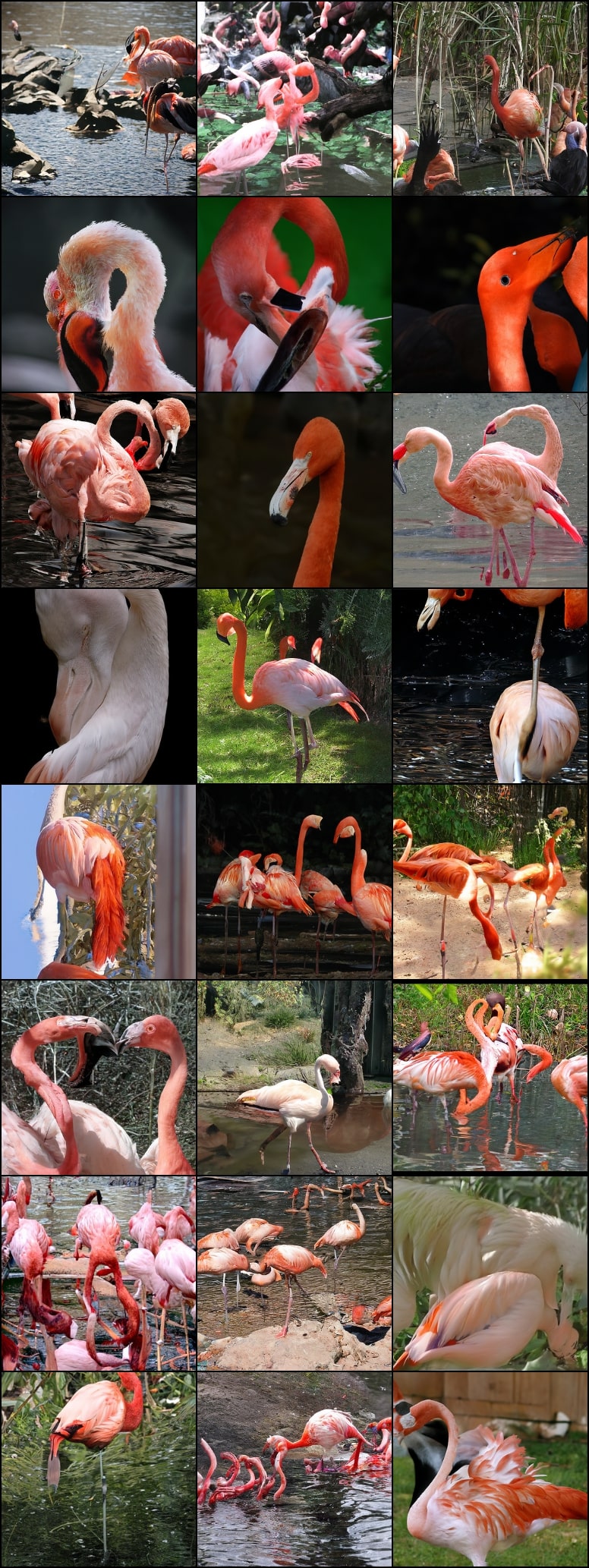}
		\subcaption{ADM-G++ (FID 4.45 recall 0.60)}
	\end{subfigure}
	\caption{Uncurated random samples from flamingo class (130) (a) ADM with poor FID (10.94) and good recall (0.63), (b) ADM-G++ with good FID (4.45) and good recall (0.60).}
	\label{fig:ImageNet256_recall_130}
\end{figure}

\begin{figure}[t]
	\centering
	\begin{subfigure}{0.48\linewidth}
		\centering
		\includegraphics[width=\linewidth]{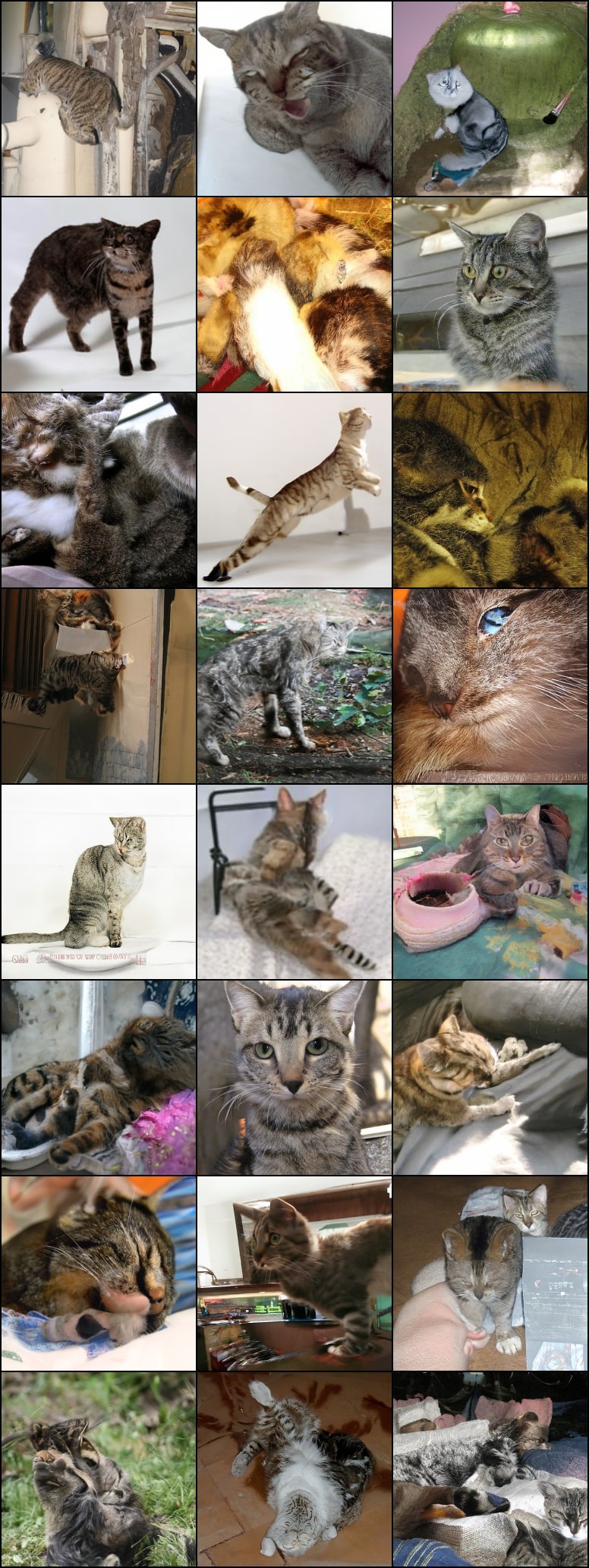}
		\subcaption{ADM (FID 10.94 recall 0.63)}
	\end{subfigure}
	\hfil
	\begin{subfigure}{0.48\linewidth}
		\centering
		\includegraphics[width=\linewidth]{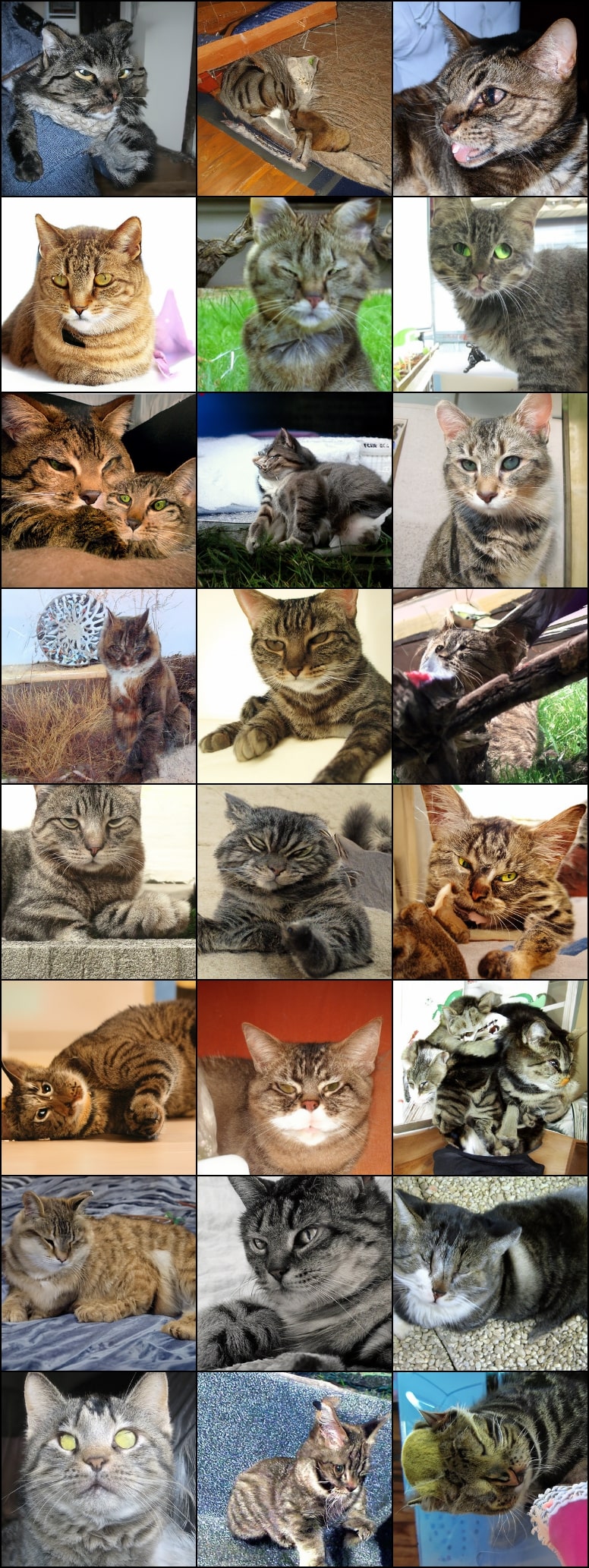}
		\subcaption{ADM-G++ (FID 4.45 recall 0.60)}
	\end{subfigure}
	\caption{Uncurated random samples from tabby cat class (281) (a) ADM with poor FID (10.94) and good recall (0.63), (b) ADM-G++ with good FID (4.45) and good recall (0.60).}
	\label{fig:ImageNet256_recall_281}
\end{figure}

\begin{figure}[t]
	\centering
	\begin{subfigure}{0.48\linewidth}
		\centering
		\includegraphics[width=\linewidth]{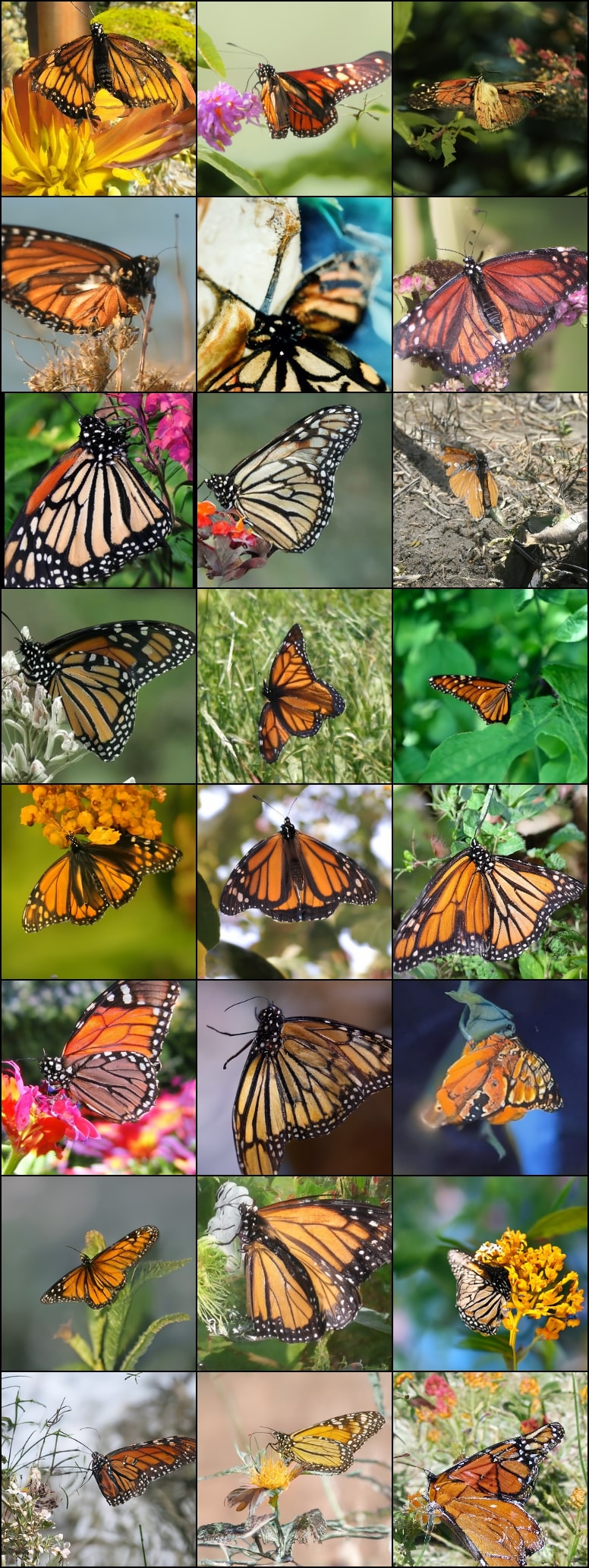}
		\subcaption{ADM (FID 10.94 recall 0.63)}
	\end{subfigure}
	\hfil
	\begin{subfigure}{0.48\linewidth}
		\centering
		\includegraphics[width=\linewidth]{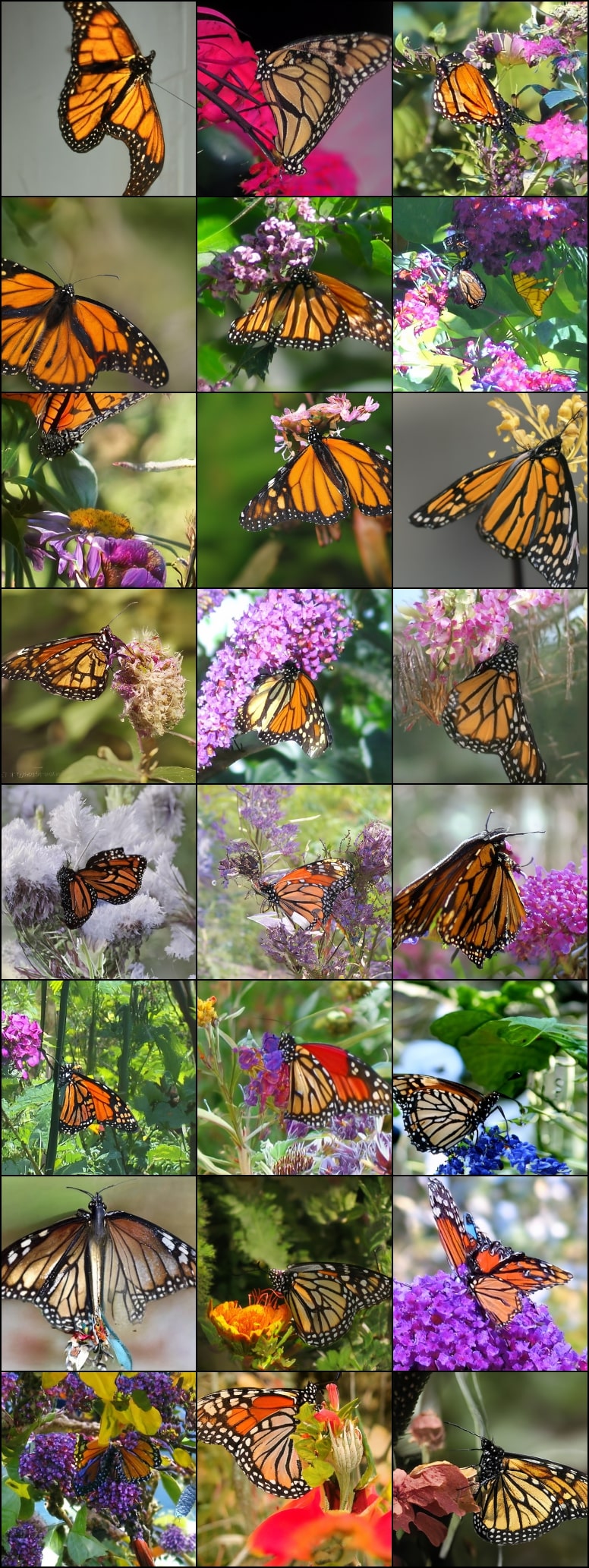}
		\subcaption{ADM-G++ (FID 4.45 recall 0.60)}
	\end{subfigure}
	\caption{Uncurated random samples from monarch butterfly class (323) (a) ADM with poor FID (10.94) and good recall (0.63), (b) ADM-G++ with good FID (4.45) and good recall (0.60).}
	\label{fig:ImageNet256_recall_323}
\end{figure}

\begin{figure}[t]
	\centering
	\begin{subfigure}{0.48\linewidth}
		\centering
		\includegraphics[width=\linewidth]{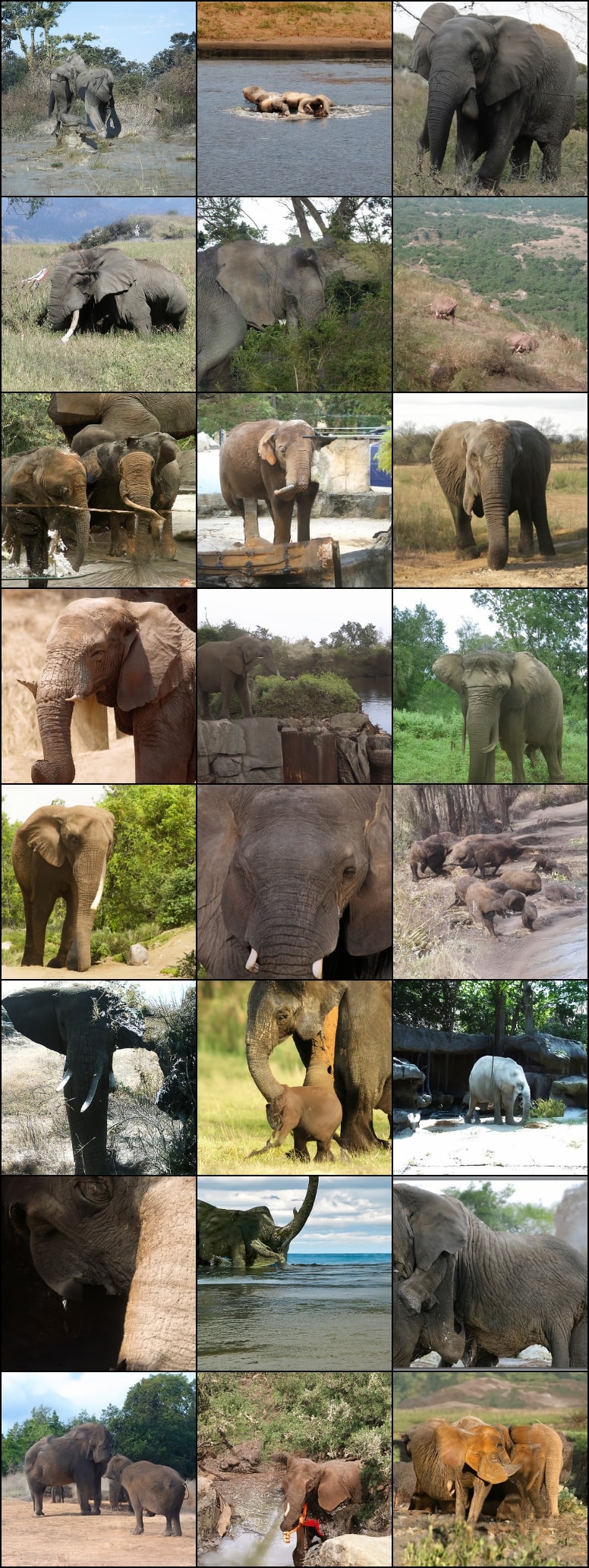}
		\subcaption{ADM (FID 10.94 recall 0.63)}
	\end{subfigure}
	\hfil
	\begin{subfigure}{0.48\linewidth}
		\centering
		\includegraphics[width=\linewidth]{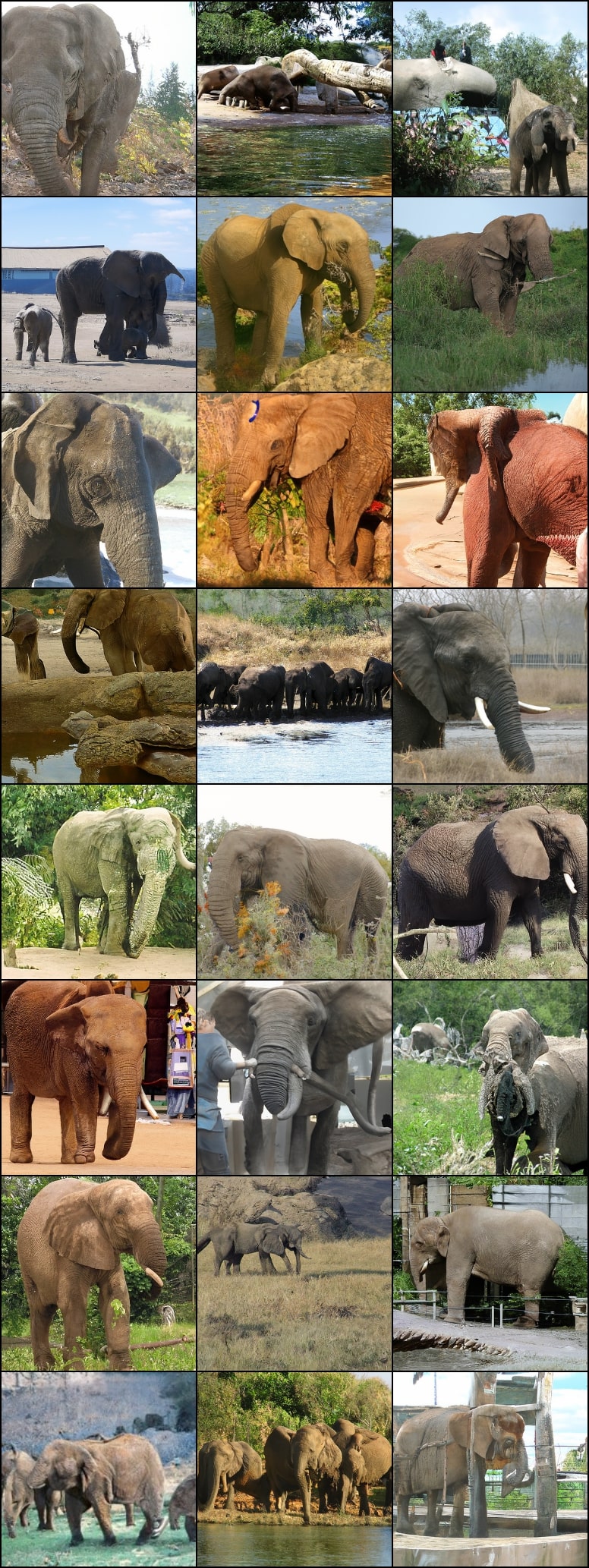}
		\subcaption{ADM-G++ (FID 4.45 recall 0.60)}
	\end{subfigure}
	\caption{Uncurated random samples from african elephant class (386) (a) ADM with poor FID (10.94) and good recall (0.63), (b) ADM-G++ with good FID (4.45) and good recall (0.60).}
	\label{fig:ImageNet256_recall_386}
\end{figure}

\begin{figure}[t]
	\centering
	\begin{subfigure}{0.48\linewidth}
		\centering
		\includegraphics[width=\linewidth]{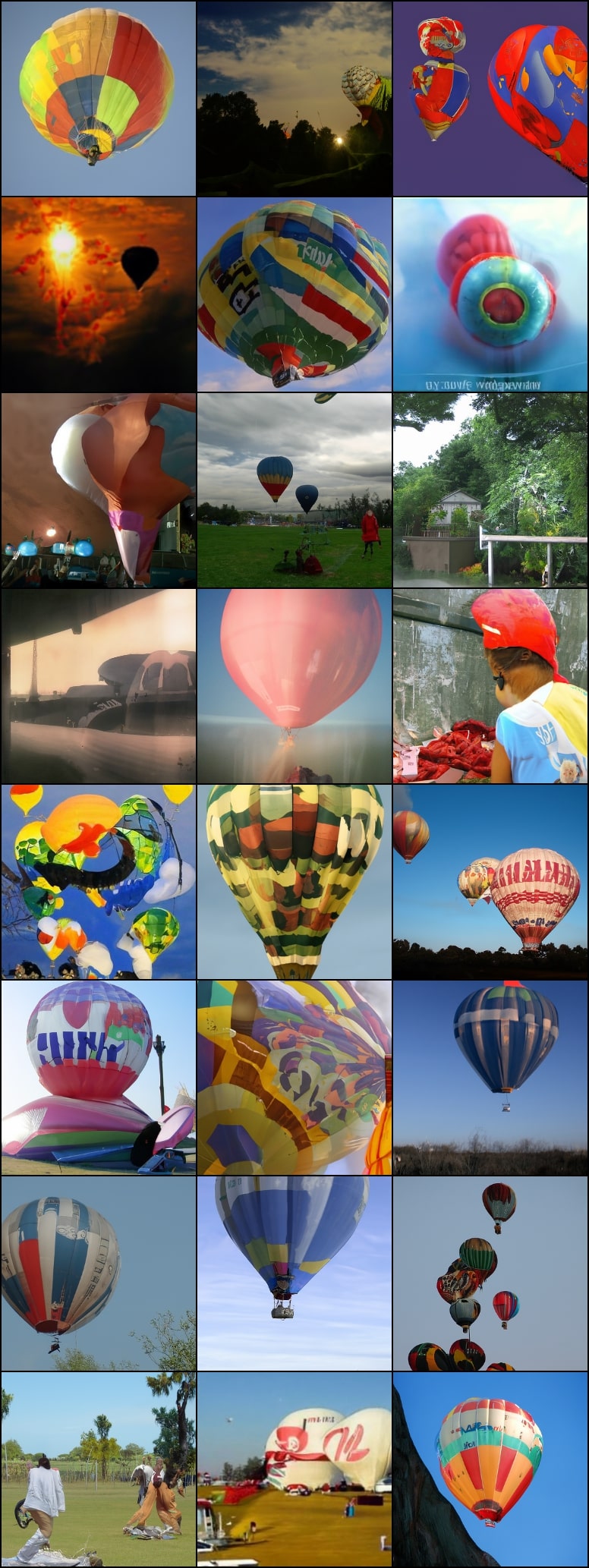}
		\subcaption{ADM (FID 10.94 recall 0.63)}
	\end{subfigure}
	\hfil
	\begin{subfigure}{0.48\linewidth}
		\centering
		\includegraphics[width=\linewidth]{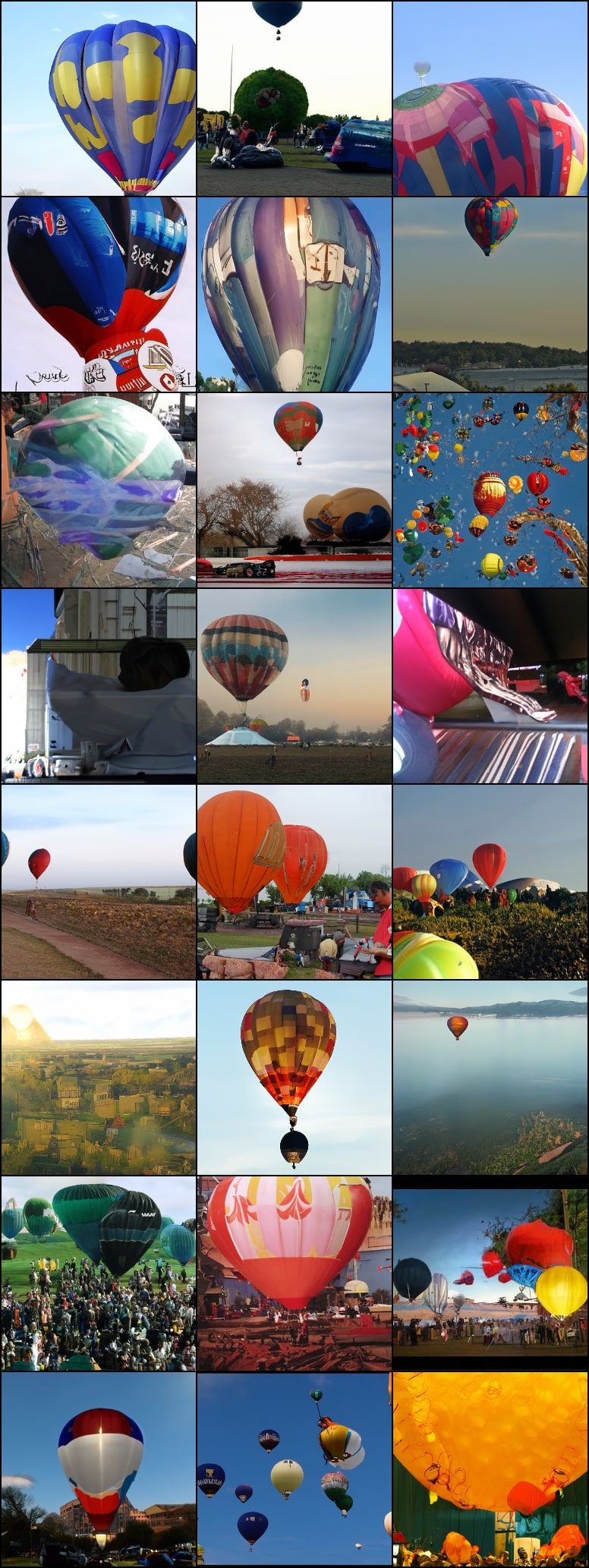}
		\subcaption{ADM-G++ (FID 4.45 recall 0.60)}
	\end{subfigure}
	\caption{Uncurated random samples from ballon class (417) (a) ADM with poor FID (10.94) and good recall (0.63), (b) ADM-G++ with good FID (4.45) and good recall (0.60).}
	\label{fig:ImageNet256_recall_417}
\end{figure}

\begin{figure}[t]
	\centering
	\begin{subfigure}{0.48\linewidth}
		\centering
		\includegraphics[width=\linewidth]{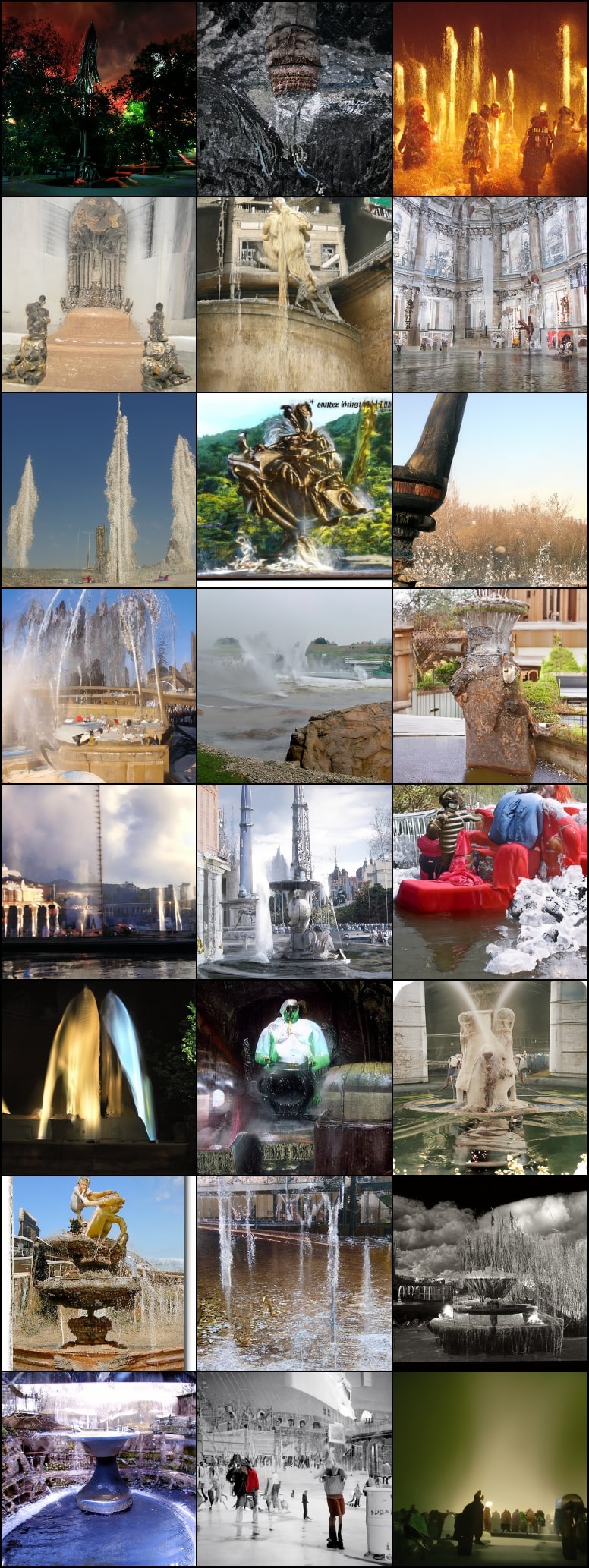}
		\subcaption{ADM (FID 10.94 recall 0.63)}
	\end{subfigure}
	\hfil
	\begin{subfigure}{0.48\linewidth}
		\centering
		\includegraphics[width=\linewidth]{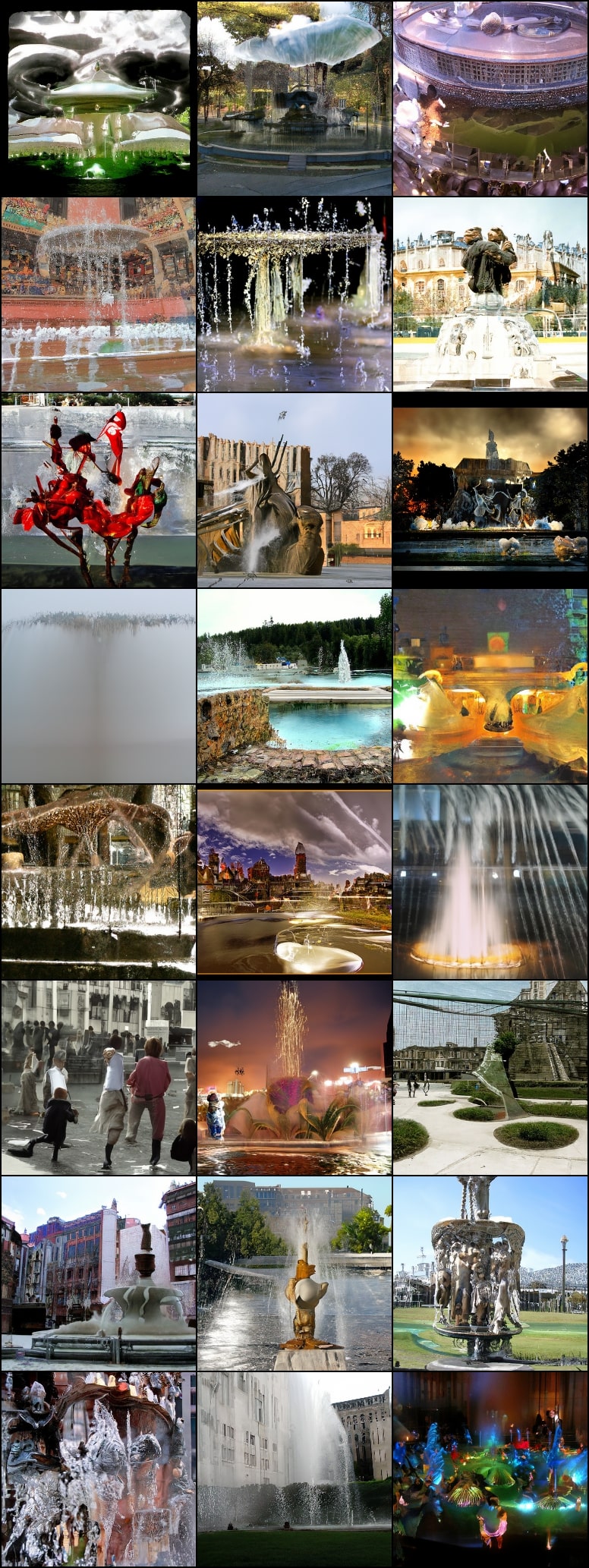}
		\subcaption{ADM-G++ (FID 4.45 recall 0.60)}
	\end{subfigure}
	\caption{Uncurated random samples from fountain class (562) (a) ADM with poor FID (10.94) and good recall (0.63), (b) ADM-G++ with good FID (4.45) and good recall (0.60).}
	\label{fig:ImageNet256_recall_562}
\end{figure}

Figures \ref{fig:ImageNet256_recall_90}, \ref{fig:ImageNet256_recall_130}, \ref{fig:ImageNet256_recall_281}, \ref{fig:ImageNet256_recall_323}, \ref{fig:ImageNet256_recall_386}, \ref{fig:ImageNet256_recall_417}, \ref{fig:ImageNet256_recall_562} compare ADM-G++ (cfg=0.10) with the vanilla ADM to illustrate how sample fidelity is improved while keeping the sample diversity. ADM-G++ (cfg=0.10) performs FID of 4.45 and recall of 0.60, and ADM performs FID of 10.94 and recall of 0.63.

\begin{figure}[t]
	\centering
	\begin{subfigure}{0.48\linewidth}
		\centering
		\includegraphics[width=\linewidth]{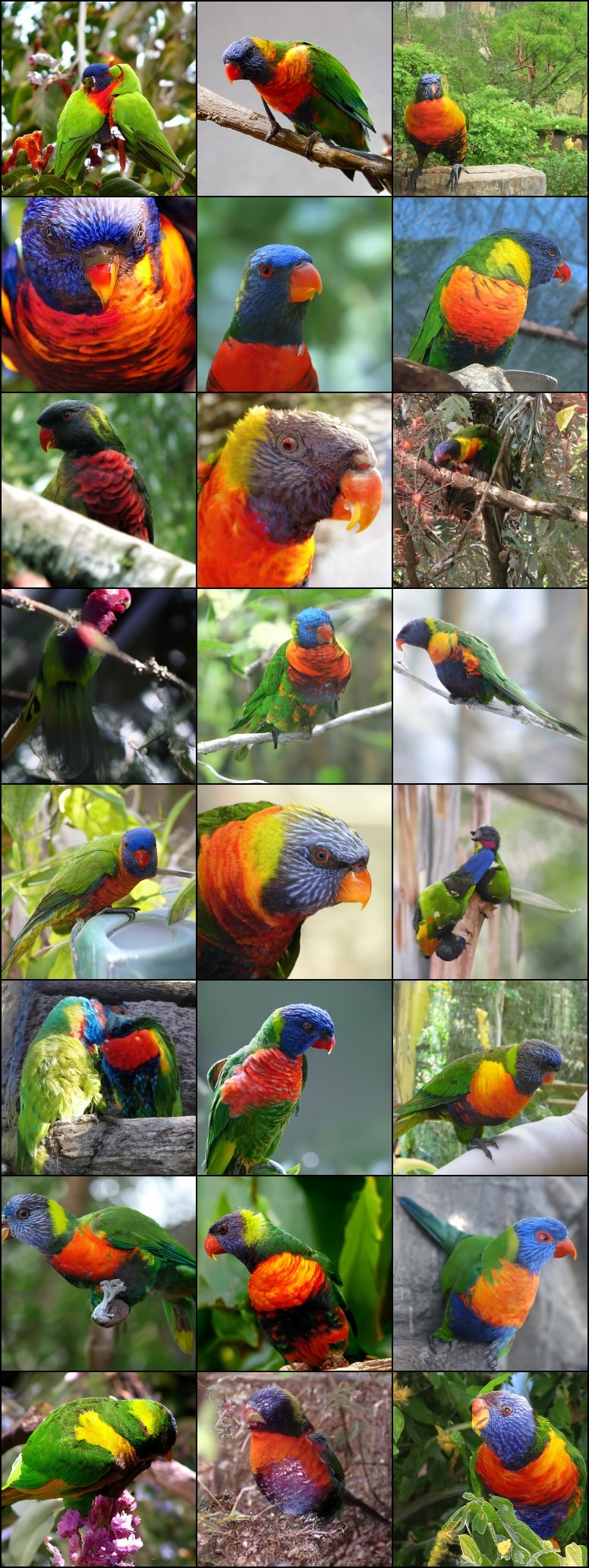}
		\subcaption{ADM-G (FID 4.59 recall 0.52)}
	\end{subfigure}
	\hfil
	\begin{subfigure}{0.48\linewidth}
		\centering
		\includegraphics[width=\linewidth]{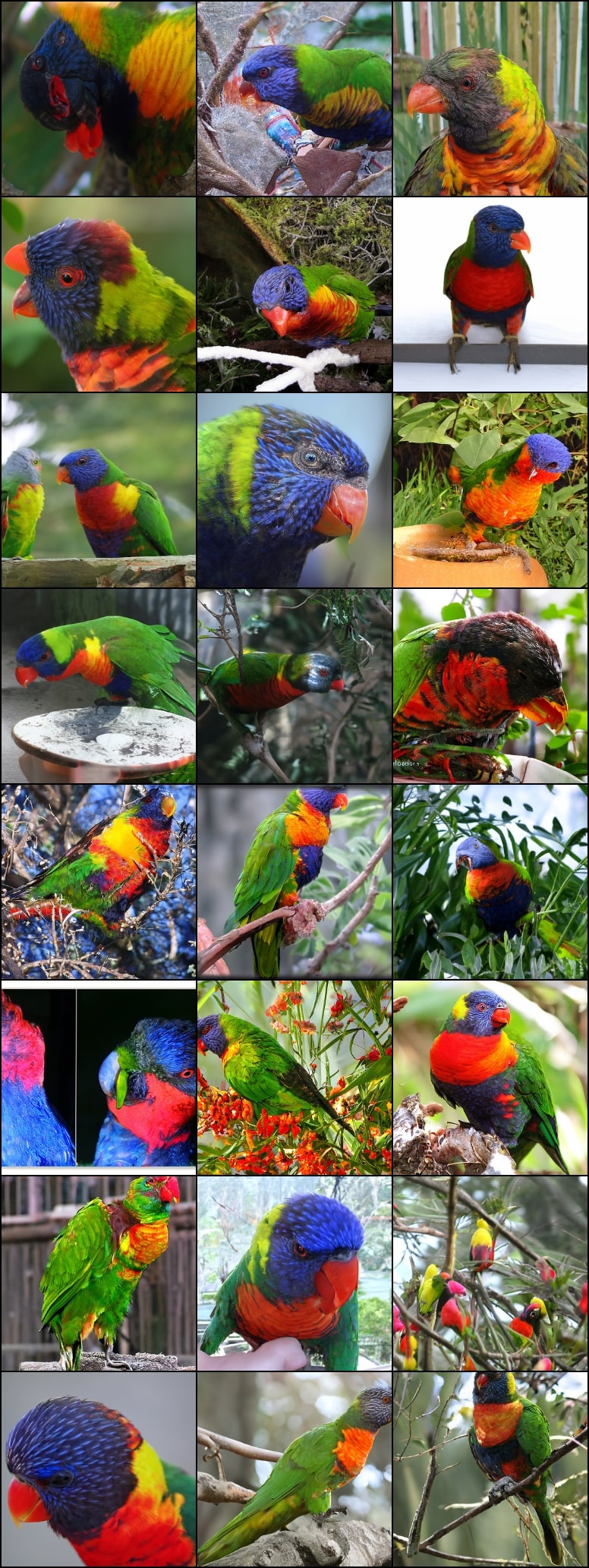}
		\subcaption{ADM-G++ (FID 3.18 recall 0.53)}
	\end{subfigure}
	\caption{Uncurated random samples from lorikeet class (90) (a) ADM-G with good FID (4.59) and poor recall (0.52), (b) ADM-G++ with SOTA FID (3.18) and moderate recall (0.53).}
	\label{fig:ImageNet256_FID_90}
\end{figure}

\begin{figure}[t]
	\centering
	\begin{subfigure}{0.48\linewidth}
		\centering
		\includegraphics[width=\linewidth]{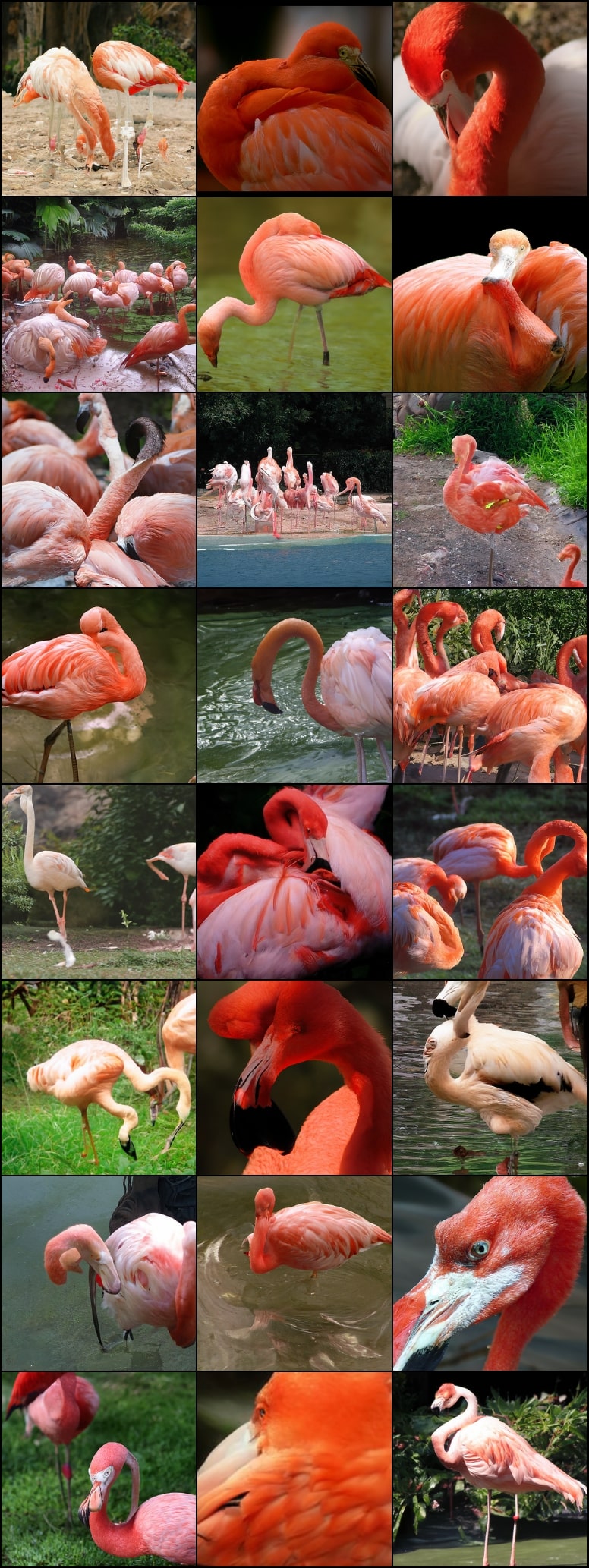}
		\subcaption{ADM-G (FID 4.59 recall 0.52)}
	\end{subfigure}
	\hfil
	\begin{subfigure}{0.48\linewidth}
		\centering
		\includegraphics[width=\linewidth]{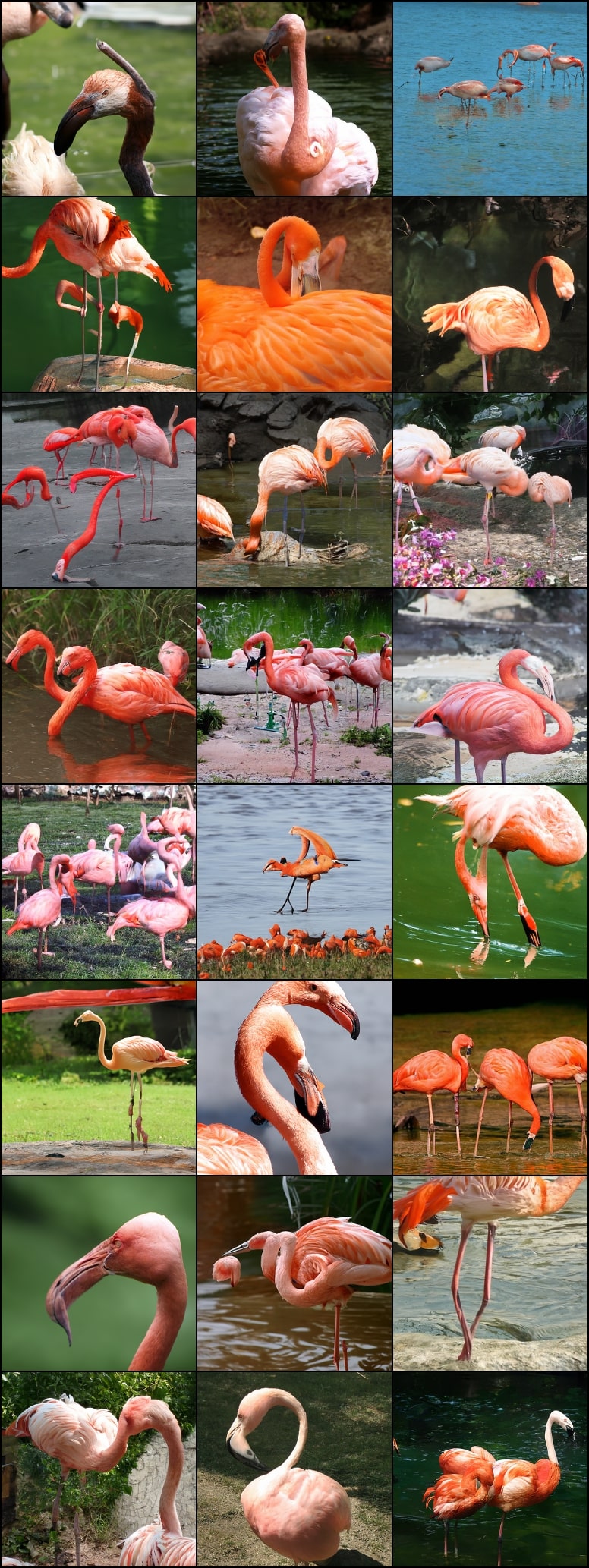}
		\subcaption{ADM-G++ (FID 3.18 recall 0.53)}
	\end{subfigure}
	\caption{Uncurated random samples from flamingo class (130) (a) ADM-G with good FID (4.59) and poor recall (0.52), (b) ADM-G++ with SOTA FID (3.18) and moderate recall (0.53).}
	\label{fig:ImageNet256_FID_130}
\end{figure}

\begin{figure}[t]
	\centering
	\begin{subfigure}{0.48\linewidth}
		\centering
		\includegraphics[width=\linewidth]{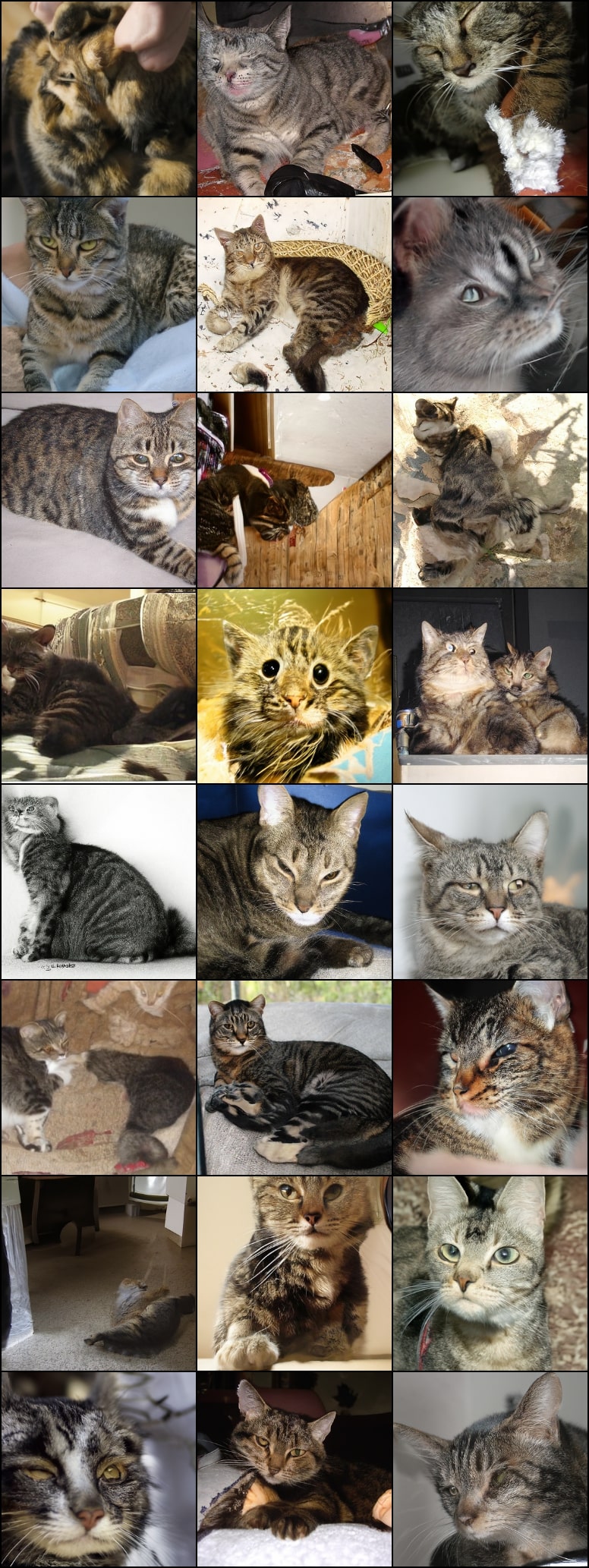}
		\subcaption{ADM-G (FID 4.59 recall 0.52)}
	\end{subfigure}
	\hfil
	\begin{subfigure}{0.48\linewidth}
		\centering
		\includegraphics[width=\linewidth]{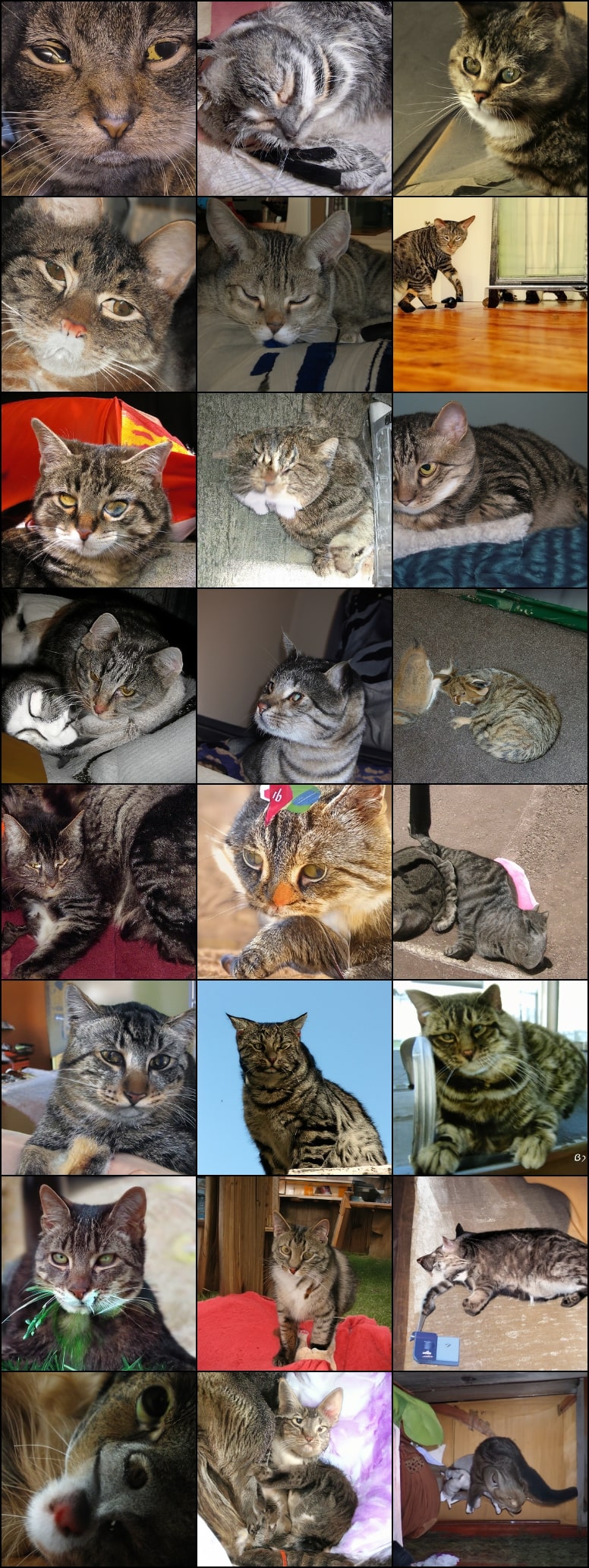}
		\subcaption{ADM-G++ (FID 3.18 recall 0.53)}
	\end{subfigure}
	\caption{Uncurated random samples from tabby cat class (281) (a) ADM-G with good FID (4.59) and poor recall (0.52), (b) ADM-G++ with SOTA FID (3.18) and moderate recall (0.53).}
	\label{fig:ImageNet256_FID_281}
\end{figure}

\begin{figure}[t]
	\centering
	\begin{subfigure}{0.48\linewidth}
		\centering
		\includegraphics[width=\linewidth]{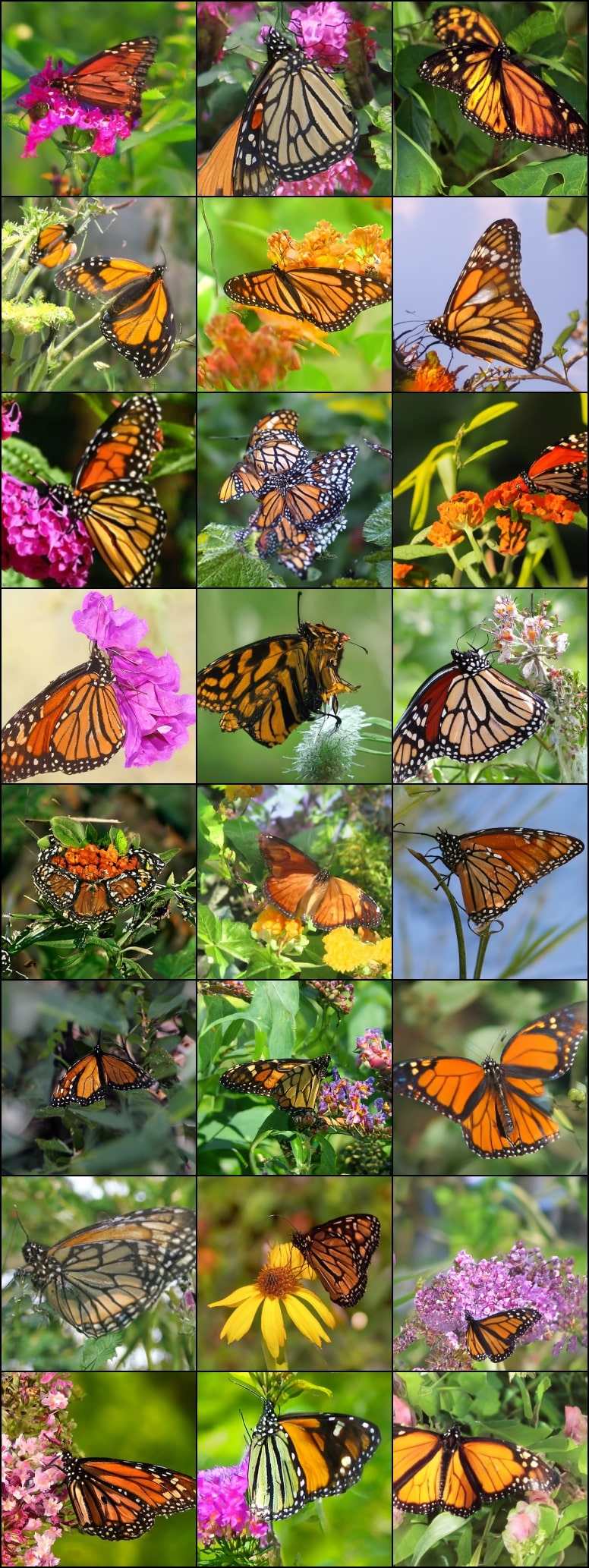}
		\subcaption{ADM-G (FID 4.59 recall 0.52)}
	\end{subfigure}
	\hfil
	\begin{subfigure}{0.48\linewidth}
		\centering
		\includegraphics[width=\linewidth]{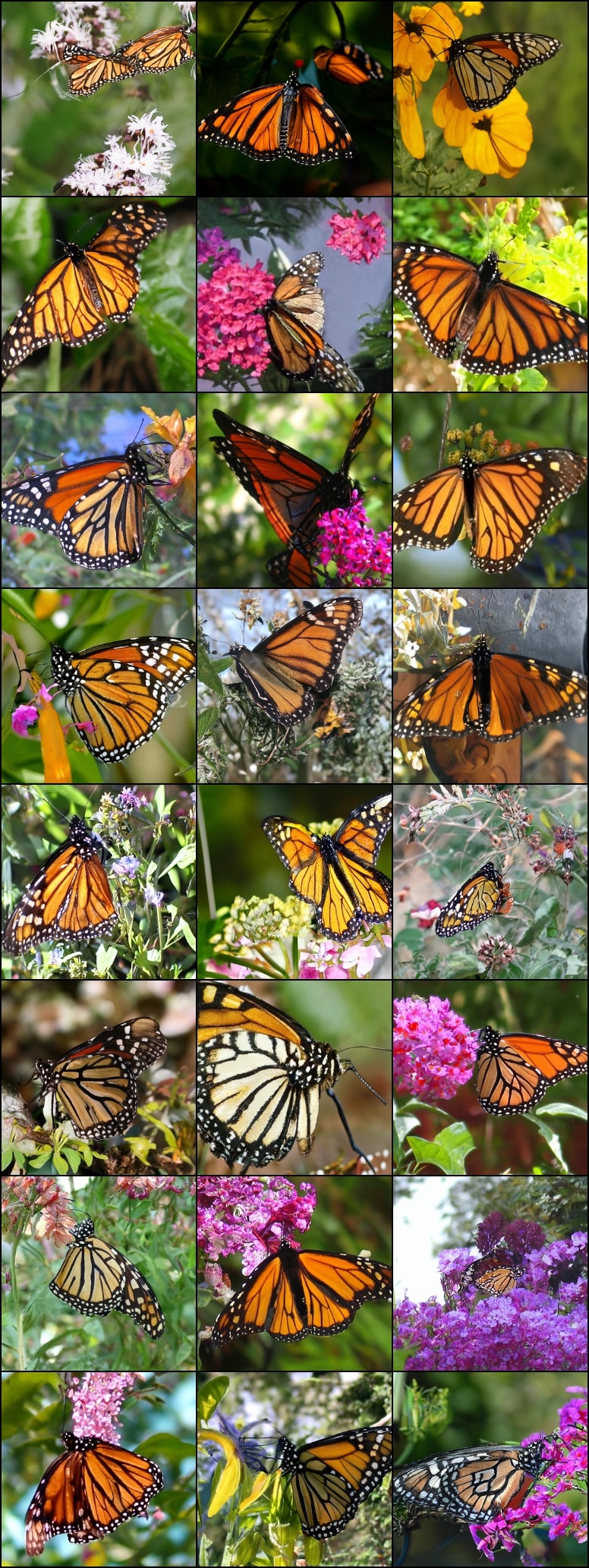}
		\subcaption{ADM-G++ (FID 3.18 recall 0.53)}
	\end{subfigure}
	\caption{Uncurated random samples from monarch butterfly class (323) (a) ADM-G with good FID (4.59) and poor recall (0.52), (b) ADM-G++ with SOTA FID (3.18) and moderate recall (0.53).}
	\label{fig:ImageNet256_FID_323}
\end{figure}

\begin{figure}[t]
	\centering
	\begin{subfigure}{0.48\linewidth}
		\centering
		\includegraphics[width=\linewidth]{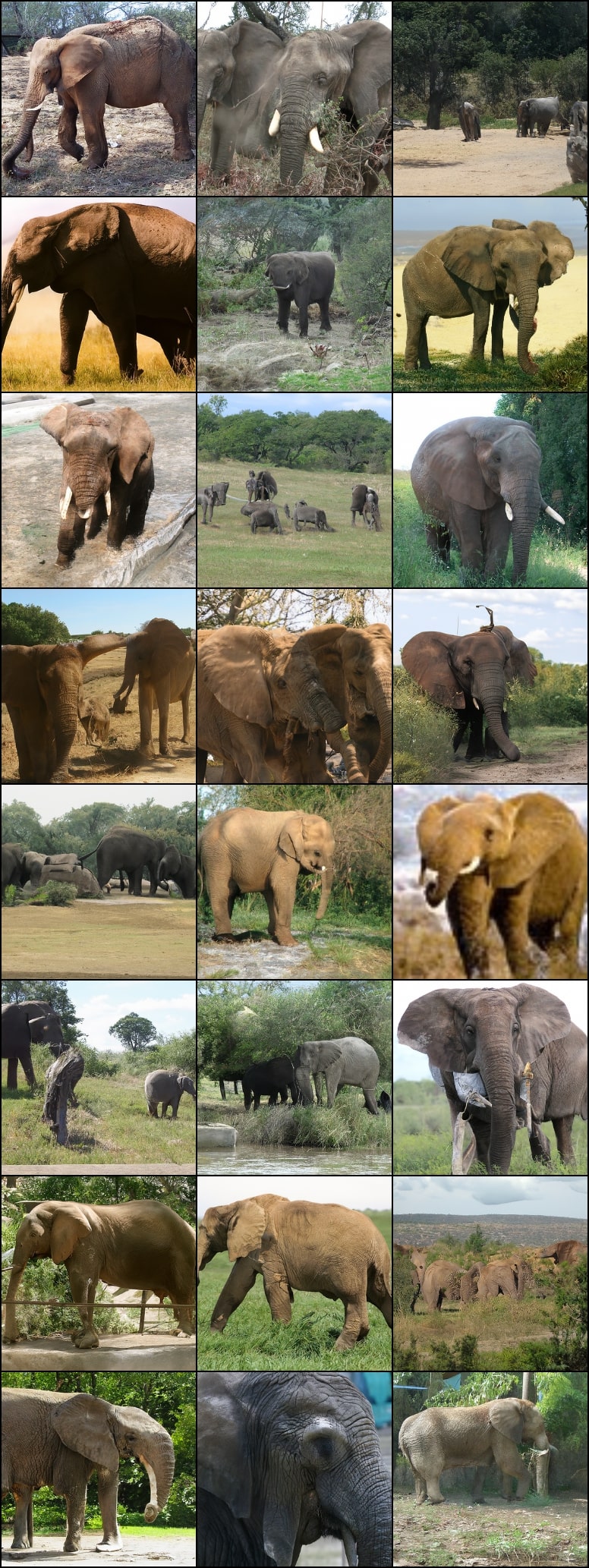}
		\subcaption{ADM-G (FID 4.59 recall 0.52)}
	\end{subfigure}
	\hfil
	\begin{subfigure}{0.48\linewidth}
		\centering
		\includegraphics[width=\linewidth]{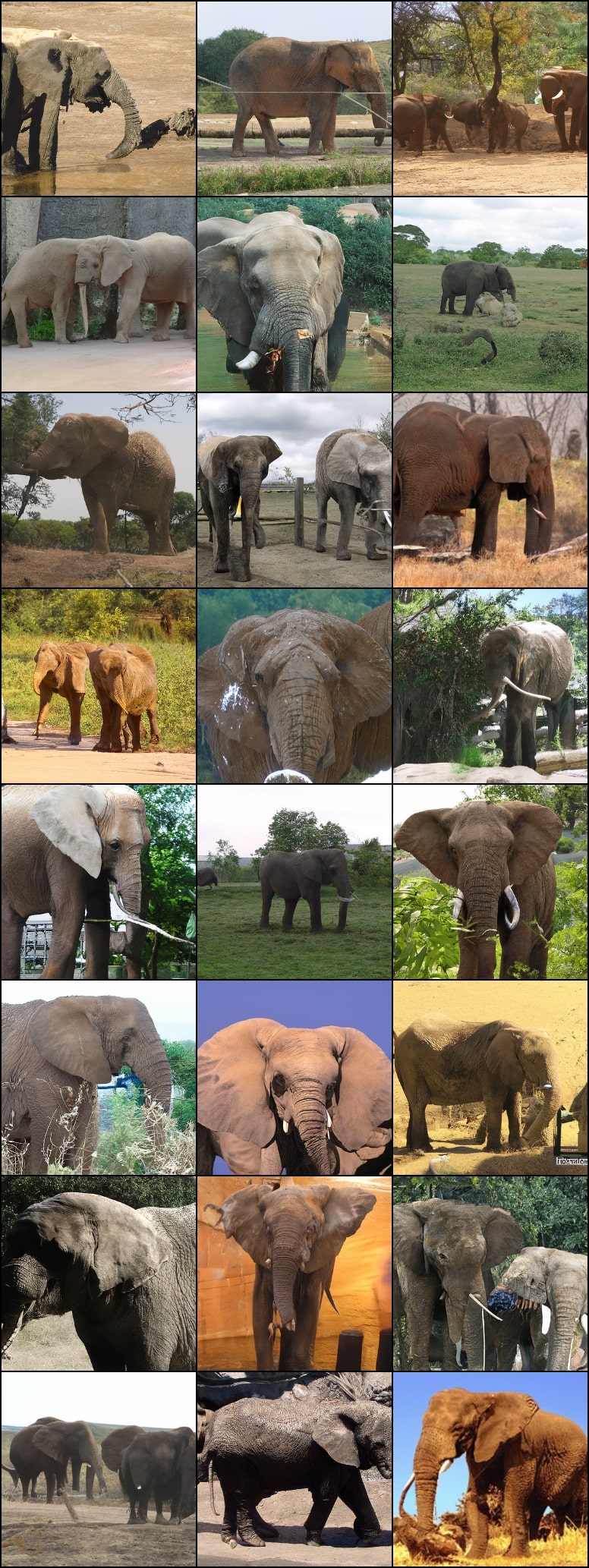}
		\subcaption{ADM-G++ (FID 3.18 recall 0.53)}
	\end{subfigure}
	\caption{Uncurated random samples from african elephant class (386) (a) ADM-G with good FID (4.59) and poor recall (0.52), (b) ADM-G++ with SOTA FID (3.18) and moderate recall (0.53).}
	\label{fig:ImageNet256_FID_386}
\end{figure}

\begin{figure}[t]
	\centering
	\begin{subfigure}{0.48\linewidth}
		\centering
		\includegraphics[width=\linewidth]{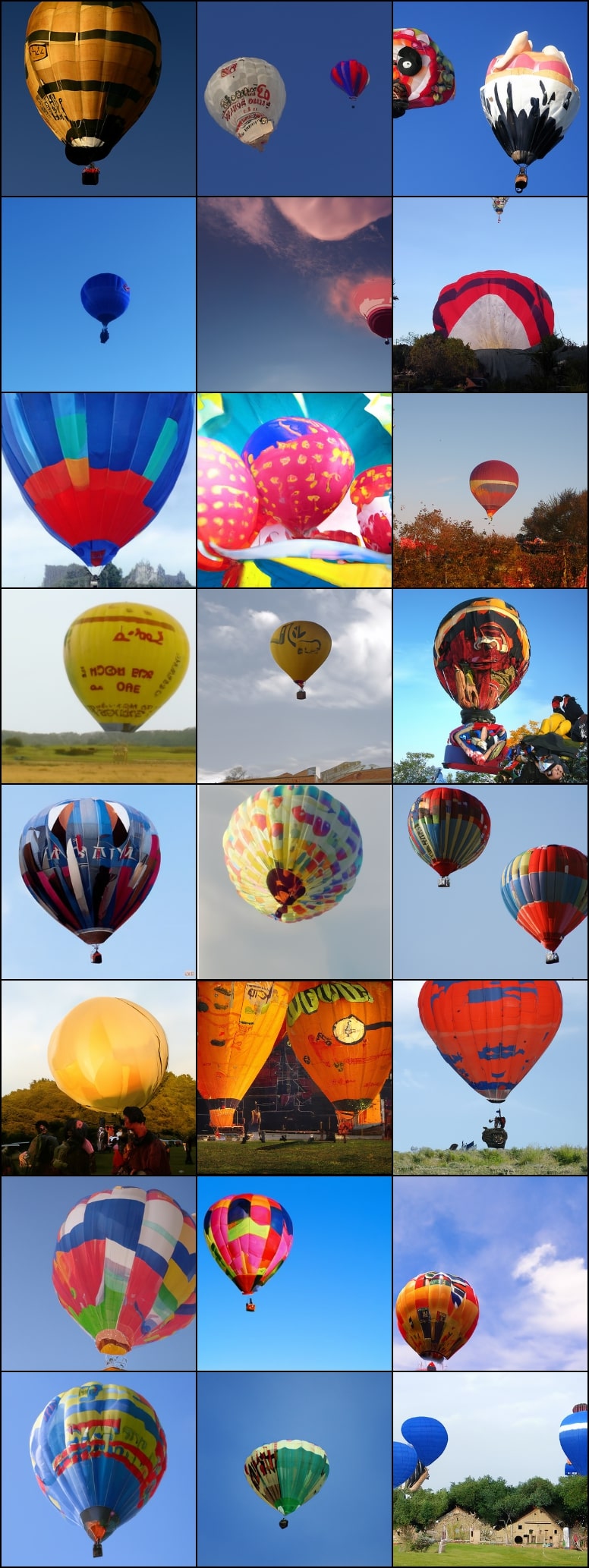}
		\subcaption{ADM-G (FID 4.59 recall 0.52)}
	\end{subfigure}
	\hfil
	\begin{subfigure}{0.48\linewidth}
		\centering
		\includegraphics[width=\linewidth]{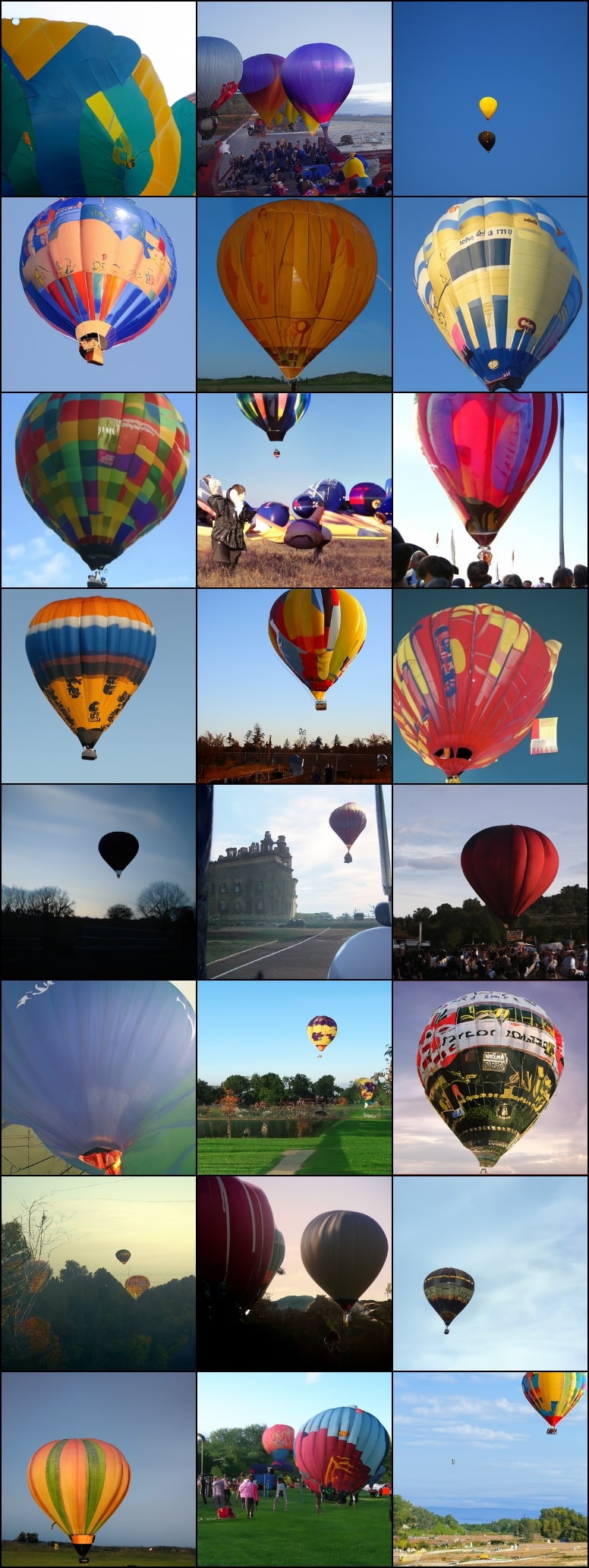}
		\subcaption{ADM-G++ (FID 3.18 recall 0.53)}
	\end{subfigure}
	\caption{Uncurated random samples from ballon class (417) (a) ADM-G with good FID (4.59) and poor recall (0.52), (b) ADM-G++ with SOTA FID (3.18) and moderate recall (0.53).}
	\label{fig:ImageNet256_FID_417}
\end{figure}

\begin{figure}[t]
	\centering
	\begin{subfigure}{0.48\linewidth}
		\centering
		\includegraphics[width=\linewidth]{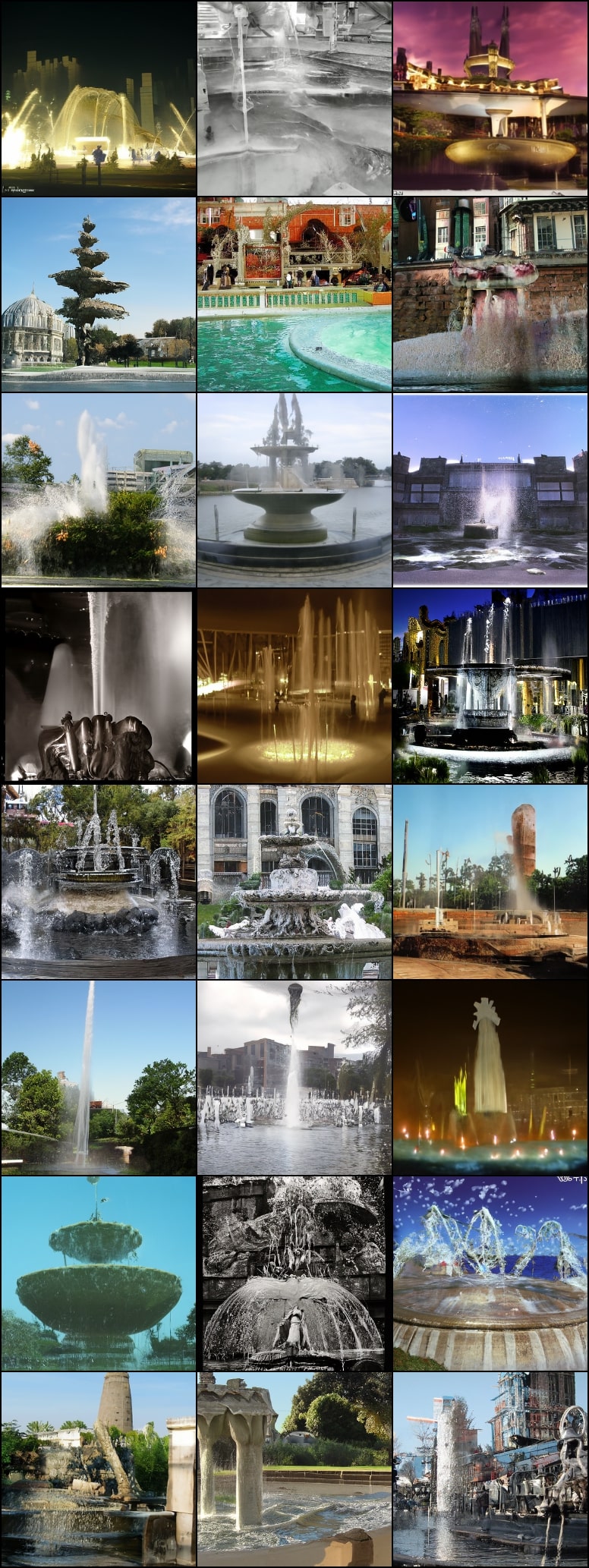}
		\subcaption{ADM-G (FID 4.59 recall 0.52)}
	\end{subfigure}
	\hfil
	\begin{subfigure}{0.48\linewidth}
		\centering
		\includegraphics[width=\linewidth]{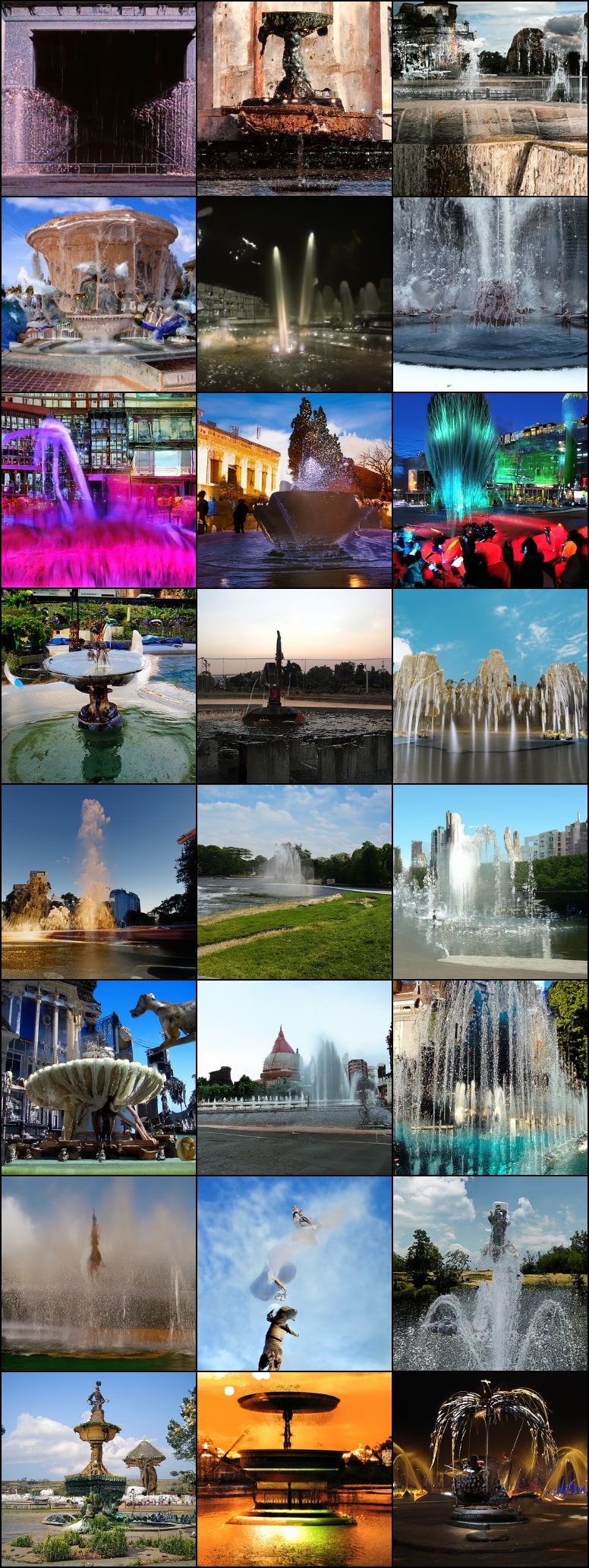}
		\subcaption{ADM-G++ (FID 3.18 recall 0.53)}
	\end{subfigure}
	\caption{Uncurated random samples from fountain class (562) (a) ADM-G with good FID (4.59) and poor recall (0.52), (b) ADM-G++ with SOTA FID (3.18) and moderate recall (0.53).}
	\label{fig:ImageNet256_FID_562}
\end{figure}

Figures \ref{fig:ImageNet256_FID_90}, \ref{fig:ImageNet256_FID_130}, \ref{fig:ImageNet256_FID_281}, \ref{fig:ImageNet256_FID_323}, \ref{fig:ImageNet256_FID_386}, \ref{fig:ImageNet256_FID_417}, \ref{fig:ImageNet256_FID_562} compare ADM-G++ (cfg=0.75) with ADM-G (cfg=1.50). These figures show the discriminator guidance is effective in high-dimensional dataset. 

\begin{figure*}[t]
	\centering
		\includegraphics[width=\linewidth]{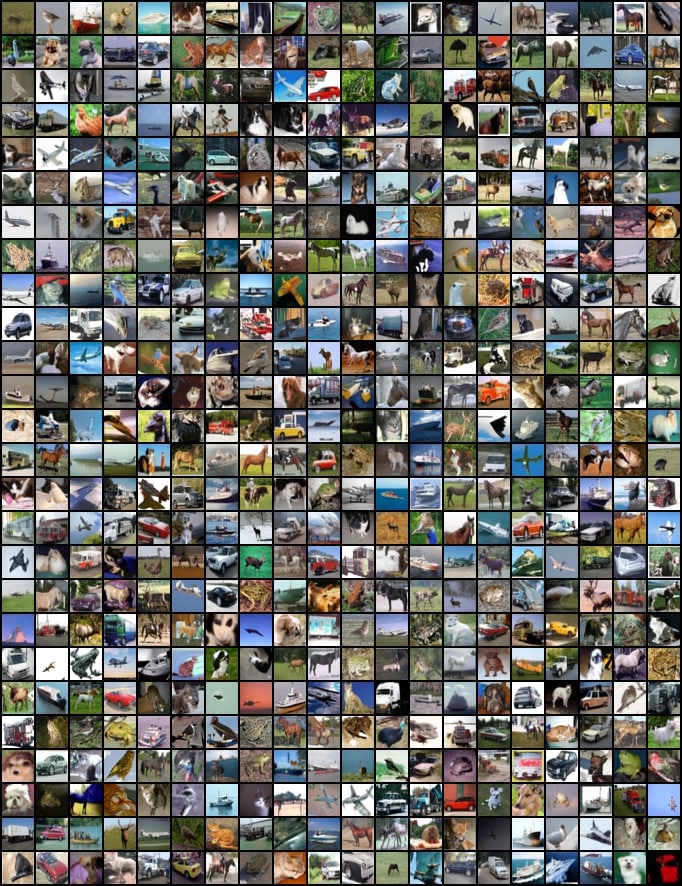}
	\caption{Uncurated random samples from LSGM-G++ on unconditional CIFAR10 (FID: 1.94).}
	\label{fig:unconditional_CIFAR10_LSGM-G++}
\end{figure*}

\begin{figure*}[t]
	\centering
		\includegraphics[width=\linewidth]{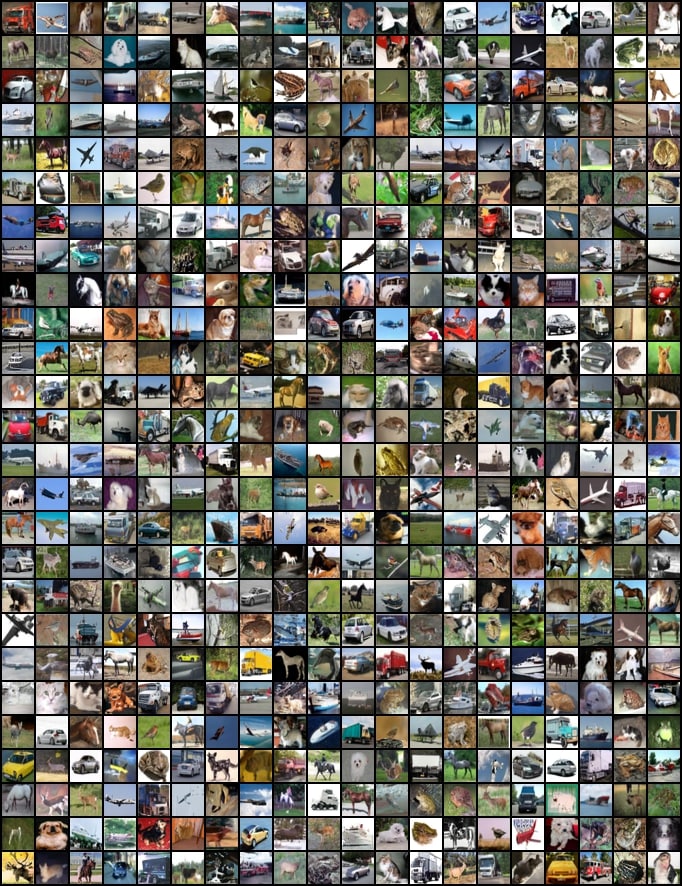}
	\caption{Uncurated random samples from EDM-G++ on unconditional CIFAR10 (FID: 1.77).}
	\label{fig:unconditional_CIFAR10_EDM-G++}
\end{figure*}

\begin{figure*}[t]
	\centering
		\includegraphics[width=\linewidth]{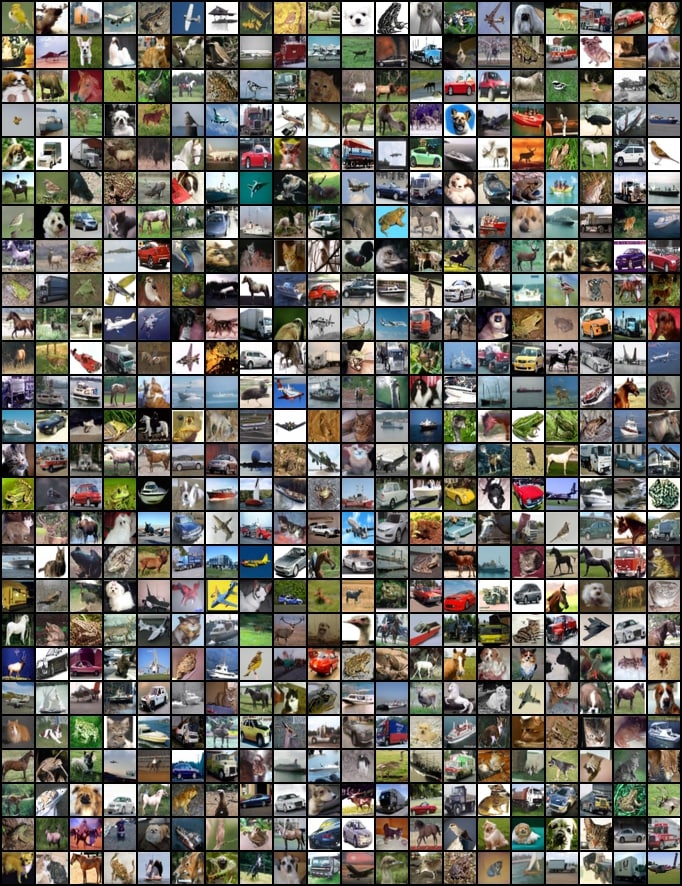}
	\caption{Uncurated random samples from EDM-G++ on conditional CIFAR10 (FID: 1.64).}
	\label{fig:conditional_CIFAR10_EDM-G++}
\end{figure*}

\begin{figure*}[t]
	\centering
		\includegraphics[width=\linewidth]{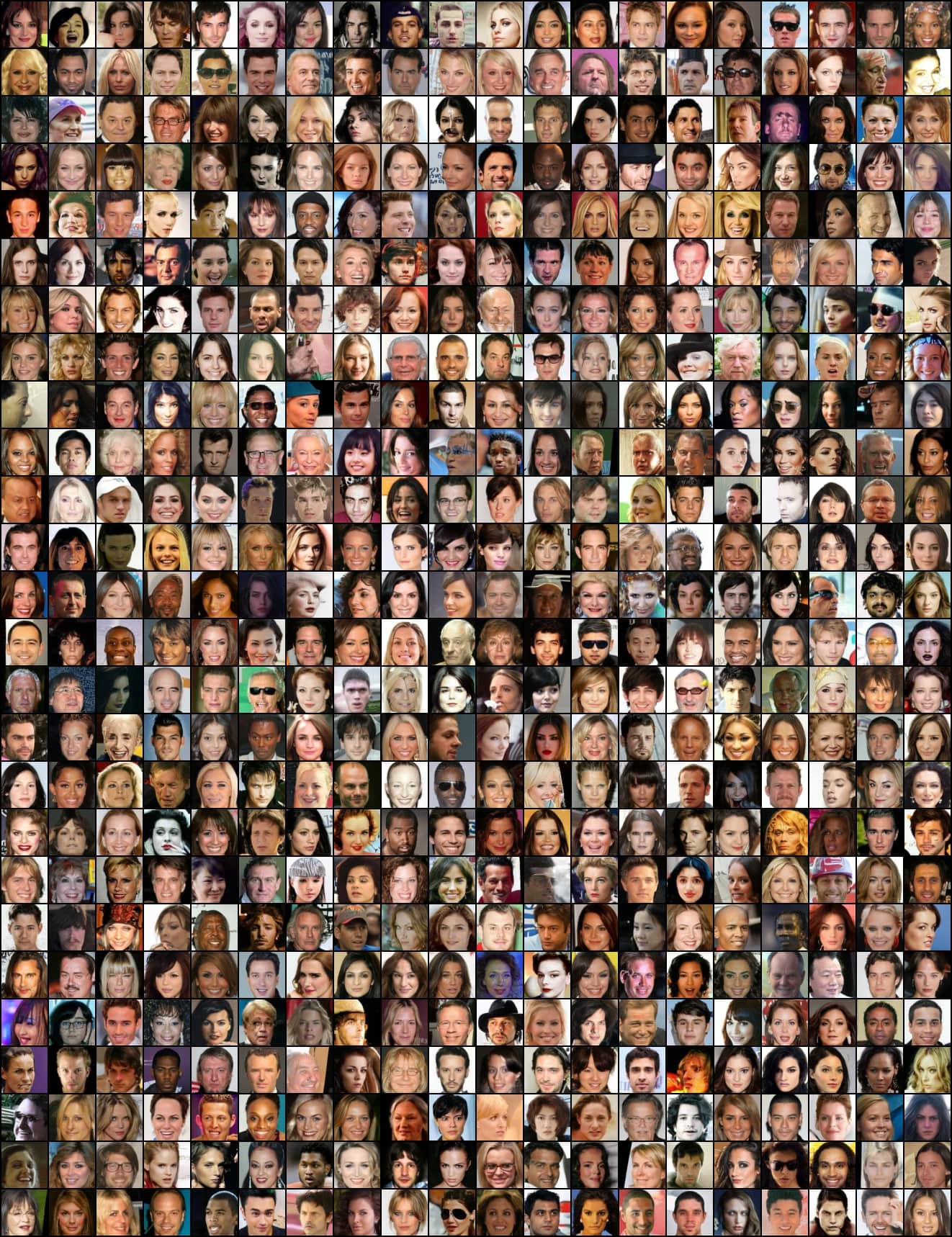}
	\caption{Uncurated random samples from Soft Truncation-G++ on unconditional CelebA (FID: 1.34).}
	\label{fig:celeba}
\end{figure*}

\begin{figure*}[t]
	\centering
		\includegraphics[width=\linewidth]{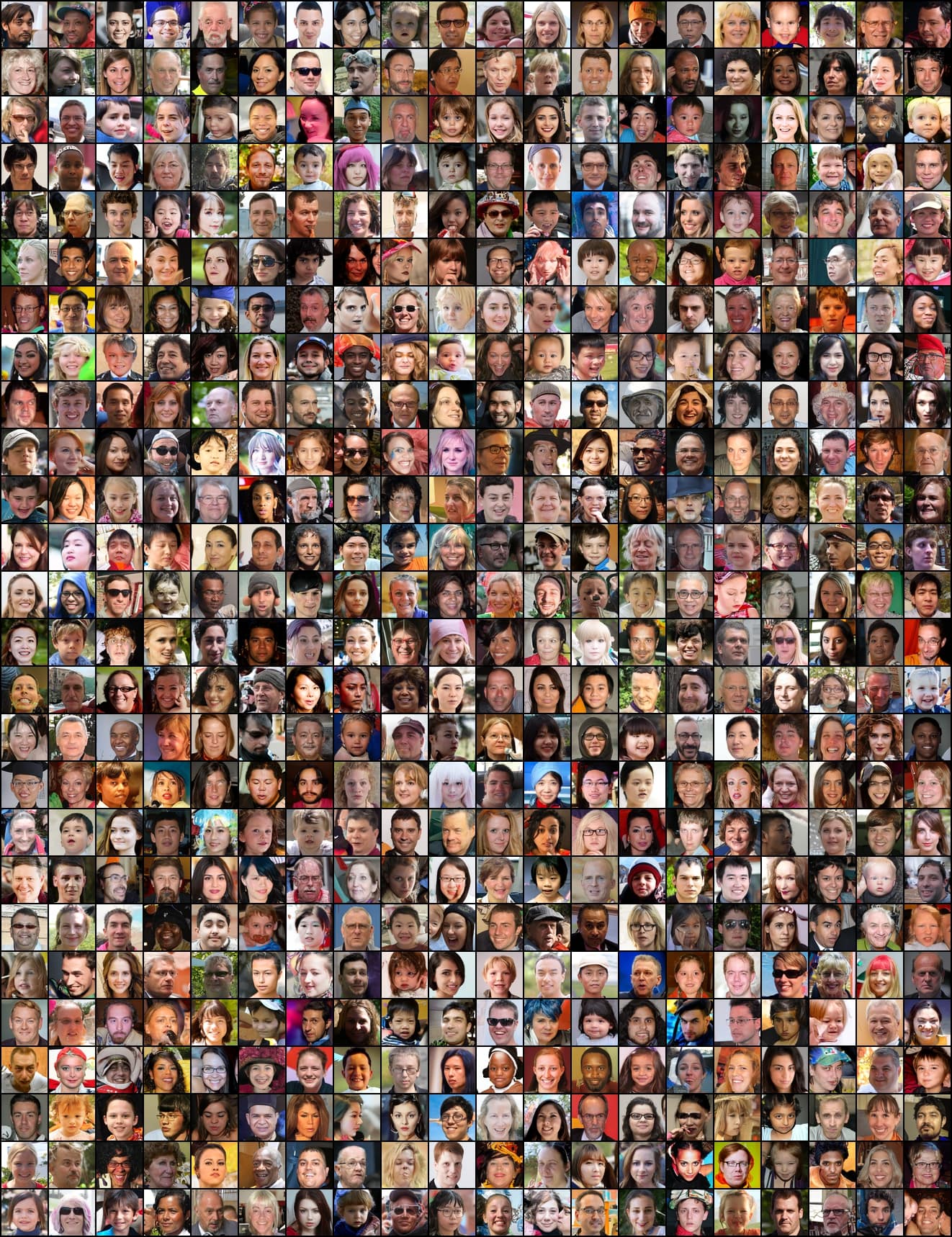}
	\caption{Uncurated random samples from EDM-G++ on unconditional FFHQ (FID: 1.98).}
	\label{fig:ffhq}
\end{figure*}

Figures \ref{fig:unconditional_CIFAR10_LSGM-G++}, \ref{fig:unconditional_CIFAR10_EDM-G++}, \ref{fig:conditional_CIFAR10_EDM-G++}, \ref{fig:celeba}, and \ref{fig:ffhq} show uncurated samples from unconditional CIFAR-10 with LSGM-G++, unconditional CIFAR-10 with EDM-G++, conditional CIFAR-10 with EDM-G++, unconditional CelebA with Soft Truncation-G++, and unconditional FFHQ with EDM-G++.

\begin{figure*}[t]
	\centering
	\begin{subfigure}{0.29\linewidth}
		\centering
		\includegraphics[width=\linewidth]{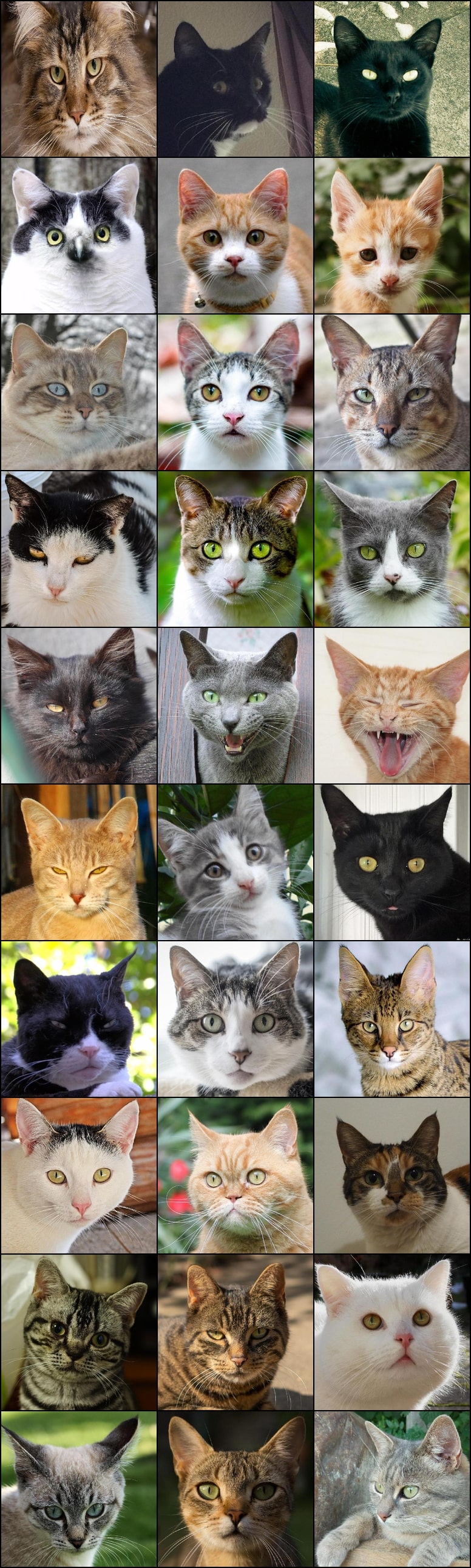}
		\subcaption{Cat (Source)}
	\end{subfigure}
	\hfil
	\begin{subfigure}{0.29\linewidth}
		\centering
		\includegraphics[width=\linewidth]{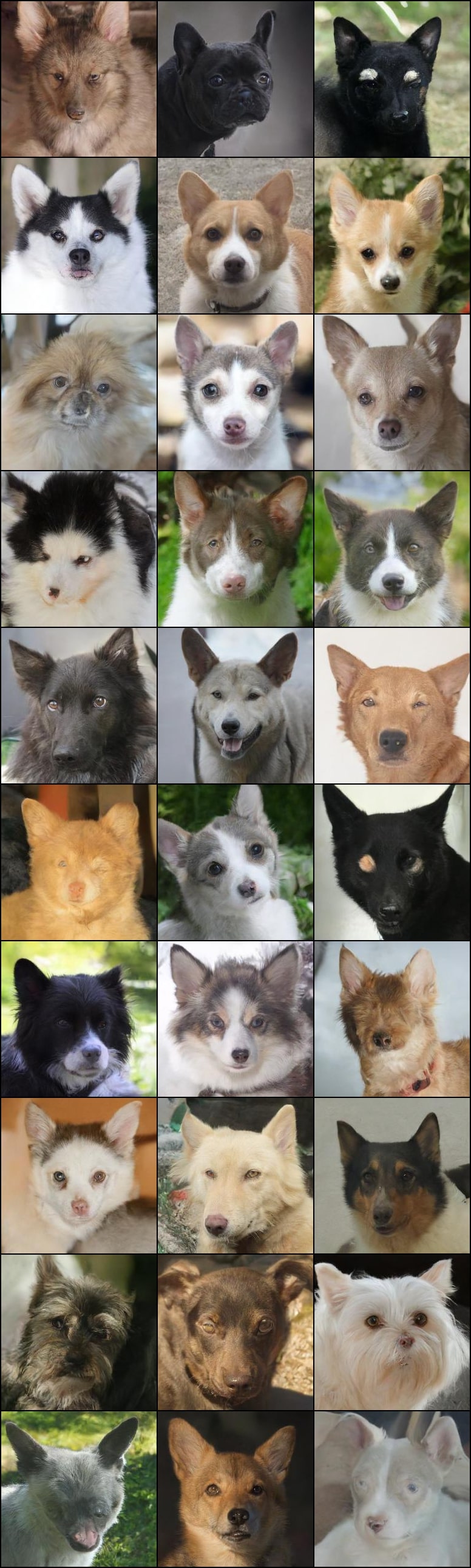}
		\subcaption{SDEdit (FID: 74.02)}
	\end{subfigure}		
	\hfil
	\begin{subfigure}{0.29\linewidth}
		\centering
		\includegraphics[width=\linewidth]{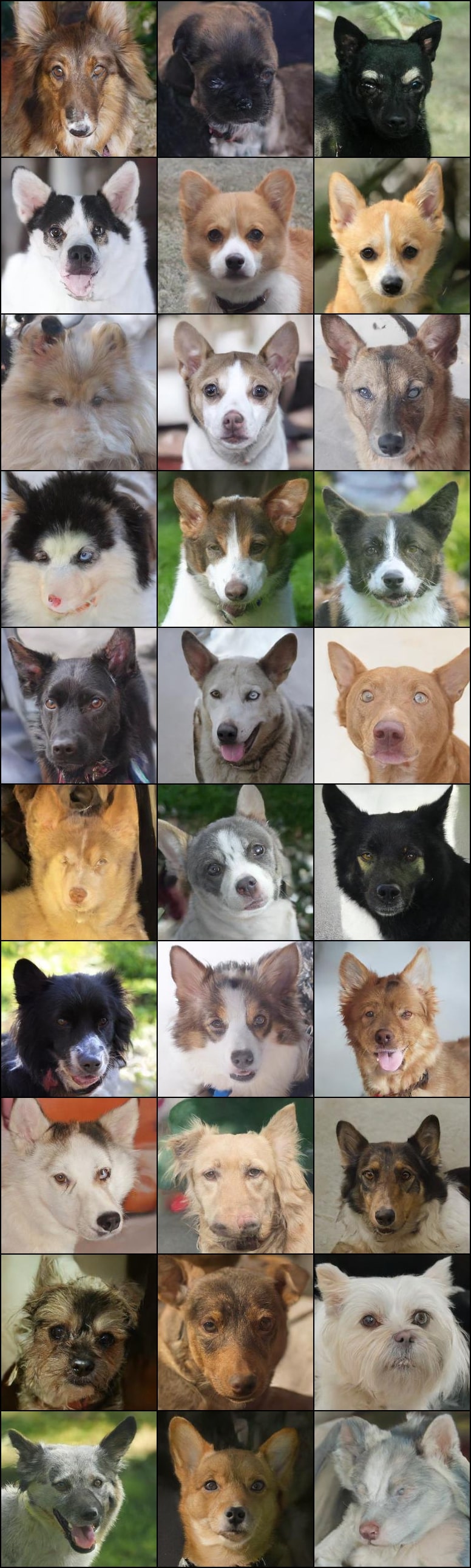}
		\subcaption{SDEdit + DG (FID: 61.92)}
	\end{subfigure}
	\caption{Uncurated random translated samples from (a) source cat, (b) SDEdit (FID: 74.02, L2: 49.22, PSNR: 19.21, SSIM: 0.42), and (c) SDEdit + DG with $w_{t}^{DG}=8$ (FID: 61.92, L2: 50.62, PSNR: 18.94, SSIM: 0.41).}
	\label{fig:I2I_uncurated}
\end{figure*}

Figure \ref{fig:I2I_uncurated} shows the uncurated samples from I2I translation task.

\end{document}